\documentclass[10pt,journal,compsoc]{IEEEtran}
\usepackage{amsmath,amsfonts}
\usepackage{algorithmic}
\usepackage{algorithm}
\usepackage{array}
\usepackage[caption=false,font=normalsize,labelfont=sf,textfont=sf]{subfig}
\usepackage{textcomp}
\usepackage{stfloats}
\usepackage{url}
\usepackage{verbatim}
\usepackage{graphicx}
\usepackage{cite}

\usepackage{CJKutf8}  
\usepackage{booktabs}
\usepackage{tabularx}
\usepackage{caption}
\usepackage{pifont}  
\usepackage{xltabular}
\usepackage{makecell}   
\usepackage{enumitem}
\usepackage{colortbl} 
\usepackage[colorlinks,linkcolor=red,anchorcolor=blue,citecolor=green]{hyperref}

\definecolor{lightgray}{gray}{0.9} 

\newcolumntype{C}[1]{>{\centering\arraybackslash}p{#1}} 
\newcolumntype{L}{>{\arraybackslash}X} 
\newcolumntype{Y}{>{\centering\arraybackslash}X} 
\newcolumntype{H}[1]{>{\centering\arraybackslash}m{#1}}

\begin{document}
\bstctlcite{IEEEexample:BSTcontrol}

\title{Large VLM-based Vision-Language-Action Models for Robotic Manipulation: A Survey}


\author{
    Rui Shao, Wei Li, Lingsen Zhang, Renshan Zhang, Zhiyang Liu, Ran Chen, Liqiang Nie
    \thanks{
    Rui Shao, Wei Li, Lingsen Zhang, Renshan Zhang, Zhiyang Liu, Ran Chen and Liqiang Nie are with the School of Computer Science and Technology, Harbin Institute of Technology (Shenzhen), China, 518055 (e-mail: shaorui@hit.edu.cn, \{liwei2024, zls030726, zhangrenshan, 25s151095, 220310324\}@stu.hit.edu.cn, nieliqiang@gmail.com).
    \\
    Wei Li is corresponding author.}
}

\markboth{Journal of \LaTeX\ Class Files,~Vol.~14, No.~8, August~2021}%
{Shell \MakeLowercase{\textit{et al.}}: Task_Assignments: VLA Survey}



\IEEEtitleabstractindextext{%
\begin{abstract}

Robotic manipulation, as a critical frontier in robotics and embodied AI, demands precise motor control and integrated understanding of visual and semantic cues in dynamic environments. Traditional approaches, grounded in predefined task specifications and rigid control policies, often struggle to scale or generalize in unstructured, novel scenarios.
In recent years, Vision-Language-Action (VLA) models, built upon Large Vision-Language Models (VLMs) pretrained on vast image-text datasets, have become an increasingly studied paradigm. By leveraging large VLMs’ capabilities in open-world generalization, hierarchical task planning, knowledge-augmented reasoning, and rich multimodal fusion, these models empower robots to interpret high-level instructions, recognize unseen environments and execute complex manipulation tasks. 
This survey provides the first systematic, taxonomy-oriented review of large VLM-based VLA models for robotic manipulation. We begin by clearly defining large VLM-based VLA models and delineating two principal architectural paradigms: \textbf{(1) monolithic models}, encompassing single-system and dual-system designs with differing levels of integration; and \textbf{(2) hierarchical models}, which explicitly decouple planning from execution via interpretable intermediate representations.
Building on this foundation, we present an in-depth examination of large VLM-based VLA models:
\textbf{(1)} integration with advanced domains, including reinforcement learning, training-free optimization, learning from human videos, and world model integration;
\textbf{(2)} synthesis of distinctive characteristics, consolidating architectural traits, operational strengths, and the datasets and benchmarks that support their development;
\textbf{(3)} identification of promising directions, including memory mechanisms, 4D perception, efficient adaptation, multi-agent cooperation, and other emerging capabilities.
This survey consolidates recent advances to resolve inconsistencies in existing taxonomies, mitigate research fragmentation, and fill a critical gap through the systematic integration of studies at the intersection of large VLMs and robotic manipulation.
We provide a regularly updated project page to document ongoing progress: \href{https://github.com/JiuTian-VL/Large-VLM-based-VLA-for-Robotic-Manipulation}{\textcolor{red}{https://github.com/JiuTian-VL/Large VLM-based VLA for Robotic Manipulation}}.
\end{abstract}

}

\maketitle

\IEEEdisplaynontitleabstractindextext

\IEEEpeerreviewmaketitle

\begin{figure}[!h]
    \centering
    \includegraphics[width=0.92\linewidth]{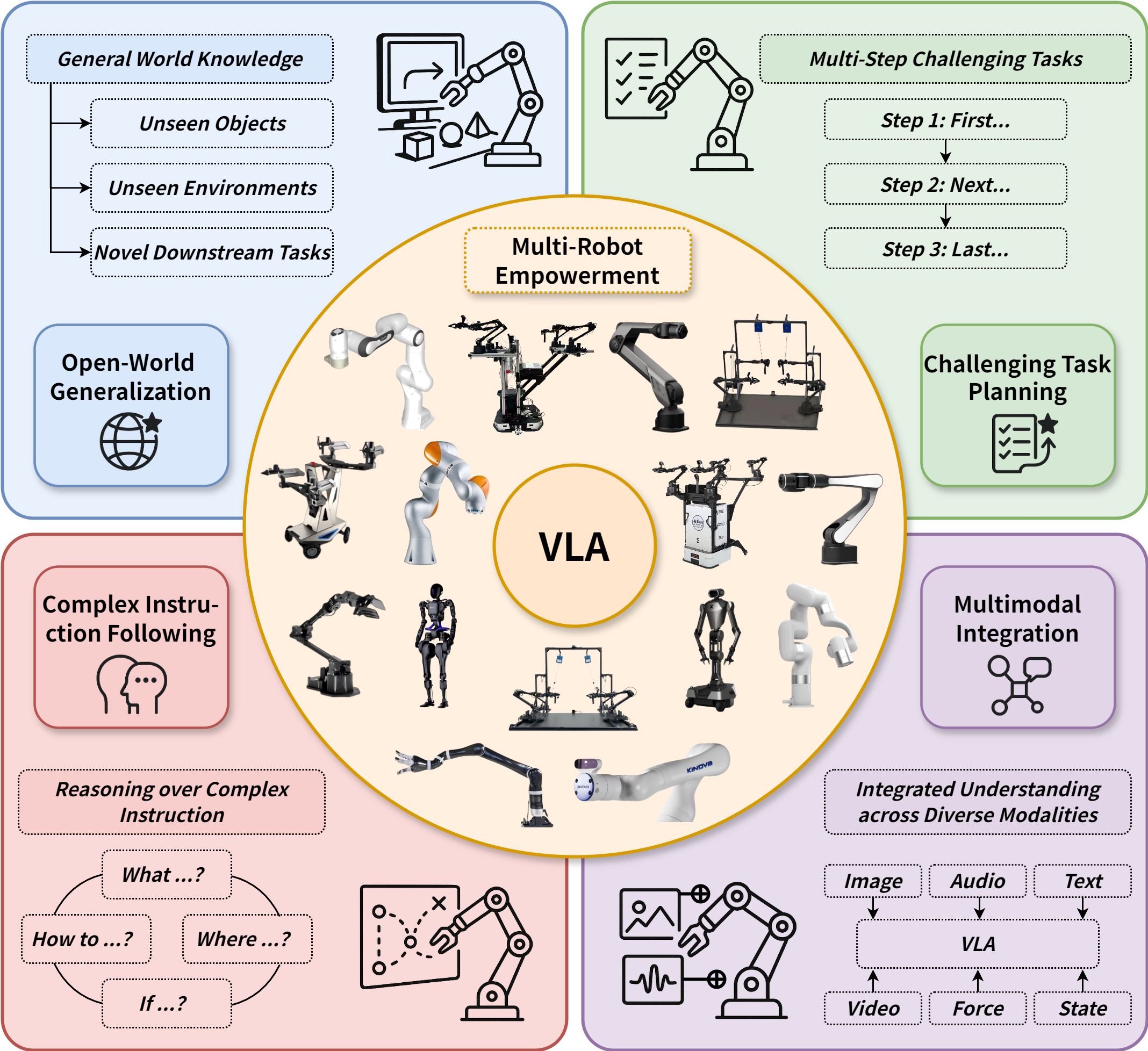}
    \caption{Illustration of core advantages of large VLM-based Vision-Language-Action (VLA) models for robotic manipulation. Large VLM-based VLA models leverages the strengths of large Vision-Language Models (VLMs), including \textbf{(1)} open-world generalization, \textbf{(2)} hierarchical task planning, \textbf{(3)} knowledge-augmented reasoning, and \textbf{(4)} rich multimodal fusion. 
    These capabilities empower diverse robotic arms and significantly enhance robotic intelligence.
}
    \label{fig:intro1}
\end{figure}

\section{Introduction}
\label{Introduction}
\IEEEPARstart{R}{obotic} manipulation stands as a pivotal challenge at the confluence of robotics and embodied artificial intelligence~\cite{intro1,intro2, intro3, intro4, intro5}. Its realization necessitates not only precise motor control but also a profound comprehension of diverse visual and semantic cues within complex, dynamic environments. The widespread utility of robotic manipulation is evident across numerous domains, including advanced manufacturing, efficient logistics, precision healthcare, and versatile domestic service~\cite{intro6, intro7, sapkota2025visionlanguageactionofactconceptsprogress}. Traditional manipulation methods~\cite{intro8, intro10, intro9, intro11, intro12, intro13, intro14} have been predominantly anchored in meticulously engineered control policies and rigidly predefined task specifications. However, these approaches demonstrably falter in unstructured, real-world settings—{particularly when confronted with novel objects~\cite{shridhar2022cliport, zeng2021transporter}, ambiguous natural language instructions~\cite{ahn2022can, huang2022inner}, or previously unseen environmental configurations~\cite{brohan2022rt, driess2023palme, intro8, kroemer2021review}}—thereby exposing inherent limitations in their scalability and generalization capabilities.


Recently, large Vision-Language Models (VLMs)~\cite{liu2024improved, dai2023instructblip, bai2023qwen, liu2023visual, chen2024expanding, li2024monkey} have emerged as a transformative paradigm. Pretrained on vast web-scale image-text datasets, large VLMs provide capabilities for aligning visual information with natural language. This innovative capacity enables large VLMs to interpret complex visual scenes in conjunction with textual descriptions, moving beyond mere object recognition to encompass a holistic contextual understanding. The subsequent integration of large VLMs into robotic systems has catalyzed the development of a novel class of models: large VLM-based Vision-Language-Action (VLA) models~\cite{kim2024openvla, brohan2023rt2visionlanguageactionmodelstransfer, belkhale2024rt, black2024pi0, pi052025phy, smolvla, bjorck2025gr00t}. 
As illustrated in Fig.~\ref{fig:intro1}, this emerging paradigm offers great potential to overcome the fundamental limitations of conventional robotic pipelines. It enables robots to interpret high-level human instructions, generalize to unseen objects and scenarios, reason about complex spatial relationships, and perform sophisticated manipulation tasks in dynamic, unstructured environments.
For instance, a VLA model could execute commands such as ``place the red mug next to the laptop onto the top shelf,'' a task demanding a sophisticated fusion of visual grounding, spatial reasoning, and sequential motion planning.

\begin{figure*}[t]
    \centering
    \includegraphics[width=1.0\linewidth]{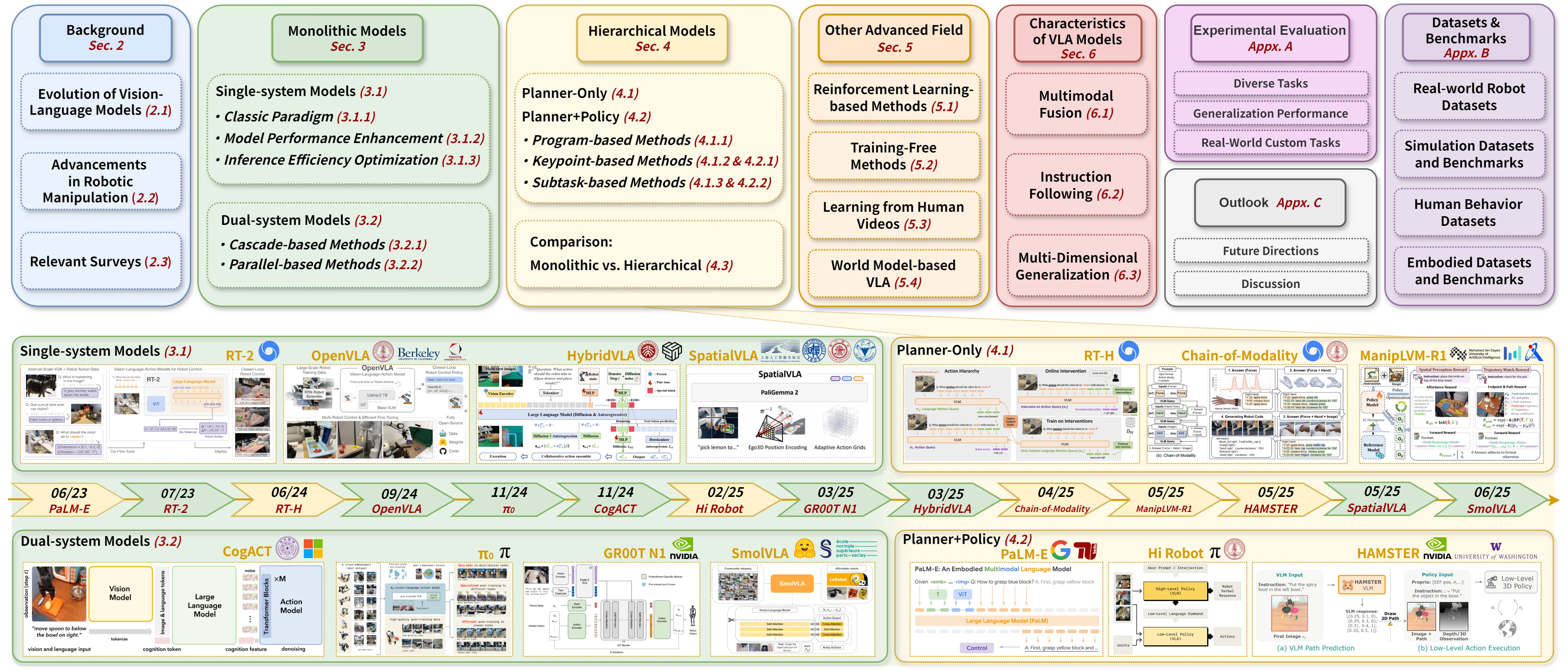}
    \caption{Outline of the organization of our comprehensive survey (top) and a chronological timeline of notable developments in large VLM-based Vision-Language-Action models for robotic manipulation (bottom). The timeline highlights representative milestones for both monolithic models and hierarchical models, providing a perspective on the field’s recent progress.}
    \label{fig:pipline & timeline}
\end{figure*}

\begin{figure*}[t]
    \centering
    \includegraphics[width=\linewidth]{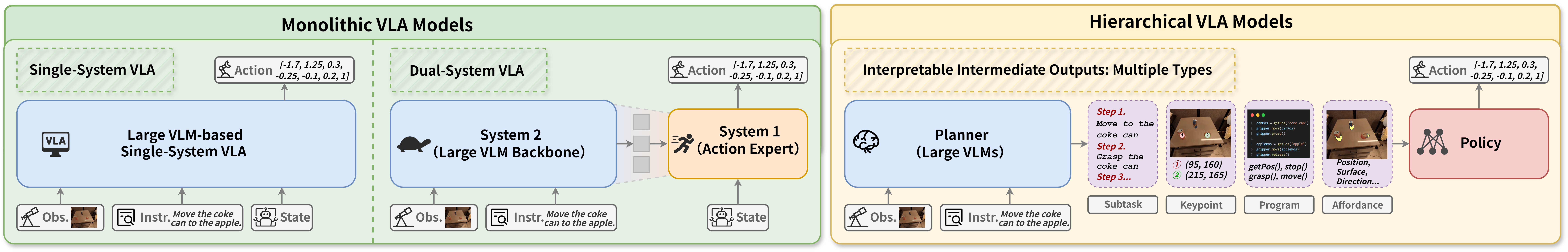}
    \caption{Comparison of the two principal categories of large VLM-based VLA models. Monolithic models (Sec.~\ref{monolithic}) integrate perception, language understanding, and action generation within single- or dual-system architectures, with the latter incorporating an additional action expert. In contrast, hierarchical models (Sec.~\ref{hi}) decouple planning from policy execution through interpretable intermediate outputs (e.g., subtasks, keypoints, programs, affordances).}
    \label{fig:thumbnail}
\end{figure*}

In this work, based on an extensive review of recent research~\cite{kim2024openvla, brohan2023rt2visionlanguageactionmodelstransfer, belkhale2024rt, black2024pi0, pi052025phy, smolvla, bjorck2025gr00t, cotVLA, hybridvla, li2025bridgevla, deervla, dan2025pi05ki} and a refined understanding of the field~\cite{cen2025worldvla, zhang2025rewind, chen2025fast, black2025real, zhou2025mitigatinghumanrobotdomaindiscrepancy, chen2025world4omnizeroshotframeworkimage}, we propose a coherent definition of large VLM-based VLA models: the model (1) leverages a large VLM to understand visual observations and natural language instructions, and (2) performs reasoning processes that directly or indirectly serve robotic action generation. We further distinguish two principal categories of large VLM-based VLA models, as shown in Fig.~\ref{fig:pipline & timeline} and Fig.~\ref{fig:thumbnail}:
\begin{itemize}
    \item \textbf{Monolithic Models} (Fig.~\ref{fig:thumbnail}, left) comprise single-system and dual-system implementations. (1) Single-system models~\cite{kim2024openvla, brohan2023rt2visionlanguageactionmodelstransfer, kim2025oft, hung2025nora} integrate both environmental comprehension (including visual perception, linguistic understanding, and robot state awareness) and action generation within a unified architecture. In contrast, (2) dual-system models~\cite{black2024pi0, pi052025phy, smolvla, bjorck2025gr00t} employ a VLM backbone for scene interpretation and an action expert for action determination, exchanging information via the propagation of latent representations.
    \item \textbf{Hierarchical Models} (Fig.~\ref{fig:thumbnail}, right)~\cite{song2025maniplvm, zhang2025embodied, li2025hamster, xu2025a0, huang2024rekep} explicitly decouple planning from policy execution. These differ from dual-system end-to-end approaches through two defining characteristics: (1) Structured intermediate outputs where planner modules generate interpretable representations such as keypoint detections, affordance maps, or trajectory proposals, which are then processed by policy modules to formulate executable actions; and (2) Decoupled training paradigms that enable independent optimization of hierarchical modules through specialized loss functions or API-mediated interactions.
\end{itemize}
This taxonomy emphasizes critical design dimensions in VLA development, particularly regarding system integration granularity and the explicitness of cognitive decomposition, while maintaining essential connections to modern representation learning paradigms.

Operating under the above definitions and taxonomy, our comprehensive survey across the broad spectrum of related work reveals several critical gaps within the nascent VLA field, with the organization of this survey illustrated in Fig.~\ref{fig:pipline & timeline}.
First, the terminology and modeling assumptions in this field remain inconsistent, and the research landscape is fragmented across disciplines (robotics, computer vision, natural language processing, etc.). 
Second, existing reviews tend to focus either on VLMs~\cite{intro22, intro15, intro19, intro20, intro21} or on robotic manipulation in isolation~\cite{intro16, intro2, intro17, intro18, intro23}, lacking a comprehensive synthesis of the unique challenges and advances that arise at their intersection. 
Therefore, there is a pressing need for a systematic and principled survey that elucidates the foundations of large VLM-based VLA models, organizes the space of relevant methods, and outlines future directions for this integrated paradigm. This survey aims to address this gap.
We present a structured and in-depth overview of advances in large VLM-based VLA models research, aiming to provide a panoramic perspective of the field to foster a deeper understanding and support future research.
Our key contributions are summarized as follows: 
\begin{itemize}
    \item \textbf{A Longitudinal Synthesis of large VLM-based VLA Models Development:} We systematically review the evolutionary trajectory of VLMs, the technical advancements in manipulation learning, and the subsequent emergence of the large VLM-based VLA paradigm. Furthermore, we examine the development of monolithic models and hierarchical models, identifying key challenges and outlining future directions. 
    \item \textbf{A Cross-Cutting Synthesis of large VLM-based VLA Modeling Practices:} We provide a more fine-grained comparative taxonomy of monolithic models and hierarchical models, examining them in detail from both structural and functional perspectives. We further explore advanced research frontiers of large VLM-based VLA models, highlighting their distinctive characteristics and the datasets that underpin their development.
\end{itemize}
This synthesis provides a high-level summary and a conceptual roadmap for understanding the field’s development and structural organization.

The remainder of this survey is organized as follows. As illustrated in Fig.~\ref{fig:pipline & timeline}, Sec.~\ref{background} presents essential background on the evolution of VLMs and the foundations of robotic manipulation. 
Sec.~\ref{monolithic} examines monolithic models, detailing single-system and dual-system architectures, their respective advantages, and design trade-offs. Sec.~\ref{hi} explores hierarchical models, categorizing them into planner-only and planner–policy frameworks, and further sub-classifying by intermediate representations such as subtasks, keypoints, and programs. Sec.~\ref{advanced} discusses other advanced fields, including RL-based optimization, training-free methods, learning from human videos, and world model-based approaches. Sec.~\ref{Characteristics} analyzes key characteristics of large VLM-based VLA models, covering multimodal fusion, instruction following, and multi-dimensional generalization. Sec.~\ref{conclusion} concludes the survey.
{Appx. A provides a comparative experimental analysis of state-of-the-art large VLM-based VLA models, evaluating generalization across benchmarks and real-world deployments, and discussing trade-offs between monolithic and hierarchical paradigms.}
{Appx. B} categorizes and analyzes the diverse datasets and benchmarks employed in large VLM-based VLA models research, spanning simulated, real-world, and human interaction data. 
Finally, {Appx. C} explores critical open challenges and promising future research directions.

\section{Background}
\label{background}
\subsection{Evolution of Vision-Language Models}
\label{Evolution of Vision-Language Models}
Recently, the emergence of large VLMss~\cite{chen2024lion, liu2024improved, dai2023instructblip, bai2023qwen, liu2023visual, chen2024expanding, li2024monkey, zhang2025falcon, shen2024mome} has been marked by a shift from task-specific architectures to unified frameworks capable of handling diverse multimodal tasks~\cite{li2025lion, li2025optimus, li2025optimus2, ye2024cat, ye2025cat+}. Modern VLMs, such as LLaVA1.5~\cite{liu2024improved} and Qwen-VL~\cite{bai2023qwen}, typically adopt a three-component architecture: a vision encoder to encode visual input, a projector to align visual features with textual embeddings, and a large language model to mediate multimodal reasoning. This design unifies multimodal tasks as textual outputs. 
{It enables VLMs to perform not only traditional tasks like visual question answering or object detection~\cite{alayrac2022flamingo, peng2023kosmos}, but also more advanced capabilities, including compositional reasoning and spatial navigation~\cite{ma2023crepe, hsieh2023sugarcrepe, shah2023lm, liu2025msnav}, which are critical for real-world applications.} 
Building on this foundation, DriveVLM~\cite{tiandrivevlm} demonstrates the integration of VLMs into autonomous driving systems, while CogAgent~\cite{hong2024cogagent} shows VLMs' ability to interact with graphical user interfaces. These advancements highlight the growing sophistication of large VLMs in bridging vision and language, while also revealing the emerging potential of using large VLMs to build broad real-world applications.

Crucially, the potential of large VLMs in real-world applications~\cite{chen2025less, lyu2025puma, zhu2025emosym, xie2025gui, shao2025robust, shao2024detecting, shao2023detecting, shao2019multi, shao2025deepfake, chen2025spabench} rests on their generalization capabilities afforded by visual instruction tuning.
By training on large-scale, carefully curated visual instruction–following datasets, VLMs acquire the flexibility to comprehend abstract or open-ended instructions and to generalize to unseen multimodal scenarios. LLaVA~\cite{liu2023visual} pioneers this paradigm by leveraging GPT-4 to convert raw image–text pairs into 150K conversational samples, significantly enhancing its ability to address open-ended queries. InternVL2~\cite{chen2024expanding} advances by introducing an LLM-guided automated filtering pipeline. It helps clean noisy annotations, reducing anomalous behaviors of the model. Collectively, these advances demonstrate how visual instruction tuning prepares large VLMs to follow high-level commands. This capability forms the foundation for extending large VLMs to VLA for robotic manipulation, which requires robust reasoning across vision, language, and action in the real world with ambiguous instructions.

Recent advancements in VLMs further emphasize scalability for enhanced multimodal perception and reasoning. Models like LLaVA-OneVision~\cite{li2025llavaonevision} unify image and video processing within a single framework. Qwen-2VL~\cite{bai2025qwen2} introduces dynamic resolution support for arbitrary input sizes, while Vision-R1~\cite{huang2025visionr1} leverages reinforcement learning to improve chain-of-thought reasoning. Collectively, these innovations empower VLMs with stronger real-world perception and decision-making capabilities, laying the groundwork for VLMs to further integrate action generation.

Building on these breakthroughs, the next frontier for large VLMs is to transition from passive observation to active interaction with physical environments. While models like DriveVLM~\cite{tiandrivevlm} and CogAgent~\cite{hong2024cogagent} already demonstrate VLMs' ability to process real-world inputs and generate high-level plans, they remain unable to engage directly with the real world in an embodied, physical manner. Equipping VLMs with robotic manipulation capabilities thus emerges as a particularly promising avenue for future research.

\subsection{Advancements in Robotic Manipulation}
\label{Advancements in Robotic Manipulation}


{Early language-conditioned robotic systems typically relied on modular designs, pairing separate vision encoders with language modules or planners rather than using unified multimodal models. For instance, CLIPort~\cite{shridhar2022cliport} combines the CLIP vision-language encoder~\cite{radford2021learning} for semantic grounding with a Transporter network~\cite{zeng2021transporter} for spatial reasoning, while RT-1~\cite{brohan2022rt} uses a CNN-based visual encoder~\cite{tan2019efficientnet} conditioned on separate language embeddings to map observations and instructions to actions. 
Beyond such pipelines, early vision-language manipulation methods like PerAct~\cite{shridhar2022peract} and VIMA~\cite{jiang2022vima} introduced transformer-based policies that jointly process visual inputs and language prompts for action prediction, while VLMBench~\cite{zheng2022vlmbench} provided a standardized benchmark for evaluating language-grounded manipulation. 
Although these approaches enable a degree of multi-task generalization by integrating pretrained visual and language components, they remain limited by task-specific supervision and lack strong generalization to novel concepts and long-horizon tasks. They also struggle with complex or ambiguous instructions, highlighting the need for unified large-scale VLM-based VLA architectures~\cite{kim2024openvla, black2024pi0, pi052025phy, bjorck2025gr00t, li2026semanticvla, li2026consisvla, li2026global, hu2026abstraction, zhu2026h, lv2025spatial, lv2025f1}.}


In contrast, the latest VLA models tightly integrate vision, language, and action control in a unified system. The RT-2~\cite{brohan2023rt2visionlanguageactionmodelstransfer} model exemplifies this shift. RT-2~\cite{brohan2023rt2visionlanguageactionmodelstransfer} begins with a pretrained large VLM backbone (e.g., PaLM-E~\cite{driess2023palme} or PaLI-X~\cite{chen2023pali}) and co-trains it on both internet-scale vision-language tasks and real robot trajectories. Crucially, robot actions are cast as text tokens and included in the same training corpus as normal language outputs. This simple scheme allows the model to ``absorb'' robot control as another language task, resulting in a unified VLA model. 
Empirically, RT-2 demonstrates improved semantic understanding. 
Compared to RT-1~\cite{brohan2022rt}, RT-2~\cite{brohan2023rt2visionlanguageactionmodelstransfer} generalizes to novel objects and unseen instructions (e.g., placing an object on a particular number) and can perform basic reasoning (choosing the smallest or closest object). In other words, the large-scale vision-language pretraining imbues the robotic policy with real-world knowledge and stronger language grounding that earlier models lack.

Following RT-2's demonstration of how Internet-scale VLM pretraining can boost robotic control, more VLA models have emerged. 
For example, $\pi_0$~\cite{black2024pi0} adopts a flow-matching architecture on top of a pretrained VLM, trained on diverse dexterous robot datasets, which yields strong zero-shot generalization and easy adaptation to new tasks via fine-tuning. In addition, OpenVLA~\cite{kim2024openvla} has been released as a 7B open-source VLA model pretrained on approximately 970k real-world robot demonstrations. It achieves superior generalist manipulation performance and supports efficient fine-tuning on consumer hardware via techniques like LoRA~\cite{hu2022lora}. Together, these advancements underscore a trend toward greater generality and open accessibility in VLA-driven robotic manipulation.


\subsection{Relevant Surveys {and Taxonomies}}
\label{Relevant Surveys}

The growing interest in VLA models for embodied AI has inspired several surveys~\cite{intro1, intro2, intro3, intro4, intro5, intro6, sapkota2025visionlanguageactionofactconceptsprogress, wang2024large, sun2025review}, yet most focus on broader architectural paradigms or diverse application domains, leaving a gap in the systematic exploration of large VLM-based VLA systems specifically tailored for robotic manipulation. 
{Crucially, the taxonomies in existing surveys fail to highlight the fundamental impact of pre-trained large models as core backbones on architectural design.}

For example, 
Wang et al.~\cite{wang2024large} offers an early overview of the integration of text-only LLMs into robotic task planning. It emphasizes their capability to generate precise action plans from natural language instructions. However, the study primarily focuses on high-level reasoning and does not address the challenge of grounding VLMs in robotic visual perception and action determination, which is a gap that recent VLA models are designed to fill. {To bridge this gap, our taxonomy introduces both monolithic and hierarchical paradigms to systematically analyze how high-level semantic plans are translated into actuator commands and how VLMs are securely grounded in low-level control mechanisms.}
Ma et al.~\cite{intro2} conduct a comprehensive survey of VLA architectures, reviewing modular, end-to-end, and hybrid approaches for integrating vision, language, and action modalities.
However, this survey lacks a focused analysis of recent trends that leverage pre-trained VLMs as foundational components. {To address this limitation, our Monolithic and Hierarchical classifications categorize models based on how the VLM backbone is structurally integrated into the control loop. Specifically, we detail whether the VLM acts as a unified action generator or a decoupled high-level planner, as shown in Fig.~\ref{fig:thumbnail}.}
Sapkota et al.~\cite{sapkota2025visionlanguageactionofactconceptsprogress} provide an extensive investigation into a broader range of VLA applications, such as autonomous driving, augmented reality navigation, and precision agriculture. However, its breadth dilutes the depth required for robotics-specific challenges such as real-time actuation constraints, sensor noise robustness, and long-term decision-making. 
{Yao et al.~\cite{yao2023bridging} categorize language-conditioned manipulation tasks based on the functional roles of language, such as serving as policy conditioning or enabling cognitive planning. Li et al.~\cite{li2024foundation} examine how various foundation models are incorporated into the high-level functional modules of robot learning. While these surveys provide task-oriented or modular perspectives, they typically treat foundation models as auxiliary cognitive tools rather than unified action generators. In contrast, this survey specifically investigates how large VLMs are structurally coupled with low-level control mechanisms.}

To address the lack of a comprehensive and in-depth survey in this emerging field, we provide a structured overview of recent VLA research. Building on the definition of VLA models described in Sec.~\ref{Introduction}, 
we trace the evolution of VLAs in robotics and analyze key learning paradigms. {Meanwhile, we provide comparative introductions and insights into advanced architectures leveraging reinforcement learning, world models, and related techniques.} Our aim is to offer a comprehensive perspective that supports deeper understanding and drives future advances.




\section{Monolithic Models}
\label{monolithic}
Monolithic VLA models are mainly implemented in two ways: single-system and dual-system architectures, as illustrated in Fig.~\ref{fig:thumbnail} left. 
In the single-system design (Sec.~\ref{single}), visual perception, language instructions, and robot states are jointly fed into a unified model that processes all modalities and decodes executable actions via autoregressive or parallel decoding.
In contrast, dual-system architectures (Sec.~\ref{dual}) divide functionality into two cooperating modules: System 2 (VLM backbone) performs slower but more generalized reflective reasoning, while System~1 (action expert) focuses on fast processing to support reactive behaviors.
{
Single-system VLAs offer streamlined development and avoid complex inter-module communication. In contrast, dual-system VLAs are highly inclusive and cover a broad range of action expert integration strategies, including hybrid variants subsumed under this category, leveraging a division of labor to combine reactive speed with reasoning accuracy.
}

\subsection{Single-system Models}
\label{single}
Single-system VLA models embody the monolithic design philosophy. They aim to transfer the semantic knowledge of large VLMs to robotic manipulation tasks through a unified model. This section reviews the paradigms of single-system models and explores two main research directions: enhancing the model’s capabilities to solve complex tasks and improving inference efficiency for practical deployment. All models are summarized in Tab. \ref{single-system}.

\begin{table*}[!ht]
\scriptsize
\renewcommand{\arraystretch}{1.2}
\caption{Single-system VLA models. In the LLM / VLM column, omission of the V-Encoder indicates a VLM; otherwise, it represents an LLM. In the Learning column, ``AD'' denotes Autoregressive Decoding and ``PD'' denotes Parallel Decoding. ``SFT'' denotes fine-tuning distinct from action-prediction imitation learning, where tasks like captioning, VQA, reasoning and others all qualify as SFT. ``A'' and ``B'' in parentheses represent the learning methods used by Action head or Backbone.}
\label{single-system}
\begin{tabularx}{\textwidth}{C{2.2cm} C{2.3cm} C{2.1cm} C{2.1cm} L X}
    \toprule
    \makecell[c]{\textbf{Model}}
    & \makecell[c]{\textbf{V-Encoder}} 
    & \makecell[c]{\textbf{LLM / VLM}}
    & \makecell[c]{\textbf{Learning}} 
    & \makecell[c]{\textbf{Contribution}}  \\ 


    \midrule \rowcolor{lightgray} \multicolumn{5}{c}{\textbf{Classic Paradigm: Autoregressive Decoding}} \\ \midrule

    RT-2
    \cite{brohan2023rt2visionlanguageactionmodelstransfer}
    & -
    & PaLI-X / PaLM-E
    & AD (A), SFT (B)
    & Represent actions as VLM tokens to enable generalization.\\

    \rowcolor{lightgray}
    RT-2-X
    \cite{o2024open}
    & ViT-22B
    & UL2
    & AD (A), SFT (B)
    & Fine-tune on cross-robot data for positive skill transfer.\\

    OpenVLA 
    \cite{kim2024openvla}
    & DINOv2 + SigLIP
    & LLaMA2-7B 
    & AD (A)
    & Open-source 7B-parameter VLA model for generalist robot control.\\

     \midrule \rowcolor{lightgray} \multicolumn{5}{c}{\textbf{Paradigm Derivations: Model Performance Enhancement}} \\ \midrule

    LEO agent
    \cite{huang2024embodied}
    & ConvNext
    & Vicuna-7B
    & AD (A), SFT (B)
    & Combine object-centric 3D features with LLM for action.\\

    \rowcolor{lightgray}
    ECoT \cite{chen2025trainingstrategiesefficientembodied}
    & DINOv2 + SigLIP
    & LLaMA2-7B
    & AD (A), SFT (B)
    & Incorporate chain-of-thought to enhance policy explainability.\\
    
    ReVLA
    \cite{dey2024revla}
    & DINOv2 + SigLIP
    & LLaMA2-7B
    & AD (A)
    & Reverse backbone gradually to preserve visual generalization.\\

    \rowcolor{lightgray}
    TraceVLA
    \cite{zheng2024tracevla}
    & DINOv2 + SigLIP
    & LLaMA2-7B
    & AD (A)
    & Propose visual trace prompting for spatiotemporal awareness.\\

    FuSe \cite{jones2025beyond}
    & -
    & PaliGemma-3B
    & AD (A), SFT (B)
    & Leverage natural language for cross-modal fine-tuning. \\

    \rowcolor{lightgray}
    UniAct
    \cite{zheng2025universal}
    & -
    & LLaVA-0.5B
    & PD (A)
    & Propose universal action space for versatile and adaptive control.\\

    SpatialVLA \cite{qu2025spatialvlaexploringspatialrepresentations}
    & SigLIP
    & Gemma2
    & AD (A)
    & Improve generalization via 3D encoding and action grid.\\

    \rowcolor{lightgray}
    UP-VLA
    \cite{zhang2025up}
    & ViT + VQ-GAN
    & Phi1.5-1.3B
    & AD (A), SFT (B) 
    & Propose unified training for semantic-spatial understanding.\\

    VLAS \cite{zhao2025vlas}
    & CLIP
    & Vicuna-7B
    & AD (A), SFT (B)
    & Introduce voice modality to VLA and construct a paired dataset.\\

    \rowcolor{lightgray}
    HybridVLA \cite{hybridvla}
    & DINOv2 + SigLIP
    & LLaMA2-7B
    & Diff., AD (A)
    & Integrate diffusion and autoregressive policies to improve success. \\

    CoT-VLA \cite{cotVLA}
    & -
    & VILA-U 
    & AD+PD(A), SFT(B)
    & Propose a visual chain-of-thought to improve planning.\\

    \rowcolor{lightgray}
    VTLA \cite{zhang2025vtla}
    & -
    & Qwen2-VL-7B
    & AD (A)
    & Integrate visual and tactile inputs to improve task success.\\

    OE-VLA \cite{zhao2025unveilingpotentialvisionlanguageactionmodels}
    & SigLIP
    & Qwen1.5-7B
    & AD (A), SFT (B)
    & Introduce four open-ended tasks to expand interaction modalities.\\

    \rowcolor{lightgray}
    ReFineVLA \cite{vanvo2025refinevlareasoningawareteacherguidedtransfer}
    & SigLIP
    & Gemma2
    & AD (A), SFT (B)
    & Propose reasoning-aware framework to fine-tune VLAs effectively.\\

    LoHoVLA \cite{yang2025lohovlaunifiedvisionlanguageactionmodel}
    & SigLIP
    & Gemma-2B
    & AD (A), SFT (B)
    & Address long-horizon tasks via hierarchical closed-loop control.\\

    \rowcolor{lightgray}
    BridgeVLA
    \cite{li2025bridgevla}
    & SigLIP
    & Gemma
    & PD (A), SFT (B)
    & Project 3D data into 2D space for efficient action prediction\\

    UnifiedVLA \cite{wang2025unified}
    & -
    & Emu3
    & AD (A), SFT (B)
    & Convert all input signals into tokens to build a unified model.\\

    \rowcolor{lightgray}
    WorldVLA
    \cite{cen2025worldvla}
    & -
    & Chameleon
    & AD (A), SFT (B)
    & Combine world and action models for bidirectional improvement.\\

    4D-VLA \cite{zhang20254dvla}
    & -
    & InternVL-4B
    & PD (A)
    & Integrate 4D spatiotemporal cues for efficient VLA pretraining. \\

    \rowcolor{lightgray}
    VOTE
    \cite{lin2025vote}
    & DINOv2 + SigLIP
    & LLaMA2-7B
    & PD (A)
    & Introduce voting strategy to increase action prediction accuracy.\\

    ST-VLA \cite{patratskiy2025stvla}
    & -
    & PaliGemma2
    & AD (A), SFT (B)
    & Project visual traces onto depth maps for better understanding. \\

    \midrule \rowcolor{lightgray} \multicolumn{5}{c}{\textbf{Paradigm Derivations: Inference Efficiency Optimization}} \\ \midrule
    
    RoboFlamingo
    \cite{li2023vision}
    & ViT
    & MPT-1B
    & PD (A), SFT (B)
    & Decouple design to adapt open-sourced VLM for robotic control.\\

    \rowcolor{lightgray}
    RoboMamba
    \cite{liu2024robomamba}
    & CLIP / SigLIP ViT-L
    & Mamba-2.8B/1.4B
    & PD (A), SFT (B)
    & Introduce the Mamba architecture to the VLA field.\\

    DeeR-VLA \cite{deervla}
    & CLIP ViT-L/14
    & MPT-1B / 7B
    & PD (A)
    & Propose dynamic early-exit to reduce inference overhead.\\

    \rowcolor{lightgray}
    OpenVLA-OFT \cite{kim2025oft}
    & DINOv2 + SigLIP
    & LLaMA2-7B
    & PD (A)
    & Boost performance via OpenVLA-based fine-tuning.\\

    PD-VLA \cite{pdvla}
    & CLIP ViT-L 
    & Vicuna1.5-7B
    & PD (A)
    & Introduce parallel decoding manner for faster robot control. \\

    \rowcolor{lightgray}
    MoLe-VLA
    \cite{zhang2025mole}
    & DINOv2 + SigLIP
    & LLaMA2-7B 
    & PD (A)
    & Reduce computation via dynamic LLM layer activation.\\

    NORA \cite{hung2025nora}
    & -
    & Qwen2.5-VL-3B
    & AD (A)
    & Build efficient low-parameter model to boost performance.\\

    \rowcolor{lightgray}
    FLashVLA \cite{tan2025thinktwiceactonce}
    & DINOv2 + SigLIP
    & LLaMA
    & AD (A)
    & Propose retraining-free acceleration to improve VLA inference.\\

    BitVLA \cite{wang2025bitvla}
    & SigLIP b1.58
    & BitNet b1.58 2B4T
    & PD (A), SFT (B)
    & Build ternary weight model to reduce deployment memory cost.\\

    \rowcolor{lightgray}
    Spec-VLA \cite{wang2025spec}
    & DINOv2 + SigLIP
    & LLaMA2-7B
    & PD (A)
    & Propose speculative decoding to speed up without success drop.\\

    CogVLA \cite{li2025cogvla}
    & DINOv2 + SigLIP
    & LLaMA2-7B
    & PD (A)
    & Propose a cognition-aligned framework for efficient manipulation.\\

    \bottomrule
\end{tabularx}
\end{table*}

\subsubsection{\textbf{Classic Paradigm: Autoregressive Decoding}}
\label{sec:Classic Paradigm: Autoregressive Decoding}
\begin{figure*}[t]
    \centering
    \includegraphics[width=\linewidth]{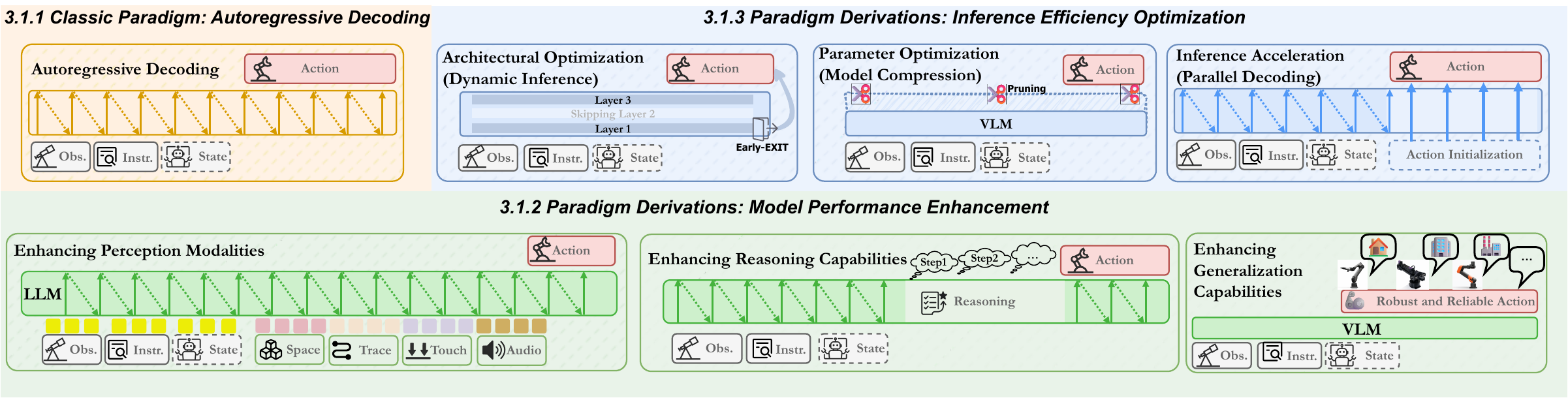}
    \caption{Comparison of representative paradigms in single-system models. Sec. \ref{sec:Classic Paradigm: Autoregressive Decoding} illustrates the model schematic of classical autoregressive decoding, where arrows denote information flow. Sec. \ref{sig:performance} presents approaches to enhance model capabilities through additional modalities, chain-of-thought reasoning, and improved generalization. Sec. \ref{sig:inference} discusses methods for improving inference efficiency through architectural refinement, parameter design, and decoding strategies.}
    \label{fig:single}
\end{figure*}

The autoregressive decoding paradigm draws directly from the sequence generation capability of LLMs. By discretizing the robot’s continuous action space into a token sequence, the model can sequentially predict action tokens. As shown in ``Autoregressive Decoding'' of Fig. \ref{fig:single}, the VLM receives the visual observation, natural language instruction, and optionally the robot state as input. Then it autoregressively generates action tokens, which can be converted into executable actions through a downstream de-tokenizer.


The RT series \cite{brohan2022rt, brohan2023rt2visionlanguageactionmodelstransfer, o2024open} and OpenVLA \cite{kim2024openvla} are typical examples of this paradigm. RT-1 \cite{brohan2022rt} introduces Transformer \cite{vaswani2017attention} into the VLA domain and encodes continuous robot actions as discrete tokens. RT-2 \cite{brohan2023rt2visionlanguageactionmodelstransfer} inherits this token-based action representation and applies it to a larger-scale VLM. It co-fine-tunes on internet-scale vision-language and robot trajectory data, successfully transferring rich visual–semantic knowledge from web data to robotic manipulation tasks. RT-2-X \cite{o2024open} further improves cross-robot skill transfer by applying co-fine-tuning on the Open X-Embodiment (OXE) dataset \cite{o2024open}.
Going a step further, OpenVLA \cite{kim2024openvla} replaces the large-parameter vision encoder in RT series with a combination of SigLIP \cite{zhai2023sigmoid} and DINOv2 \cite{oquab2023dinov2}. Through fine-tuning on large-scale real-world robotic manipulation data, it achieves superior performance with fewer model parameters. Its fully open-sourced nature positions it as a widely used baseline in subsequent research.

{These models demonstrate that large-scale vision-language pretraining knowledge can be transferred directly to robot control. However, the classic paradigm relies heavily on a single 2D visual input and reactive decision-making. As a result, they often perform poorly in complex real-world environments due to limited capabilities and low efficiency, setting the stage for subsequent research on performance enhancement and inference efficiency optimization.}

\subsubsection{\textbf{Model Performance Enhancement}}
\label{sig:performance}

{To address these performance bottlenecks in complex environments, recent studies have systematically enhanced VLA models from three perspectives. They expand perception modalities to capture richer physical surroundings, strengthen reasoning capabilities to transition from reactive control to deliberate decision-making, and improve generalization capabilities to ensure robust execution.}

\textbf{Enhancing Perception Modalities}. 
As shown in Fig.~\ref{fig:single}, these models improve real-world information acquisition by expanding input modalities. 
\textbf{(1) 3D perception.} Since the real world is three-dimensional and continuous, large VLM-based VLA models cannot fully understand it by relying solely on 2D images. Accordingly, Leo Agent \cite{huang2024embodied} directly obtains object-centric 3D point clouds from simulation environments and encodes them using PointNet++ \cite{qi2017pointnet++}; the encoded features are then fed into a spatial transformer \cite{chen2022language} to generate ``Space'' information. In real-world settings, SpatialVLA \cite{qu2025spatialvlaexploringspatialrepresentations} proposes Ego3D position encoding, which estimates depth from 2D images and back-projects it into egocentric 3D coordinates, fusing pixel-wise 3D positions with 2D features to produce egocentric ``Space'' information. Similarly, BridgeVLA \cite{li2025bridgevla} reconstructs 3D point clouds from RGB-D images and generates three orthographic projection views, enabling ``Space'' information to be represented in a 2D format compatible with large VLMs.

\textbf{(2) 4D perception.} To better capture spatial structure and temporal dynamics in robotic manipulation, 4D perception is required for more comprehensive scene understanding. TraceVLA \cite{zheng2024tracevla} overlays sampled motion-point trajectories on the current image to create the ``Trace'' information in Fig.~\ref{fig:single}. Combined with the original observation, it equips the large VLM with spatiotemporal understanding. Furthermore, 4D-VLA \cite{zhang20254dvla} integrates 3D coordinates into visual features to resolve spatial coordinate inconsistency, and leverages memory bank sampling of historical keyframes to mitigate temporal state ambiguity, thereby improving spatiotemporal reasoning. ST-VLA \cite{patratskiy2025stvla} introduces the ``Spatial Traces'' approach, which combines historical temporal ``Trace'' information with spatial depth information of the scene, surpassing prior SpatialVLA \cite{qu2025spatialvlaexploringspatialrepresentations} and TraceVLA \cite{zheng2024tracevla} models in certain robotic manipulation tasks.

\textbf{{(3) Additional modality perception.}} {Purely vision-based models often fail in contact-rich operations. To reduce the perceptual gap with the real world during manipulation, VTLA \cite{zhang2025vtla} encodes tactile information with a vision encoder to align it with visual sequences, and then feeds it into the LLM together with visual and textual tokens, thereby constructing a ``Vision-Tactile-Language-Action model''. Beyond tactile information, auditory cues in the real world also deserve attention. VLAS \cite{zhao2025vlas} employs the ``Whisper Encoder'' to extract speech features and projects them into the LLM embedding space through an MLP, thereby introducing the audio information shown in Fig.~\ref{fig:single}. Moreover, Fuse \cite{jones2025beyond} learns to align ``Touch'' and ``Audio'' modalities with language concepts during fine-tuning, enabling multimodal fusion with limited modality specific annotations. Beyond touch and audio, OE-VLA \cite{zhao2025unveilingpotentialvisionlanguageactionmodels} incorporates multimodal instructions by leveraging diverse multimodal data and a two-stage curriculum learning strategy, thereby extending instructions beyond text to images and videos.}

\textbf{Enhancing Reasoning Capabilities}. To move VLA models from simple reactive control toward more advanced deliberative decision-making, enhancing reasoning ability is essential. As shown in ``Enhancing Reasoning Capabilities'' of Fig. \ref{fig:single}, the LLM generates a chain-of-thought ``Reasoning'' process, which is then used as contextual information to produce the final ``Action''. 
ECoT \cite{zawalski2025robotic} generates a reasoning chain that sequentially combines high-level task planning with visually grounded features before outputting the final action. 
CoT-VLA \cite{cotVLA} further introduces a visual chain-of-thought reasoning by predicting a subgoal observation that represents a planned state in pixel space. 
In contrast to reasoning-oriented perspectives, LoHoVLA \cite{yang2025lohovlaunifiedvisionlanguageactionmodel} employs a ``Hierarchical Closed-Loop Control'' mechanism to address planning errors, action failures, and external disturbances, thereby handling long-horizon tasks. 
ReFineVLA \cite{vanvo2025refinevlareasoningawareteacherguidedtransfer} adopts a ``Selective Transfer Fine-Tuning'' strategy with dual learning objectives to fine-tune only the upper layers, enabling the model to enhance multimodal understanding.
These studies collectively underscore the central role of reasoning in enabling VLA models to achieve more reliable and generalizable action prediction.


\textbf{Enhancing Generalization Capabilities}. 
The generalization capability of a model refers to its ability to perform diverse tasks across different platforms and scenarios. 
As illustrated in ``Enhancing Generalization Capabilities'' of Fig. \ref{fig:single}, UniAct \cite{zheng2025universal} defines a ``Universal Action Codebook'' that abstracts heterogeneous robot actions into a unified representation. This unified encoding eliminates action-space heterogeneity and enables cross-embodiment learning and reasoning, thereby enhancing generalization.
ReVLA \cite{dey2024revla} employs a reversible training strategy that gradually restores the vision encoder to its original pre-trained state, mitigating catastrophic forgetting during fine-tuning and improving out-of-distribution visual generalization.

Another approach to enhancing generalization is to generate more robust and reliable actions, as depicted in Fig. \ref{fig:single}. 
HybridVLA \cite{hybridvla} integrates diffusion and autoregressive decoding within a unified model, and employs a Collaborative Action Ensemble mechanism to adaptively fuse them, selecting the most suitable generation strategy for different tasks, thereby improving control robustness.
Similarly, VOTE \cite{lin2025vote} introduces an adaptive ``Ensemble Voting'' strategy that groups past action predictions by similarity to the current one. It averages the majority set to produce more robust actions, balancing responsiveness and stability.

Enabling the model to better understand physical dynamics can promote the generation of more reliable actions. WorldVLA \cite{cen2025worldvla} integrates an action model and a world model within a unified autoregressive framework for capturing physical dynamics. Likewise, UnifiedVLA \cite{wang2025unified} introduces a world model as a post-training task, enabling it to learn causal dynamics of the physical world from large-scale unlabeled videos. UP-VLA \cite{zhang2025up} proposes a training framework that implicitly learns the physical dynamics of the world by predicting the next-frame image during pre-training. Consequently, these models generate more reliable actions that better adhere to physical laws. 

{Overall, the core strategies for improving the performance of single-system models include introducing new modalities and incorporating techniques such as chain-of-thought reasoning. The former injects diverse prior knowledge about the physical world into the model and compensates for the blind spots of 2D vision, whereas the latter endows the model with more human-like complex reasoning. Together with other advances, these strategies ultimately lead to broader generalization capabilities. However, although these methods substantially improve model capability, incorporating new modalities still relies on task-specific fine-tuning data and has not yet developed general pre-trained representations comparable to those for 2D image-text data. Nevertheless, integrating these capabilities inevitably exacerbates the computational burden, highlighting a critical trade-off between performance enhancement and the real-time constraints of robotic control.}

\subsubsection{\textbf{Inference Efficiency Optimization}}
\label{sig:inference}
{To resolve the slow inference speeds that conflict with high-frequency robotic control requirements, recent studies have systematically optimized VLA efficiency from three perspectives. They refine model architectures to enable dynamic and lightweight computation, reduce overall parameter sizes to lower deployment barriers, and accelerate decoding strategies to generate actions more rapidly.}

\textbf{Architectural Optimization}. These works aim to improve efficiency by adjusting the VLM architecture in VLA. As illustrated in ``Architectural Optimization'' of Fig. \ref{fig:single}, dynamic inference can be achieved through layer skipping or early exiting, avoiding redundant computation. 
MoLe-VLA \cite{zhang2025mole} introduces a ``Spatio-Temporal Aware Router (STAR)'' over the joint projection space of visual and language inputs, dynamically computing routing weights to select the most relevant LLM layers. This mechanism efficiently activates critical layers while skipping redundant ones, enabling fast and semantically sensitive inference.
DeeR-VLA \cite{deervla} enables dynamic inference by adding multiple early exits within the LLM. Adjacent exits compare output consistency to determine whether inference can be terminated early, avoiding unnecessary deep-layer computation.
Unlike dynamic inference, RoboMamba~\cite{liu2024robomamba} uses a
linear-complexity Mamba backbone \cite{gu2023mamba} and pose head, reporting 9.0~Hz on A100 versus 3.4~Hz for OpenVLA-7B~\cite{kim2024openvla}. However, this $ \sim 3\times$ gain over OpenVLA is based on an SE(3) pose-prediction interface and does not imply universal superiority over generalist autoregressive VLAs.

\textbf{Parameter Optimization}. Reducing model size directly lowers deployment difficulty. As shown in Fig. \ref{fig:single} under ``Parameter Optimization'', model compression directly reduces the parameter count. BitVLA \cite{wang2025bitvla} introduces the first ``1-bit'' weight VLA model, using distillation-aware training to quantize the vision encoder and adopting the ``1-bit'' weight BitNet b1.58 \cite{nielsen2024bitnet} as its LLM. On the other hand, NORA \cite{hung2025nora} pairs a compact, high-quality VLM with a FAST+ \cite{pertsch2025fast} tokenizer that compresses high-dimensional actions into short token sequences. It demonstrates that a well-designed small model can outperform larger models.

\textbf{Inference Acceleration}. This line of work seeks to enhance inference efficiency by accelerating the process of action decoding during inference. The most notable aspect is that most models replace autoregressive decoding with parallel decoding. Fig. \ref{fig:single} under ``Inference Acceleration'' shows that decoding speed can be improved by modifying the LLM decoding process or using a specialized action head. In this way, a complete action sequence can be generated in a single forward pass. RoboFlamingo \cite{li2023vision} separates perception and policy, using an independent MLP action head to convert VLM outputs into actions. This enables the generation of stacked actions in one pass. OpenVLA-OFT \cite{kim2025oft} replaces the causal attention mask with a bidirectional one, enabling empty action embeddings to be filled in one pass and converted to continuous actions via an MLP. PD-VLA \cite{pdvla} reformulates autoregressive decoding as a nonlinear equation system solved via fixed-point iteration. Bidirectional attention updates action tokens until convergence, thereby enabling parallel decoding.

Different from parallel decoding, Spec-VLA \cite{wang2025spec} explores speculative decoding in VLA models. With Relaxed Acceptance to increase the verification model's acceptance rate for draft model predictions, it achieves a 1.42$\times$ speedup over the baseline. Beyond decoding, FlashVLA \cite{tan2025thinktwiceactonce} addresses redundancy by using FlashTrigger to evaluate the stability of actions and the environment and decide whether to skip the current inference, significantly reducing latency.

{In summary, current optimization methods significantly improve the efficiency of model inference and deployment by compressing parameter sizes, pruning redundant computations, and parallelizing the decoding process. While they successfully bridge the gap between heavy foundation models and real-time robotic constraints, these efficiency gains often come at the expense of reasoning depth and task accuracy. Aggressive quantization and pruning strategies can degrade the model's capability to handle complex tasks. Furthermore, most approaches merely adapt standard LLM optimization techniques rather than addressing the unique spatial-temporal nature of embodied control. Moving forward, true efficiency will require a shift from simple structural compression to embodiment-native designs. Developing dual-system paradigms that decouple high-level deliberate reasoning from high-frequency instinctive motor control offers a promising solution to these challenges. Ultimately, the field must seek an optimal equilibrium that preserves the emergent intelligence of massive models within the ultra-low latency required by the physical world.}

\vspace{-6pt}
\subsection{Dual-system Models}
\label{dual}

\begin{figure}[t]
    \centering
    \includegraphics[width=\linewidth]{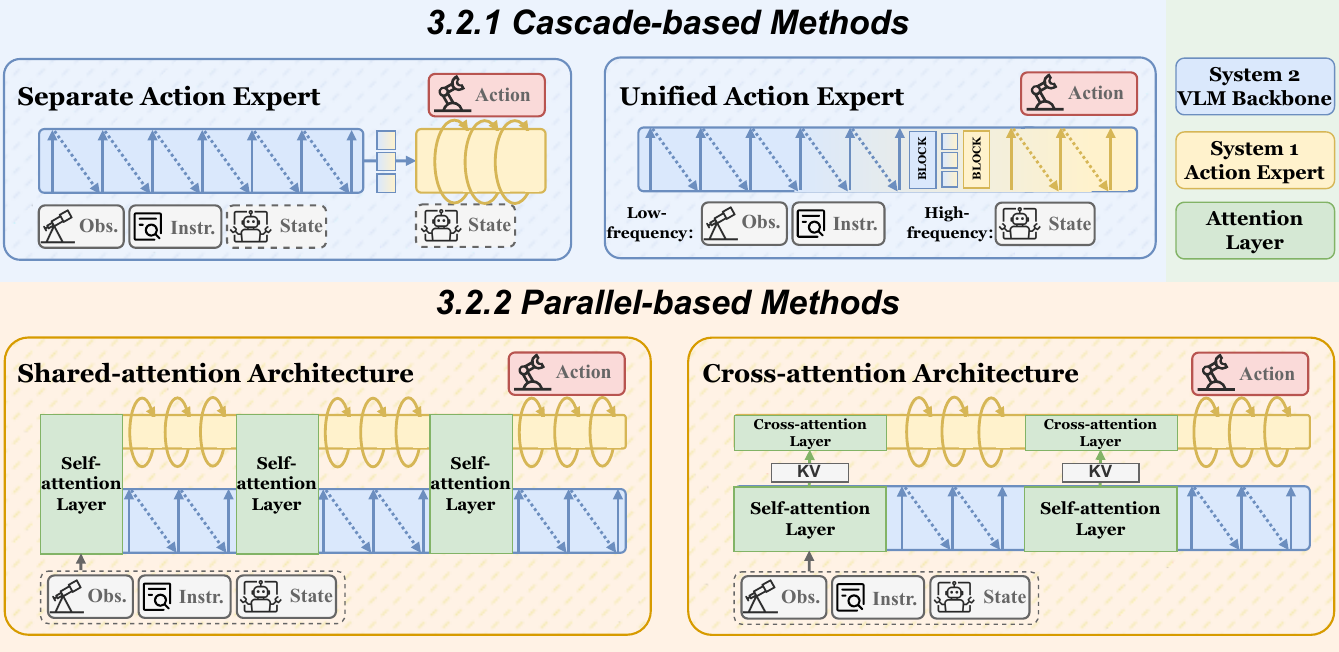}
    \caption{Comparison of representative paradigms in monolithic dual-system models. Sec.~\ref{sec:Cascade-based Methods} introduces cascade-based methods, where the VLM backbone is cascaded with the action expert to forward its output features in a single pass to the action expert. Sec.~\ref{sec:Parallel-based Methods} introduces parallel-based methods, where the VLM backbone and the action expert operate in parallel and interact in various ways.}
    \label{fig:dual}
\end{figure}

The need to simultaneously handle the demands of deep reasoning and real-time action generation has prompted the development of monolithic dual-system VLA architectures. 
These models adopt two distinct systems: a larger and slower System 2 for high-level reasoning and planning, and a smaller and faster System 1 for low-level action generation. This separation enables a faster response time and smoother real-world deployment. It delivers both solid strategic planning and high-speed control for robotic manipulation. The separated low-level control module is also more adaptable to task-specific optimization. The key distinction between this approach and single-system models lies in the introduction of an action expert to explicitly decouple action generation. In contrast to hierarchical models, its major difference is that it does not produce an interpretable intermediate output. Tab. \ref{tab:dual} summarizes representative dual-system methods. Fig. \ref{fig:dual} illustrates two representative architectures: cascade-based methods and parallel-based methods. {This categorization is grounded in how System 2 interacts with System 1. Cascade-based designs follow a sequential interaction pattern, where high-level planning precedes low-level execution, while parallel-based designs adopt a tightly coupled paradigm with continuous interaction between the two systems during inference. This distinction leads to different trade-offs in information flow, system coupling, and real-time responsiveness.}

\subsubsection{\textbf{Cascade-based Methods}}
\label{sec:Cascade-based Methods}

{Following the above categorization, we first examine cascade-based methods, which adopt a sequential interaction pattern between the two systems. In this paradigm, high-level reasoning is performed prior to low-level action execution, resulting in a clear temporal separation between planning and control.}

As shown in the upper part of Fig. \ref{fig:dual}, cascade-based dual-system VLA models separate high-level semantic reasoning from low-level real-time control in a serial manner. The high-level system (System 2) typically employs VLMs to process multimodal input, perform semantic grounding, and generate action plans. These plans are encoded as latent cognitive representations rather than being executed directly. The low-level system (System 1) then decodes these representations into executable robot actions at a higher frequency, significantly improving run-time efficiency and facilitating real-world deployment. Fig. \ref{fig:dual} shows two architectures, differing in how the action expert is designed.

\begin{table*}[t]
\scriptsize
\renewcommand{\arraystretch}{1.2}
\caption{Dual-system VLA models. The ``System 2 Backbone'' column lists the VLM backbone used as the System 2 component in dual-system methods. The ``System 1 Learning'' column lists the learning methods used by the action experts as System 1. ``Diff.'' denotes diffusion-based learning, “FM” denotes flow-matching, ``MSE'' denotes mean squared error, ``BCE'' denotes binary cross-entropy, and ``AR'' denotes autoregressive learning.}
\label{tab:dual}
\begin{tabularx}{\textwidth}{C{2.6cm} C{1.4cm} C{1.3cm} L X}
    \toprule
    \textbf{Model} & \textbf{System 2 Backbone} & \textbf{System 1 Learning} & \makecell[c]{\textbf{Contribution}} \\ \midrule
    

    \rowcolor{lightgray}
    \multicolumn{4}{c}{\textbf{Cascade-based}} \\ \midrule

    DP-VLA \cite{han2024dual}
    & OpenVLA
    & Regression
    & Propose a dual-system architecture for robot manipulation with efficiency and performance. \\

    \rowcolor{lightgray}
    RoboDual \cite{bu2024towards}
    & OpenVLA
    & Diff.
    & Combine a VLA-based generalist for reasoning and a DIT specialist for control. \\

    LCB 
    \cite{shentu2024llms}
    & LLaVA
    & Diff.
    & Leverage an added special token to encode VLM reasoning and act as conditions for policy. \\

    \rowcolor{lightgray}
    GR00T N1 \cite{bjorck2025gr00t}
    & Eagle-2
    & FM
    & Combine a VLM and DiT for humanoid robots manipulation. \\

    CogACT \cite{li2024cogact}
    & OpenVLA
    & Diff.
    & {\scriptsize Propose an action ensemble algorithm to integrate the action diffusion process into VLA. } \\
    
    \rowcolor{lightgray}
    HiRT 
    \cite{zhang2024hirt}
    & InstructBLIP
    & Regression
    & Propose a dual-system model with System 2 running at a lower frequency.  \\

    Fast-in-Slow \cite{chen2025fast}
    & Prismatic
    & Diff., AR
    & Propose a unified dual-system model that embeds fast execution within a VLM-based reasoner.  \\

    \rowcolor{lightgray}
    OpenHelix \cite{intro18}
    & LLaVA
    & Diff.
    & Conduct auxiliary training on the token bridging VLM and policy. \\

    ChatVLA \cite{chatvla}
    & Qwen2-VL
    & Diff.
    & Unifie vision-language-action via MoE-shared attention with separate perception/control FFNs. \\

    \rowcolor{lightgray}
    ChatVLA-2 \cite{chatvla2}
    & Qwen2-VL
    & Diff.
    & Enable open-world robotic reasoning via dynamic MoE routing and Reasoning-Following MLP. \\

    Diffusion-VLA \cite{diffusionvla}
    & Qwen2-VL 
    & Diff.
    & Merge Qwen2-VL reasoning with diffusion actions via FiLM-modulated reasoning injection. \\

    


    \rowcolor{lightgray}
    TriVLA \cite{liu2025trivla}
    & Eagle-2
    & Diff.
    & Introduce a world-dynamics perception module as system 3 to complement static perception. \\

    GF-VLA \cite{li2025information}
    & LLaMA 2
    & Regression
    & Enable interpretable bimanual manipulation via information-theoretic graphs from human videos. \\

    \rowcolor{lightgray}
    RationalVLA \cite{song2025rationalvla}
    & LLaVA-v1.5
    & Diff.
    & Introduce a learnable latent interface to enable instruction rejection for robust manipulation. \\

    VQ-VLA \cite{wang2025vq}
    & OpenVLA
    & VQ-VAE
    & Develop a vector quantization-based action tokenizer for efficient and smoother control. \\

    \rowcolor{lightgray}
    TinyVLA
    \cite{wen2025tinyvla}
    & LLaVA
    & Diff.
    & Demonstrate that high-performance VLAs require no large-scale robotic pretraining.\\
    
    \midrule \rowcolor{lightgray} \multicolumn{4}{c}{\textbf{Parallel-based}} \\ \midrule
    
    $\pi_0$
    \cite{black2024pi0}
    & PaliGemma
    & FM
    & Combine a pre-trained Vision-Language Model with a Flow Matching-based Action Expert.\\

    \rowcolor{lightgray}
    $\pi_0$-FAST
    \cite{pertsch2025fast}
    & $\pi_0$
    & AR
    & Propose a DCT-based action tokenization enabling efficient autoregressive VLA training.\\

    $\pi_{0.5}$
    \cite{pi052025phy}
    & PaliGemma
    & FM
    & Convert high-level prompts into more fine-grained subtask predictions before feeding into $\pi_0$ \\
    
    \rowcolor{lightgray}
    $\pi_{0.5}$-KI
    \cite{dan2025pi05ki}
    & PaliGemma
    & FM
    & Prevent gradients from the action expert from flowing into the VLM backbone during training.\\


    ForceVLA
    \cite{yu2025forcevla}
    & $\pi_0$
    & Diff.
    & Treat force sensing as a first-class modality via MoE, improving contact-rich manipulation. \\

    \rowcolor{lightgray}
    SmolVLA  \cite{smolvla}
    & SmolVLM-2
    & FM
    & Propose a lightweight VLA with frozen SmolVLM-2 and flow-matching transformer. \\

    OneTwoVLA \cite{lin2025onetwovla}
    & $\pi_0$
    & FM
    & Integrate acting/reasoning in shared VLA backbone processing multi-view inputs. \\

    \rowcolor{lightgray}
    Tactile-VLA \cite{huang2025tactile}
    & $\pi_0$
    & FM
    & Integrate tactile sensing to enable force-aware, generalizable contact-rich manipulation. \\

    GR-3 \cite{cheang2025gr3}
    & Qwen2.5-VL
    & FM
    & Combine VL data and few-shot trajectories for robust manipulation in long-horizon or unseen tasks. \\

    \rowcolor{lightgray}
    villa-X \cite{chen2025villa}
    & PaliGemma
    & FM
    & Integrate proprioceptively grounded latent actions and robot actions in a joint diffusion process. \\

    GraspVLA \cite{deng25gra}
    & InternLM2
    & FM
    & Enable sim-to-real and open-vocabulary grasping with synthetic data. \\
    
    \bottomrule
\end{tabularx}
\end{table*}


To decouple the capabilities of ``cognition'' and ``action'' of the model, many methods select a separate model to serve as the action expert. As shown in the ``Separate Action Expert'' part at the top-left of Fig. \ref{fig:dual}, the VLM backbone transmits features to the action expert to convey information derived from visual, textual, and robot-specific state inputs. For example, CogACT \cite{li2024cogact} introduces the diffusion transformer (DiT) \cite{peebles2023dit} as an action model. In particular, it also proposes an adaptive action ensemble algorithm for smoother and more efficient movement trajectories. GR00T N1 \cite{bjorck2025gr00t} also adopts a dual-system architecture in which DiT is introduced as the low-level action model. Both systems are tightly coupled and jointly trained end-to-end, enabling much faster policy steps. Some models adopt a similar architecture, but do not employ a DiT. For example, DP-VLA \cite{han2024dual} uses a Behavioral Cloning transformer \cite{mandlekar2021matters} instead.

Many models have made variations based on this. The HiRT model \cite{zhang2024hirt} uses VLM to operate at a lower frequency. This process extracts features for long-term understanding of the scene. Then a lightweight visual action strategy is deployed at a higher frequency. This design enables more efficient robotic manipulation and allows the system to keep up with changes in the real-world environment during deployment. TriVLA \cite{liu2025trivla} is a three-module system incorporating a video generation model to predict future frames, a VLM to interpret instructions, and a diffusion action expert. GF-VLA \cite{li2025information} extracts information–based hand–object and object–object scene graphs from human demonstrations, then fuses them with an LLM to generate interpretable behavior trees and low-level Cartesian actions for dual-arm control. RationalVLA \cite{song2025rationalvla} couples a VLM with a diffusion policy through a learnable latent interface. The VLM emits a token serving as the condition of the controller for action generation or a token to refuse infeasible commands. VQ-VLA \cite{wang2025vq} adopts a convolutional residual VQ-VAE \cite{van2017VQVAE}, which is pretrained with action sequence, to take the place of the binning method of OpenVLA \cite{kim2024openvla}. The model shows linear performance gains from more simulated data and has less sim-to-real gap. Some models integrate the action expert into the VLA backbone, Fast-in-Slow \cite{chen2025fast} is a typical example. As shown in the ``Unified Action Expert'' part at the top-right of Fig. \ref{fig:dual}, the action expert utilizes the final transformer blocks of the VLM backbone. They run at different frequencies, enabling seamless coordination between the two systems within a single pretrained model.

{In cascaded VLA models, the action expert serves as the core component. Based on the underlying learning paradigm, they can be broadly categorized into four classes: regression, discretization, diffusion, and flow matching. Each class adopts distinct architectures: \textbf{(1)} regression-based methods typically employ a Transformer encoder with an MLP head to directly predict continuous actions; \textbf{(2)} discretization-based methods leverage VQ-VAE~\cite{van2017VQVAE} to compress continuous actions into discrete codebooks; \textbf{(3)} diffusion-based methods utilize 1D U-Net~\cite{ronneberger2015u} or DiT~\cite{peebles2023dit} to model complex action distributions via iterative denoising; and \textbf{(4)} flow matching methods generally build vector field networks upon DiT, enabling efficient generation through ordinary differential equation trajectories. These paradigms reflect fundamental trade-offs among architectural simplicity, expressive power for complex action distributions, computational efficiency, and quantization error.}

{Overall, cascade-based designs emphasize modular decomposition between cognition and action, while differing primarily in the implementation of the action expert, setting the stage for more tightly coupled interaction paradigms explored in parallel-based methods. However, this explicit separation also introduces inherent limitations: the lack of real-time feedback from the action system can reduce adaptability in dynamic environments, and errors in high-level planning may propagate to execution without timely correction. As a result, although cascade-based methods offer strong interpretability and stable optimization, they often face challenges in responsiveness and robustness under rapidly changing conditions.}

\subsubsection{\textbf{Parallel-based Methods}}
\label{sec:Parallel-based Methods}

{In contrast to cascade-based designs, parallel-based methods adopt a tightly coupled interaction pattern between high-level reasoning and low-level control. Instead of enforcing a strict temporal separation, the two systems operate simultaneously and continuously exchange information during inference.}

As shown in the lower part of Fig. \ref{fig:dual}, parallel-based dual-system VLA designs an action expert operating in parallel with the VLM backbone. The two components interact to exchange information during inference. Based on the choice of action expert and the interaction mechanism, two categories in Fig. \ref{fig:dual} are defined.

As shown in the ``Shared-attention Architecture'' part at the bottom-left of Fig. \ref{fig:dual}, this architecture draws inspiration from the mix-of-experts (MoE) framework \cite{noam2017outr}. They leverage token interactions within self-attention layers, which are shared by the VLM backbone and the action expert to separate high-level reasoning from low-level execution. Information derived from visual and textual inputs interacts with noise and robot-specific inputs within the shared self-attention layers. This design facilitates task-specific optimization. For example, this design allows for the incorporation of a true MoE or other specially designed architectures within the action expert. A typical example is $\pi_0$ \cite{black2024pi0}. Its backbone weights are initialized from a pre-trained VLM. To handle robot-specific inputs and action generation, a second set of independent weights, the flow-matching-based action expert, is introduced and trained from scratch. ForceVLA\cite{yu2025forcevla} uses $\pi_0$ \cite{black2024pi0} as the base model, the FVLMoE module with MoE is used to introduce the force modality into VLA. OneTwoVLA \cite{lin2025onetwovla} is based on $\pi_0$ \cite{black2024pi0} and can switch between two modes: explicitly reasoning and generating actions based on the most recent reasoning. This architecture makes it easier for the two systems to operate asynchronously and further improves efficiency.

Many innovations are built upon this. $\pi_{0.5}$ \cite{pi052025phy} builds on $\pi_0$ \cite{black2024pi0} by introducing an additional step. The VLA module integrates visual information to convert high-level prompts into more fine-grained subtask predictions, which are then processed by the dual-system architecture of $\pi_0$ \cite{black2024pi0}. $\pi_{0.5}$-KI \cite{dan2025pi05ki} builds on $\pi_{0.5}$ \cite{pi052025phy}. During training, it prevents gradients from the action expert from flowing into the VLM backbone during training to preserve the VLM’s knowledge advantage. $\pi_0$-FAST \cite{pertsch2025fast} introduces a DCT-driven action tokenization approach that facilitates efficient autoregressive training of VLA models. GraspVLA\cite{deng25gra} proposes a dual-system model via Progressive Action Generation (PAG), which unifies autoregressive perception tasks and flow-matching-based action generation into a Chain-of-Thought process. This design enables joint training on synthetic and Internet data, achieving direct sim-to-real transfer and open-vocabulary grasping. villa-X \cite{chen2025villa} grounds latent actions in robot states and jointly models latent and robot actions via joint diffusion for structured vision-action integration. Tactile-VLA \cite{huang2025tactile} integrates tactile sensing into VLA models with hybrid position–force control. As shown in the ``Cross-attention Architecture'' part at the bottom-right of Fig. \ref{fig:dual}, this architecture feeds the visual, textual, and state inputs into the VLM, where the self-attention layers generate key–value and pass key–value to the action expert’s cross-attention layers. For instance, SmolVLA\cite{smolvla} adopts this architecture. It improves efficiency by leveraging a lightweight and frozen VLM backbone and training only a downstream Flow Matching Transformer as the action expert. Another example is GR-3 \cite{cheang2025gr3}, which unifies vision-language understanding with robot trajectory learning.

{Overall, parallel-based designs emphasize tight coupling and continuous interaction between cognition and action, enabling real-time adaptability and responsive behavior in dynamic environments. By allowing bidirectional information flow during inference, these methods can better adjust actions based on evolving observations and intermediate reasoning states. However, this strong interdependence also introduces challenges: the increased system coupling complicates optimization and training stability, and the lack of clear modular boundaries can reduce interpretability and flexibility for component-level improvements. As a result, while parallel-based methods excel in responsiveness and adaptability, they often face difficulties in scalability and maintainability compared to cascade-based designs.}

\begin{figure*}[t]
    \centering
    \includegraphics[width=1.0\linewidth]{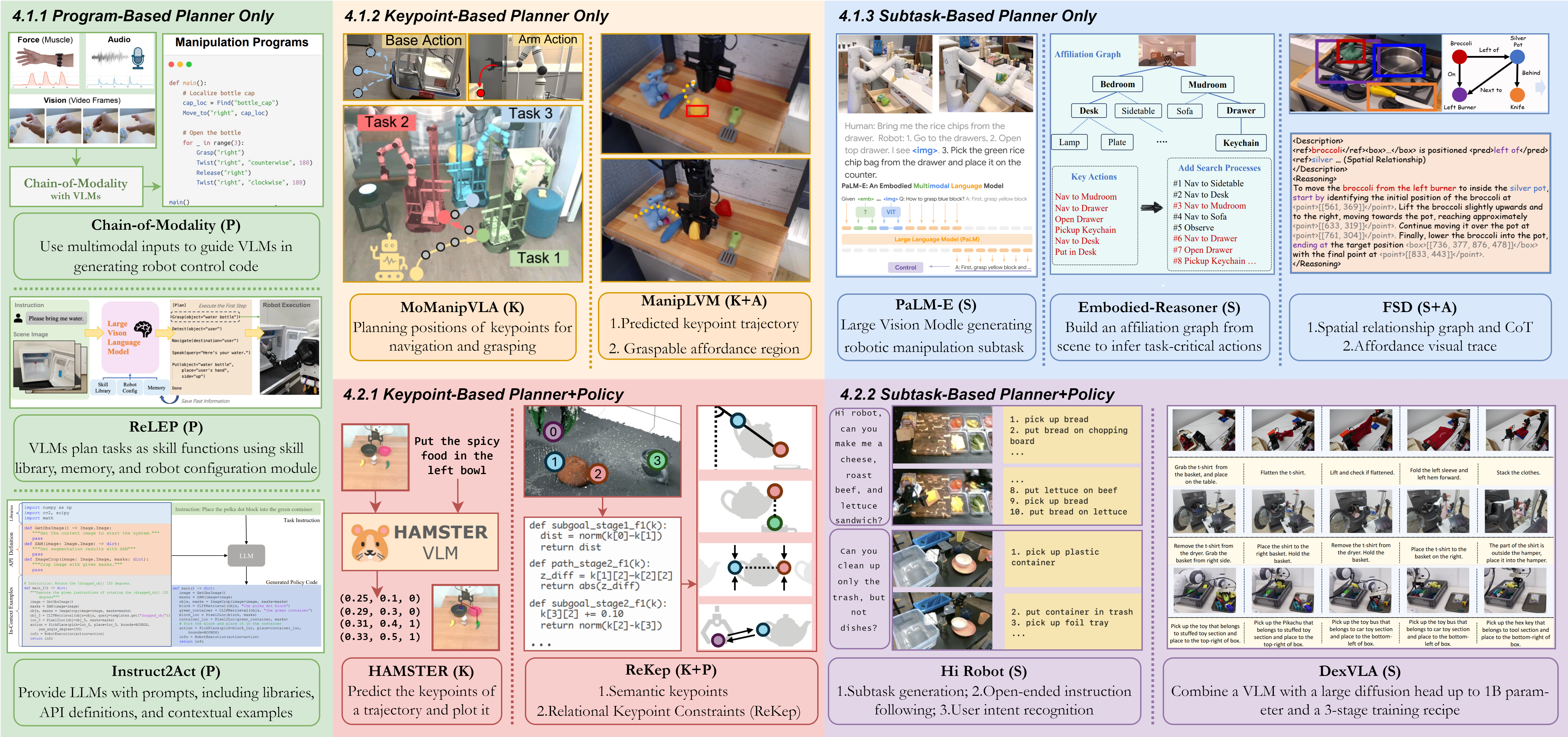}
    \caption{A diagram showing the hierarchical models in this survey. These models are divided into two categories according to their constitutions, i.e., Planner-Only (Sec.~\ref{sec:hie-planner-only}) and Planner+Policy (Sec.~\ref{sec:hie-planner-policy}). Based on the intermediate representation type, one category can be further divided into subtask-based (S), keypoint-based (K), and program-based (P). ``(A)'' denotes methods combining affordance for auxiliary purposes.}
    \label{fig:midvis}
\end{figure*}

\section{Hierarchical Models}

\label{hi}


Hierarchical modelling serves as a foundational paradigm in large VLM-based VLA models, particularly in scenarios where long-horizon reasoning, spatial abstraction, or action decomposition is required. These models are typically composed of a high-level planner and a low-level policy, as illustrated in Fig.~\ref{fig:thumbnail} right. The planner receives instructions and observations, transforming them into interpretable intermediate representations. The policy then accepts these representations and generates action sequences or codes that robots can directly execute. A summary of hierarchical models is provided in Tab. \ref{hierarchical}.

Importantly, the planner and policy in hierarchical models can operate independently, without either module being strictly dependent on the other. This modularity enables flexible combinations: many works focus solely on designing the planner component, leveraging existing off-the-shelf policies for execution. Therefore, we divide hierarchical models into two categories: Planner-Only (Sec.~\ref{sec:hie-planner-only}) and Planner+Policy (Sec.~\ref{sec:hie-planner-policy}). Unlike dual-system VLA models, which also involve multiple modules, the intermediate representations in hierarchical models are explicitly interpretable to humans. Based on their nature, each category can be further divided into subtask-, keypoint-, and program-based methods. A comparative analysis of monolithic and hierarchical models is presented in Sec.~\ref{sec:hie-comp}.


\begin{table*}[htbp]
\scriptsize
\renewcommand{\arraystretch}{1.2}
\caption{Hierarchical VLA models. The ``Type'' column denotes the output type of the planner, where ``K'' represents Keypoint, ``S'' represents Subtask, and ``P'' represents Program. The ``Learning'' column specifies the learning method adopted by the model, where ``SFT'' refers to Supervised Fine-Tuning, ``RL'' denotes Reinforcement Learning, ``IM'' indicates Imitation Learning, and ``API'' is a special case referring to the invocation of pre-existing models.}
\label{hierarchical}
\begin{tabularx}{\textwidth}{C{2.5cm} C{0.6cm} C{2.4cm} C{1.3cm} X}
    \toprule
    \makecell[c]{\textbf{Model} \\[0.4ex]}
    & \makecell[c]{\textbf{Type} \\ [0.4ex]} 
    & \makecell[c]{\textbf{Backbone} \\ [0.4ex]} 
    & \makecell[c]{\textbf{Learning} \\[0.4ex]} 
    & \makecell[c]{\textbf{Contribution} \\[0.4ex]} \\ \midrule
    \rowcolor{lightgray}
    \multicolumn{5}{c}{\textbf{Planner-Only}} \\ \midrule

    MoManipVLA \cite{momanipvla}
    & K
    & OpenVLA-7B 
    & IM
    & Leverage VLA models to predict waypoints and optimize full-body trajectories. \\
    
    \rowcolor{lightgray}
    ManipLVM-R1 \cite{song2025maniplvm}
    & K
    & Qwen2.5-VL-3B
    & RL
    & GRPO tuning for affordance and trajectory, robust performance in OOD situations. \\

    PaLM-E \cite{driess2023palme}
    & S
    & PaLM
    & SFT
    & Train a VLM capable of general VQA and robot manipulation instruction generation. \\

    \rowcolor{lightgray}
    {Emb-Reas}\cite{zhang2025embodied}
    & S
    & Qwen2-VL-7B
    & SFT
    & Construct a VLM and open-sourced dataset with planning, reasoning, and reflection.\\

    RoboPoint\cite{yuan2024robopoint}
    & K
    & Vicuna-v1.5-13B
    & SFT
    & Finetune VLM for spatial affordance prediction in the form of keypoints.\\

    \rowcolor{lightgray}
    Reinforced \cite{wu2025reinforced}
    & S
    & Qwen2.5-VL-7B
    & SFT, RL
    & Conduct GRPO on a finetuned model, bringing better generalization to unseen.\\

    CoM~\cite{wang2025chain}
    & P
    & Gemini 1.5 Pro 
    & API
    & Sequential multimodal prompting to extract force-aware manipulation from demos. \\

    \rowcolor{lightgray}
    RoVI \cite{li2025robotic}
    & K, P
    & GPT-4o / LLaVA-13B
    & SFT
    & Visual sketch-based instruction and hierarchical pipeline for precise manipulation. \\

    ReLEP \cite{liu2025longhorizonembodiedplanningimplicit}
    & P
    & LLaVA-1.6-7B
    & SFT
    & A planning framework with implicit logical inference and hallucination mitigation. \\

    \rowcolor{lightgray}
    ViLa \cite{hu2023lookleapunveilingpower}
    & S
    & GPT-4V
    & API
    & A VLM planner integrating perception and reasoning without affordance models. \\

    RoboBrain \cite{ji2025robobrain}
    & K, S
    & LLaVA
    & SFT
    & Provide a hierarchical VLA focus on planning, affordance, and trajectory.\\
    
    \midrule \rowcolor{lightgray} \multicolumn{5}{c}{\textbf{Planner+Policy}} \\ \midrule

    HAMSTER \cite{li2025hamster}
    & K
    & VILA-1.5-13B
    & SFT, IM
    & Propose an out-of-the-box way for trajectory prediction to assist the low-level policy. \\

    \rowcolor{lightgray}
    HiRobot \cite{shi2025hi}
    & S
    & PaliGemma-3B, $\pi_0$
    & SFT, IM 
    & A hierarchical VLA with high explainability and capacity for complex tasks. \\ 

    Agentic Robot \cite{yang2025agentic}
    & S
    & GPT-4o
    & SFT 
    & A closed-loop hierarchical pipeline where a VLM is attached to a completion judge. \\

    
    \rowcolor{lightgray}
    DexVLA \cite{wen2025dexvla}
    & S
    & Qwen2-VL
    & SFT, IM
    & Combine a VLM with a large diffusion head up to 1B and a 3-stage training recipe. \\

    Instruct2Act \cite{huang2023instruct2act}
    & P
    & ChatGPT 
    & API
    & Generate programs that call APIs for mapping from instructions to actions. \\
    
    \rowcolor{lightgray}
    RoboMatrix \cite{mao2024robomatrix}
    & S
    & Vicuna 1.5
    & SFT
    & VLA with modular scheduling layer, skill layer, and hardware layer. \\

    PointVLA \cite{li2025pointvla}
    & S
    & Qwen2-VL
    & SFT
    & Attach VLA with a point cloud encoder and injector to equip spatial perception. \\

    \rowcolor{lightgray}
    $A_0$ \cite{xu2025a0}
    & K
    & Qwen2.5-7B
    & SFT, API
    & Hierarchical affordance-aware diffusion with embody-agnostic keypoint prediction. \\

    FSD \cite{yuan2025seeing}
    & S
    & CLIP, Vicuna
    & SFT
    & Propose the generation of visual aids via SrCoT for zero-shot manipulation. \\

    \rowcolor{lightgray}
    RoBridge \cite{zhang2025robridge}
    & S
    & GPT-4o
    & IM, RL
    & Bridge VLM cognition with RL execution via invariant operable representation. \\

    Robocerebra \cite{han2025robocerebralargescalebenchmarklonghorizon}
    & S
    & GPT-4o, Qwen2.5-VL
    & SFT
    & A novel benchmark and hierarchical framework for long-horizontal evaluation. \\


    \rowcolor{lightgray}
    DexGraspVLA
    \cite{zhong2025dexgraspvla}
    & S
    & Qwen-VL
    & API, IM
    & Combine a VLM as a high-level planner with a low-level diffusion-based policy.\\

    RT-H
    \cite{belkhale2024rt}
    & S
    & PaLI-X 55B
    & SFT, IM
    & An action hierarchy architecture using language motion as a middle representation.\\

    \rowcolor{lightgray}
    ReKep \cite{huang2024rekep}
    & K, P
    & GPT-4o
    & API
    & Training-free trajectory generation by keypoint constraints for manipulation. \\

    VoxPoser
    \cite{huang2023voxposer}
    & P
    & GPT-4
    & API
    & Propose language-guided 3D value maps with zero-shot generalization capabilities.\\

    \rowcolor{lightgray}
    SkillDiffuser \cite{liang2024skilldiffuser}
    & S
    & Transformer 
    & IM
    & A hierarchical VLA with high-level model and low-level model.\\

    RT-Affordance \cite{nasiriany2024rtaffordanceaffordancesversatileintermediate}
    & K
    & PaLM-E 2
    & SFT, IM
    & Use visual affordances as intermediate features for web-robot knowledge transfer. \\

    \rowcolor{lightgray}
    HiBerNAC \cite{wu2025hibernac}
    & S
    & PaLM2 
    & SFT
    & Propose an asynchronous model that mimics the hierarchical structure of the brain\\

    LLARVA  
    \cite{niu2024llarva}
    & K
    & Llama 2 7B
    & SFT, IM
    & A vision-action instruction tuning paradigm and a large instruction tuning dataset.\\

    \rowcolor{lightgray}
    MALMM
    \cite{singh2024malmm}
    & S, P
    & GPT-4-Turbo
    & API
    & Three-agent system combining planner, supervisor, and coder without post-training. \\


    VLA-Touch \cite{bi2025vlatouch}
    & S
    & GPT-4o
    & IM
    & Integrate tactile sensing into VLA control via diffusion-based imitation learning. \\
    
    \bottomrule
\end{tabularx}
\end{table*}
\subsection{Planner-Only}
\label{sec:hie-planner-only}
\subsubsection{\textbf{Program-based Methods}} 

{Among the intermediate representations, programs are uniquely characterized by their capacity to encapsulate complex logic, conditional control flows, and precise parameterizations into highly structured text. By leveraging the inherent code-generation proficiencies of large foundational models, this category focuses on translating high-level task goals into strict algorithmic sequences.} In this approach, planners generate intermediate programs for robotic manipulation, which fall into two categories: robot-executable programs and auxiliary programs. Robot-executable programs are built on robot libraries and can be directly executed to control the robot. For instance, Chain-of-Modality \cite{wang2025chain} employs a multimodal prompting strategy, where the VLM is engaged in a multi-turn conversation across different modalities and ultimately generates a robot-executable Python program to reproduce the task. Similarly, Instruct2Act \cite{huang2023instruct2act} produces Python code that invokes APIs to control robot actions. In contrast, auxiliary programs support the policy in task understanding but cannot be executed directly. ROVI \cite{li2025robotic} exemplifies this category by generating auxiliary programs to describe the potential action and resolve the actual execution through translational and rotational costs. Likewise, ReLEP \cite{liu2025longhorizonembodiedplanningimplicit} uses a VLM with a memory bank to decompose tasks into basic skills from a skill library. It produces plans in the form of auxiliary programs that enable strong long-horizon performance. 

{In summary, program-based methods excel in orchestrating complex, multi-step tasks by abstracting manipulation into semantic logic and API calls. While this algorithmic abstraction is highly effective for tasks with clear procedural structures, other scenarios demand precise, continuous geometric reasoning directly within the visual observation. This alternate focus on spatial grounding characterizes keypoint-based methods, as discussed next.}

\subsubsection{\textbf{Keypoint-based Methods}}

{While program-based methods emphasize logical control flow, keypoint-based planners shift the focus toward precise spatial reasoning directly within the visual observation. More broadly, we use ``keypoint'' as a general term for grounded intermediate subgoals that specify where or how to interact. These representations extend beyond sparse points to include continuous waypoints, visual sketches, and even non-visual coordinate sequences. Low-level controllers then consume these spatial descriptors to generate continuous actions. Under this broad definition, we categorize existing methods by the complexity of their spatial representations, ranging from discrete waypoints to rich visual sketches. At the foundational level,} several methods tackle robotic manipulation by predicting {explicit} waypoints. MoManipVLA \cite{momanipvla} generates a key waypoint at each step through a VLA model, which is subsequently refined into executable actions via a bi-level trajectory optimization framework. Other approaches emphasize affordance-driven keypoints {to capture task-relevant semantics}. RoboPoint \cite{yuan2024robopoint} interprets natural language instructions to generate visual keypoints that specify precise manipulation targets. {Building on this, recent models often form hybrid representations.} ManipLVM-R1 \cite{song2025maniplvm} trains a VLM with Group Relative Policy Optimization (GRPO) \cite{guo2025deepseek} to predict both the affordance area for grasping and the trajectory of the target object, thereby yielding a more generalizable planner. Similarly, RoboBrain \cite{ji2025robobrain} integrates task planning, affordance perception, and trajectory estimation by combining LLaVA \cite{liu2023visual} for high-level planning with an A-LoRA module to identify interactable regions and a T-LoRA module to predict trajectory waypoints. Moreover, some works explore {richer forms of spatial guidance, such as trajectories and visual sketches}. RoVI \cite{li2025robotic}, for instance, employs a sketch-based interface, extracting starting points, waypoints, and endpoints from hand-drawn annotations using YOLOv8, which are then applied as trajectory constraints.

{Collectively, these keypoint-based methods highlight the advantage of bridging high-level VLM reasoning and low-level control through interpretable, spatially grounded intermediate representations.}

\subsubsection{\textbf{Subtask-based Methods}} 

{Distinct from the rigid logic of programs and the geometric constraints of keypoints, subtask-based methods leverage natural language instructions to maximize semantic flexibility and open-world generalization.}In this approach, the planner is typically a large VLM that receives high-level implicit instructions (e.g., clean up the table) along with observations, and decomposes them into step-by-step textual commands. Since these models produce interpretable intermediate instructions rather than executable actions, a low-level control policy is still required in practical deployment. Early efforts such as PaLM-E \cite{driess2023palme}, a LLaVA-style VLM trained with robotic manipulation data, demonstrate the feasibility of unifying general VQA capabilities with robot command generation. Building upon this direction, Embodied-Reasoner \cite{zhang2025embodied} introduces Observation–Thought–Action trajectories to support spatial analysis, reflection, and verification during step-wise planning. Reinforced Planning \cite{wu2025reinforced} retains this subtask-decomposition setup but boosts generalization through a two-stage pipeline of SFT followed by GRPO-based reinforcement fine-tuning. Meanwhile, some systems decouple perception from reasoning: Embodied-R \cite{zhao2025embodied} combines a large VLM for perception with a small LM for reasoning for embodied spatial reasoning on video, and its reasoning traces can be leveraged to derive stepwise subtask proposals for manipulation. In contrast, ViLA \cite{hu2023lookleapunveilingpower} leverages GPT-4V as an external planner, prompting it to generate candidate task plans from textual and visual inputs and then executing only the first command, discarding the remainder. 

{Overall, subtask-based planners translate complex instructions and visual observations into interpretable textual commands to guide execution policies. They excel in generating step-by-step text plans and performing complex tasks, yet their real-world performance remain limited by the capability of low-level control policies.}

\subsection{Planner+Policy}
\label{sec:hie-planner-policy}

\subsubsection{\textbf{Keypoint-based Methods}} 


{Within the integrated Planner+Policy architecture, keypoint-based methods leverage spatial primitives as a direct geometric bridge between high-level reasoning and low-level execution. By explicitly pairing a VLM planner with a downstream policy designed to consume these visual anchors, these models ensure that the planner's spatial intent is rigorously adhered to during continuous control.}
{This type of model generally uses a large VLM to ground subgoals as spatial primitives, spanning from discrete keypoints and continuous 2D trajectories to affordance representations and visual sketches.} The low-level policy then consumes these primitives to predict continuous trajectories and control. For instance, HAMSTER \cite{li2025hamster} first predicts trajectory keypoints from instructions and observations. Then it links these keypoints into an ordered path with a gradient color, and overlays this guidance for policy execution. Besides, ReKep \cite{huang2024rekep} utilizes DINOv2 \cite{oquab2023dinov2} and SAM \cite{kirillov2023segment} to produce keypoint proposals, then uses GPT-4o \cite{hurst2024gpt} to turn those keypoints into cost functions. These functions are subsequently solved into waypoints and actions by an optimizer. Moreover, $A_0$ \cite{xu2025a0} adopts an affordance-aware hierarchy: the planner predicts contact points and post-contact motion as an embodiment-agnostic affordance representation, which the action policy then converts into control. {To further enrich the spatial context, some works leverage visualized images—such as 2D trajectories or sketches overlaid on visual observations—to represent subsequent subtasks. Complementary to sparse point outputs, these trajectory- and sketch-conditioned formulations provide a lightweight yet expressive interface for downstream control. For example, RT-Trajectory \cite{gu2023rttrajectory} enables the policy to follow coarse motion guidance while grounding it in the current visual scene. Similarly, RT-Sketch \cite{pmlr-v270-sundaresan25a} interprets user-provided sketches as subtask goals, effectively guiding the robot even under ambiguous language instructions or visual distractors.} 

{In summary, coupling a keypoint-generating planner with a dedicated action policy provides robust spatial grounding, making these architectures highly adept at precision-demanding tasks.}

\subsubsection{\textbf{Subtask-based Methods}}

{Shifting from the geometric focus of spatial primitives to semantic abstraction, subtask-based hierarchical models utilize natural language as the universal interface between cognition and control. Subtask-based hierarchical models bridge the planner and policy with instructions}. The planner plays the same role as a subtask-based planner-only hierarchical model, and a low-level policy is appended to generate action sequences. {This explicit pairing allows the high-level VLM to handle complex, open-ended reasoning, while delegating the intricacies of high-frequency physical execution to a specialized downstream policy.} A representative example is HiRobot \cite{shi2025hi}, the planner of which accepts open-ended user instructions and then decomposes them into atomic commands for the policy. Similarly, DexVLA \cite{wen2025dexvla} features a VLM planner and a diffusion-based action policy. The policy is conditioned on the subtask tokens given by the VLM planner, excelling in complex and long-horizon robotic manipulation tasks. Built on DexVLA, PointVLA \cite{li2025pointvla} enhances spatial perception by incorporating a point cloud encoder and injector into the policy, enabling the model to follow the planner’s instructions in geometrically complex scenes. RoBridge \cite{zhang2025robridge} prompts the planner to generate text instructions of primitive actions and form an invariant operable representation for the policy to execute. Besides, SkillDiffuser \cite{liang2024skilldiffuser} decomposes complex tasks into subtasks through a high-level model that predicts a set of skills, while a low-level diffusion policy realizes concrete actions. Analogously, RoboMatrix \cite{mao2024robomatrix} organizes execution into a three-layer hierarchy: a modular scheduling layer generates subtask sequences, the skill layer encodes and selects reusable behaviors, and the hardware layer implements robot control. HiBerNAC \cite{wu2025hibernac} proposes an asynchronous, hierarchical framework. In this framework, the multi-agent neural structure first decomposes high-level instructions into structured subtasks. Then, an asynchronous pipeline manages these subtasks and coordinates the reactive VLA to execute the final low-level actions. Moreover, MALMM \cite{singh2024malmm} includes a planner, a supervisor, and a coder. The planner generates subtasks for the coder, and the supervisor coordinates transitions between modules. The coder serves as the policy and converts plans into executable robot code, including actions and positions. 

Ultimately, subtask-based Planner+Policy frameworks demonstrate how natural language can effectively decouple abstract reasoning from motor control. By converting open-ended goals into structured skill sequences, these models offer a highly interpretable and scalable approach to long-horizon manipulation, complementing the spatial precision of keypoint-based architectures.

\subsection{Comparison: Monolithic vs. Hierarchical}
\label{sec:hie-comp}
\textbf{Architectural comparison.}
The difference between monolithic and hierarchical architectures in large VLM-based VLA models for robotic manipulation lies primarily in how they map visual inputs and linguistic instructions into actions, either through a unified or modular approach. Monolithic models emphasize a single, integrated pipeline that jointly optimizes perception, reasoning, and control to directly translate high-level multimodal semantics into low-level actions. This design enables holistic and tightly coupled learning in robotic manipulation. Conversely, hierarchical architectures adopt a multi-stage design that explicitly separates high-level planning from low-level policy execution, promoting modularity and interpretability. This system decomposition allows components to be independently designed, trained, or replaced, enhancing flexibility and easing the integration of domain knowledge or adaptation to new robotic manipulation tasks. 

Another core distinction lies in the inherent nature of the intermediate processing. Monolithic models, while potentially embedding intricate internal reasoning, do so within latent spaces that are often opaque to external inspection. This implicit strategy allows them to fully exploit the representational capacity of large models, potentially discovering efficient task decompositions not explicitly designed by humans. Hierarchical systems, on the other hand, explicitly commit to generating explicit, human-understandable intermediate outputs. This makes them particularly advantageous in robotic manipulation scenarios that demand explainability, detailed task monitoring, or compatibility with traditional robotics pipelines, where high-level plans can be independently validated or modified.

Despite these differences, both approaches contribute unique strengths to the evolving landscape of VLA for robotic manipulation. Monolithic models highlight the power of unified learning and minimal manual decomposition, offering a streamlined route to generalization across diverse tasks. Hierarchical frameworks, by explicitly layering cognition and control, provide greater transparency and enhanced modular flexibility, which can be critical for complex multi-stage tasks or safety-critical deployments. Together, these paradigms illustrate complementary strategies for bridging vision-language understanding and embodied action, each offering valuable insights into the design of next-generation intelligent robotic manipulation systems. {To further quantify and empirically analyze the performance and inherent trade-offs of these two architectures across practical simulation benchmarks and physical deployments, we present a comprehensive comparative experimental evaluation in \textbf{Appx. A}.}

\textbf{Practical model selection.}
In practice, the architectural choice should be determined by deployment requirements, including control frequency and action precision, reasoning horizon and interpretability, computational and data budgets, and task-specific sensing requirements. To keep the main taxonomy concise, we provide a detailed baseline-centered model-selection guideline in \textbf{Appx. D}. The accompanying tables summarize representative single-system, dual-system, and hierarchical methods in terms of their corresponding baselines, addressed limitations, associated trade-offs, and recommended application scenarios.

\section{Other Advanced Field}
\label{advanced}
Beyond core VLA architectures, we highlight four directions for robustness, efficiency, and long-horizon planning: RL-based optimization, training-free improvements, learning from human videos, and world model-based VLA. Tab. \ref{tab:other} summarizes representative methods.
\label{Other Advanced Field}
\subsection{Reinforcement Learning-based Methods}
\label{RL-based methods}
Reinforcement learning (RL) plays an important role in enhancing VLA's generalization ability \cite{zhang2024grape} and task completion rate \cite{liu2025can}. It can be broadly categorized as online, which optimizes policies through real-time interaction, and offline, which learns from pre-collected trajectories. Most VLA methods are built upon traditional RL algorithms. We list several approaches in Tab. \ref{tab:RL}. 

Unlike RL in LLMs or VLMs \cite{ouyang2022training, jaech2024openai, huang2025visionr1} that usually finish generation in a few turns, VLA models typically involve long-horizon tasks with hundreds of steps in one trajectory, making rule-based reward functions prone to sparsity and instability \cite{lu2025vlarl, zhang2024grape}. To address this, many works incorporate learned dense reward signals. VLA-RL \cite{lu2025vlarl} trains a Robotic Process Reward Model (RPRM) to predict the success likelihood of action sequences. ReWiND \cite{zhang2025rewind} models reward as progress toward the final goal state, assigning higher rewards to states visually closer to task completion. In addition to training a reward model, Grape \cite{zhang2024grape} and TGRPO \cite{chen2025tgrpo} leverage powerful VLMs \cite{achiam2023gpt, hurst2024gpt} via prompting to generate feedback-based reward signals.

\begin{table}[t]
\scriptsize
\renewcommand{\arraystretch}{1.05}
\caption{Representative VLA methods grouped into four advanced categories: Reinforcement Learning-based, Training-Free, Learning from Human Videos, and World Model-based approaches. The grouping follows the methodological analysis in Sec.~\ref{Other Advanced Field}.}
\label{tab:other}

\begin{tabularx}{\linewidth}{p{3.5cm} X}
    \toprule
    \makecell[c]{\textbf{Description} \\[0.4ex]} &
    \makecell[c]{\textbf{Method} \\[0.4ex]} \\ 
    \midrule

    \rowcolor{lightgray} \multicolumn{2}{c}{\textbf{Reinforcement Learning-based Methods}} \\ \midrule
    Use RL to optimize robotic policies through interaction or pre-collected trajectories. & 
    VLA-RL \cite{lu2025vlarl}, ReWiND \cite{zhang2025rewind}, Grape \cite{zhang2024grape}, TGRPO \cite{chen2025tgrpo}, HIL-SERL \cite{luo2024precise}, ConRFT \cite{chen2025conrft}, RLDG \cite{xu2024rldg}, iRe-VLA \cite{guo2025improving} \\ 
    \midrule

    \rowcolor{lightgray} \multicolumn{2}{c}{\textbf{Training-Free Methods}} \\ \midrule
    Improve VLA models via architectural or computational optimizations without retraining. & 
    FlashVLA \cite{tan2025thinktwiceactonce}, EfficientVLA \cite{yang2025efficientvlatrainingfreeaccelerationcompression}, VLA-Cache \cite{xu2025vlacache}, PD-VLA \cite{pdvla}, SP-VLA \cite{li2025sp}, BAC \cite{ji2025bac}, FAST \cite{pertsch2025fast}, RTC \cite{black2025real} \\ 
    \midrule

    \rowcolor{lightgray} \multicolumn{2}{c}{\textbf{Learning from Human Videos}} \\ \midrule
    Leverage human videos to adapt robot policies, enabling cross-domain transfer. & 
    Human-Robot Semantic Alignment \cite{zhou2025mitigatinghumanrobotdomaindiscrepancy}, UniVLA \cite{bu2025learning}, LAPA \cite{ye2024latent}, VPDD \cite{he2024learningactionablediscretediffusion}, 3D-VLA \cite{zhen20243dvla3dvisionlanguageactiongenerative}, Humanoid-VLA \cite{ding2025humanoid} \\ 
    \midrule

    \rowcolor{lightgray} \multicolumn{2}{c}{\textbf{World Model-based VLA}} \\ \midrule
    Integrate predictive world models into VLA to model environment dynamics. & 
    World-VLA \cite{cen2025worldvla}, World4Omni \cite{chen2025world4omnizeroshotframeworkimage}, 3D-VLA \cite{zhen20243dvla3dvisionlanguageactiongenerative}, RIGVid \cite{patel2025robotic}, FoundationPose \cite{wen2024foundationpose}, V-JEPA 2-AC \cite{assran2025v} \\
    
    \bottomrule
\end{tabularx}
\end{table}

Besides reward sparsity, online learning is sample-inefficient due to slow simulation. To address this, several methods adopt hybrid offline-online training. ReWiND integrates offline Implicit Q-Learning \cite{kostrikov2021offline} with online Soft Actor-Critic \cite{haarnoja2018soft}, accelerating learning and enhancing safety in real-world deployment. HIL-SERL \cite{luo2024precise} is based on RLPD and introduces human-in-the-loop intervention in the training process. ConRFT \cite{chen2025conrft} employs a two-phase training scheme: an offline phase using Cal-ConRFT (combining behavior cloning and Q-learning) to initialize policy, followed by online HIL-ConRFT, which balances supervised and RL losses with human-in-the-loop intervention.

Furthermore, some methods leverage RL as a data engine to enhance generalist robotic models. RLDG \cite{xu2024rldg} is a prime example, which trains expert policies via HIL-SERL \cite{luo2024precise} to 100\% task success, then distills them into foundation models via sft. Similarly, iRe-VLA \cite{guo2025improving} alternates between online RL (to collect new successful trajectories) and sft on the expanded dataset, progressively improving the model through iterative imitation and exploration.

\subsection{Training-Free Methods}
\label{Traning-Free}
Training-free methods typically leverage modular and extensible designs to improve existing VLA architectures without training. This enables rapid prototyping, ablation studies, and targeted enhancements while preserving the model’s original capabilities and avoiding additional costs.

A range of training-free methods has been proposed to improve the efficiency of large VLM-based VLA models efficiency without retraining or architectural changes. A compelling example is FlashVLA \cite{tan2025thinktwiceactonce}, which employs a trigger mechanism that skips full decoding when action and visual cues remain stable, selectively reusing or pruning visual tokens. Similarly, EfficientVLA \cite{yang2025efficientvlatrainingfreeaccelerationcompression} prunes redundant language layers, filters task-relevant visual tokens, and caches intermediate features, while VLA-Cache \cite{xu2025vlacache} reuses cached key–value representations \cite{vaswani2017attention} of static tokens with task-relevance filtering and layer-adaptive reuse. Beyond these, SP-VLA \cite{li2025sp} combines spatio–semantic token pruning with an action-aware scheduler that routes intuitive steps to a lightweight generator and complex ones to the full VLA. Likewise, PD-VLA \cite{pdvla} reformulates autoregressive decoding under action chunking as a parallel fixed-point iteration, enabling simultaneous token prediction. FAST \cite{pertsch2025fast} compresses action sequences via discrete cosine transform and byte-pair encoding to reduce redundancy in high-frequency control, achieving up to 5× faster training. Furthermore, RTC \cite{black2025real} optimizes control-time scheduling by monitoring task progress and adjusting control frequency, reducing unnecessary computation during inference.

\subsection{Learning from Human Videos}

\begin{table}[t]
\scriptsize
\renewcommand{\arraystretch}{1.05}
\caption{Representative RL approaches for VLA. \ding{51} stands for online and \ding{55} stands for offline. ``S'' stands for sparse reward, ``D'' for dense reward, ``RM'' denotes a pre-trained reward model, ``GPT'' denotes a reward given by GPT, and ``TC'' denotes a reward function based on task completion.}
\label{tab:RL}
\begin{tabularx}{\linewidth}{C{1.8cm} C{1.5cm} C{1cm} X}
    \toprule
    \textbf{Method}
    & \textbf{RL Algo.}
    & \textbf{Online}
    & \textbf{Reward Formulation} \\ \midrule

    \rowcolor{lightgray}
    VLA-RL \cite{lu2025vlarl}
    & PPO
    & \ding{51}
    & TC (S) + RM (D) \\

    RIPT-VLA \cite{tan2025interactive}
    & LOOP
    & \ding{51}
    & TC (S) \\

    \rowcolor{lightgray}
    RLVLA \cite{liu2025can}
    & PPO
    & \ding{51}
    & TC (S) + Object Grasp (S) \\

    TPO \cite{zhang2024grape}
    & DPO
    & \ding{51}
    & TC (S) + GPT (S) \\ 

    \rowcolor{lightgray}
    ConRFT \cite{chen2025conrft}
    & -
    & \ding{55} + \ding{51}
    & TC (S)\\

    TGRPO \cite{chen2025tgrpo}
    & GRPO
    & \ding{51}
    & TC (S) + GPT (D) \\

    \rowcolor{lightgray}
    iRe-VLA \cite{guo2025improving}
    & PPO
    & \ding{51}
    & TC (S)\\

    RLDG \cite{xu2024rldg}
    & HIL-SERL
    & \ding{51}
    & TC (S)\\
    
    \rowcolor{lightgray}
    ReWiND \cite{zhang2025rewind}
    & IQL SAC
    & \ding{55} + \ding{51}
    & TC (S) + RM (D) \\

    HIL-SERL \cite{luo2024precise}
    & RLPD
    & \ding{55} + \ding{51}
    & TC (S) \\
    \bottomrule
\end{tabularx}
\end{table}
\label{Learning from Human Videos}
Recent advancements in VLA models have introduced several novel learning paradigms that surpass conventional policy learning strategies. Among them, a notable trend emerges: leveraging human video data to guide robot policy learning, using structural similarities in human-object and robot-object interactions to align visual and temporal cues.

This method aims to narrow the embodiment gap by transferring task-relevant knowledge from rich human video data. It uses human videos to pre-train perception or adapt policies. For instance, Human-Robot Semantic Alignment \cite{zhou2025mitigatinghumanrobotdomaindiscrepancy} aligns vision encoders across domains using paired human-robot videos, while UniVLA \cite{bu2025learning} learns task-centric latent actions from unlabeled human and robot videos to unify policy planning. Similarly, LAPA \cite{ye2024latent} leverages VQ-VAE \cite{van2017VQVAE} quantized latent actions to pre-train on large-scale video-language pairs, enabling transfer from actionless human videos. Besides, VPDD \cite{he2024learningactionablediscretediffusion} applies discrete diffusion over unified video tokens, facilitating cross-domain knowledge transfer through future dynamics prediction. 3D-VLA \cite{zhen20243dvla3dvisionlanguageactiongenerative} integrates human-object interaction videos with robot demonstrations for richer 3D reasoning, and Humanoid-VLA ~\cite{ding2025humanoid} exploits pose-recovered motion trajectories from online videos to enhance motion diversity. Collectively, these methods demonstrate that human video data can substantially enrich robot policy learning, yielding robustness and versatility even with limited robot data.


\subsection{World Model-based VLA}
\label{World Model}
World models, characterized by their ability to learn compact latent representations of environment dynamics, have emerged as powerful tools for enabling predictive reasoning and long-horizon planning. Recently, the integration of world models into VLA systems has gained traction as a promising strategy to enhance action planning through explicit modeling of environment dynamics.

Rather than directly generating actions from current observations, these approaches simulate future states, enabling agents to anticipate the consequences of their actions and refine decision-making accordingly. For instance, WorldVLA \cite{cen2025worldvla} introduces an autoregressive action world model that jointly learns to predict visual outcomes and generate actions within a unified token-based architecture. This mutual reinforcement between the world model and the action model improves both visual imagination and action fidelity, supporting more robust long-horizon planning. World4Omni \cite{chen2025world4omnizeroshotframeworkimage} employs a large-scale world model to produce subgoal images that depict intermediate task states. These generated visual cues are then used to guide a modular low-level policy, enabling zero-shot manipulation across diverse environments and robot embodiments. 3D-VLA \cite{zhen20243dvla3dvisionlanguageactiongenerative} employs a generative world model to predict future goal images and point clouds conditioned on instructions, effectively simulating scene dynamics. RIGVid \cite{patel2025robotic} uses a diffusion-based world model to generate candidate task videos, which are then filtered by a VLM for feasibility, with FoundationPose \cite{wen2024foundationpose} extracting gripper poses for execution. V-JEPA 2-AC \cite{assran2025v} builds a latent action-conditioned world model on top of a massive self-supervised video encoder trained on internet-scale data. During inference, it performs model-based planning by simulating future latent states conditioned on goal images, bridging dense video understanding with zero-shot robotic manipulation.


\section{Characteristics of VLA Models}
\label{Characteristics}

\subsection{Multimodal Fusion}
\label{Multimodal Fusion}
\textbf{Shared Embedding Space.} A defining characteristic of large VLM-based VLA models is the use of a shared, semantically aligned representation space for visual observations and linguistic instructions. Such alignment establishes direct semantic correspondence between perception and commands, providing a unified representation basis for subsequent action generation \cite{sapkota2025visionlanguageactionofactconceptsprogress}.

\textbf{Multimodal Token-Level Integration.} Beyond representation alignment, many VLA models integrate vision, language, proprioception, and actions at the token level within a unified model \cite{chen2025fast,wang2025unified,wang2024omnijarvisunifiedvisionlanguageactiontokenization}. This fine-grained interaction enables information from different modalities to dynamically condition one another throughout the perception--action cycle.

\textbf{Comprehensive Modal Compatibility.} Another characteristic is the extensibility of VLA models to richer sensory modalities beyond conventional vision and language. As discussed in Sec.~\ref{sig:performance} recent VLA models have incorporated information such as 3D perception, tactile sensing, and audio to complement visual observations. From a multimodal-fusion perspective, these developments demonstrate the flexibility of VLA frameworks in integrating heterogeneous sensory information while maintaining a unified perception--language--action paradigm.

\subsection{Instruction Following}
\label{Instruction Following}
\textbf{Semantic Instruction Grounding.} 
Instruction following in large VLM-based VLA models goes beyond the brittle pipelines of traditional robotic systems. These models leverage the world knowledge of pre-trained VLMs to ground natural language instructions into actions dynamically, enabling context-sensitive understanding rather than fixed templates. For instance, ChatVLA‑2 \cite{chatvla2} can interpret math problems written on whiteboards. Despite lacking explicit training in mathematics, it demonstrates an ability to generalize from visual-linguistic priors.

\textbf{Task Decomposition and Collaboration.} Beyond direct instruction-to-action translation, instruction following often involves hierarchical task decomposition, especially for long-horizon tasks. Specifically, high-level commands are broken down into sub-goals, which are then executed by low-level controllers \cite{shi2025hi}. LoHoVLA \cite{yang2025lohovlaunifiedvisionlanguageactionmodel} 
exemplifies this paradigm by generating intermediate subtasks from visual and language inputs, which guide action generation and maintain semantic alignment throughout execution. By continuously synchronizing subtask intentions with actions, the model maintains coherence in long-horizon tasks, supporting reliable goal completion.

\textbf{Explicit Reasoning via Chain-of-Thought.} A key fact of the instruction following capability of VLA models is the integration of chain‑of‑thought reasoning into the decision-making loop. In CoT‑VLA \cite{cotVLA}, for instance, the model interposes latent visual goals—predicting future images—before generating corresponding action sequences, effectively embedding a sub-goal generation mechanism. This explicit reasoning ties instructions to anticipated visual outcomes. It helps mitigate shortsighted hallucinations frequently observed in monolithic systems. In addition, it allows iterative refinement of plans, thereby facilitating more reliable execution of complex, multi-step tasks.

\subsection{Multi-Dimensional Generalization}
\label{Multi-Dimnsional Generalization}

\textbf{Cross-Task Generalization.} {Most VLA models exhibit zero-shot and few-shot generalizations, representing a leap beyond the brittle performance of traditional robotic manipulation models that often require extensive retraining.} A prime example is DexVLA \cite{wen2025dexvla}, which couples a large VLM with a billion-parameter diffusion action expert and an embodied curriculum to transfer skills across embodiments and rapidly adapt to unseen tasks. Without task-specific tuning, it attains consistently high success on diverse manipulations and generalizes to novel objects and scenes. Under direct language prompting, it also completes complex long-horizon workflows and outperforms state-of-the-art baselines such as OpenVLA \cite{kim2024openvla} and $\pi_0$ \cite{black2024pi0}.

\textbf{Cross-Domain Data Generalization.} This capacity extends further: $\pi_{0.5}$ and similar ``foundation VLA'' models are co-trained on heterogeneous data—web text, simulation videos, and cross-robot embodiments—to acquire broad contextual and semantic understanding. As a result, they successfully deploy ``out-of-the-box'' in new home environments, executing multi-stage tasks like dishwashing with OOD success rates exceeding 90\% \cite{black2024pi0}. This wide compatibility starkly contrasts with traditional pipelines, which typically collapse outside their training distribution.

\textbf{Cross-Embodiment and Sim-to-Real Generalization.} Cross‑embodiment and sim‑to‑real transfer represent another dimension of generalization. Hierarchical VLA architectures, which learn high-level planners in VLM space and delegate low-level control to domain-specific decoders, enable leveraging off‑domain data (e.g., hand-drawn sketches or simulations in diverse setups) to generalize across varying dynamics and robot morphologies. One such model, HAMSTER \cite{li2025hamster}, achieves a 20\% success boost over OpenVLA across seven axes of generalization.

\section{Conclusion}
\label{conclusion}

In conclusion, this survey presents a comprehensive synthesis of large VLM-based VLA models in robotic manipulation. By proposing a taxonomy of Monolithic and Hierarchical architectures, alongside a unified framework of datasets, benchmarks, and advanced learning paradigms, we structure the current landscape and its defining characteristics. Looking forward, future priorities must address cross-embodiment adaptation, scalable real-world deployment, and tighter coupling of reasoning and execution using internet-scale data. We aim for this survey to serve as a foundational step toward embodied AI that unifies perception, language, and action.


\begingroup
\let\,\relax
\bibliographystyle{IEEEtran}
\bibliography{IEEEabrv,main}

\begin{thebibliography}{100}
\providecommand{\url}[1]{#1}
\csname url@samestyle\endcsname
\providecommand{\newblock}{\relax}
\providecommand{\bibinfo}[2]{#2}
\providecommand{\BIBentrySTDinterwordspacing}{\spaceskip=0pt\relax}
\providecommand{\BIBentryALTinterwordstretchfactor}{4}
\providecommand{\BIBentryALTinterwordspacing}{\spaceskip=\fontdimen2\font plus
\BIBentryALTinterwordstretchfactor\fontdimen3\font minus \fontdimen4\font\relax}
\providecommand{\BIBforeignlanguage}[2]{{%
\expandafter\ifx\csname l@#1\endcsname\relax
\typeout{** WARNING: IEEEtran.bst: No hyphenation pattern has been}%
\typeout{** loaded for the language `#1'. Using the pattern for}%
\typeout{** the default language instead.}%
\else
\language=\csname l@#1\endcsname
\fi
#2}}
\providecommand{\BIBdecl}{\relax}
\BIBdecl

\bibitem{intro1}
Z.~Xu, K.~Wu, J.~Wen \emph{et~al.}, ``A survey on robotics with foundation models: toward embodied ai,'' \emph{arXiv:2402.02385}, 2024.

\bibitem{intro2}
Y.~Ma, Z.~Song, Y.~Zhuang \emph{et~al.}, ``A survey on vision-language-action models for embodied ai,'' \emph{arXiv:2405.14093}, 2024.

\bibitem{intro3}
W.~Wu, H.~He, Y.~Wang \emph{et~al.}, ``Metaurban: a simulation platform for embodied ai in urban spaces,'' in \emph{ICLR}, 2025.

\bibitem{intro4}
K.~Zhang, P.~Yun, J.~Cen \emph{et~al.}, ``Generative artificial intelligence in robotic manipulation: a survey,'' \emph{arXiv:2503.03464}, 2025.

\bibitem{intro5}
Y.~Liu, W.~Chen, Y.~Bai \emph{et~al.}, ``Aligning cyber space with physical world: a comprehensive survey on embodied ai,'' \emph{arXiv:2407.06886}, 2024.

\bibitem{intro6}
Y.~Liu, X.~Cao, T.~Chen \emph{et~al.}, ``A survey of embodied ai in healthcare: techniques, applications, and opportunities,'' \emph{INFORM FUSION}, vol. 119, p. 103033, 2025.

\bibitem{intro7}
S.~Nahavandi, R.~Alizadehsani, D.~Nahavandi \emph{et~al.}, ``Machine learning meets advanced robotic manipulation,'' \emph{INFORM FUSION}, vol. 105, p. 102221, 2024.

\bibitem{sapkota2025visionlanguageactionofactconceptsprogress}
R.~Sapkota, Y.~Cao, K.~I. Roumeliotis \emph{et~al.}, ``Vision-language-action models: concepts, progress, applications and challenges,'' \emph{arXiv:2505.04769}, 2025.

\bibitem{intro8}
A.~Billard and D.~Kragic, ``Trends and challenges in robot manipulation,'' \emph{Science}, vol. 364, p. eaat8414, 2019.

\bibitem{intro10}
C.~Wen, X.~Lin, J.~So \emph{et~al.}, ``Any-point trajectory modeling for policy learning,'' in \emph{RSS}, 2024, p.~92.

\bibitem{intro9}
P.-L. Guhur, S.~Chen, R.~G. Pinel \emph{et~al.}, ``Instruction-driven history-aware policies for robotic manipulations,'' in \emph{CoRL}, 2023, pp. 175--187.

\bibitem{intro11}
M.~Xu, Z.~Xu, Y.~Xu \emph{et~al.}, ``Flow as the cross-domain manipulation interface,'' in \emph{CoRL}, 2024.

\bibitem{intro12}
J.~Gu, S.~Kirmani, P.~Wohlhart \emph{et~al.}, ``Rt-trajectory: Robotic task generalization via hindsight trajectory sketches,'' in \emph{ICLR}, 2024, pp. 2475--2499.

\bibitem{intro13}
J.~Zhao, N.~Kuppuswamy, S.~Feng \emph{et~al.}, ``Polytouch: A robust multi-modal tactile sensor for contact-rich manipulation using tactile-diffusion policies,'' in \emph{ICRA}, 2025.

\bibitem{intro14}
V.~Mengers and O.~Brock, ``No plan but everything under control: Robustly solving sequential tasks with dynamically composed gradient descent,'' in \emph{ICRA}, 2025.

\bibitem{shridhar2022cliport}
M.~Shridhar, L.~Manuelli, and D.~Fox, ``Cliport: what and where pathways for robotic manipulation,'' in \emph{CoRL}, 2022, pp. 894--906.

\bibitem{zeng2021transporter}
A.~Zeng, P.~Florence, J.~Tompson \emph{et~al.}, ``Transporter networks: rearranging the visual world for robotic manipulation,'' in \emph{CoRL}, 2021, pp. 726--747.

\bibitem{ahn2022can}
M.~Ahn, A.~Brohan, N.~Brown \emph{et~al.}, ``Do as i can, not as i say: grounding language in robotic affordances,'' \emph{arXiv:2204.01691}, 2022.

\bibitem{huang2022inner}
W.~Huang, F.~Xia, T.~Xiao \emph{et~al.}, ``Inner monologue: Embodied reasoning through planning with language models,'' \emph{arXiv:2207.05608}, 2022.

\bibitem{brohan2022rt}
A.~Brohan, N.~Brown, J.~Carbajal \emph{et~al.}, ``Rt-1: robotics transformer for real-world control at scale,'' in \emph{RSS}, 2023.

\bibitem{driess2023palme}
D.~Driess, F.~Xia, M.~S.~M. Sajjadi \emph{et~al.}, ``Palm-e: an embodied multimodal language model,'' in \emph{ICML}, 2023, pp. 8469--8488.

\bibitem{kroemer2021review}
O.~Kroemer, S.~Niekum, and G.~Konidaris, ``A review of robot learning for manipulation: Challenges, representations, and algorithms,'' \emph{Journal of machine learning research}, vol.~22, no.~30, pp. 1--82, 2021.

\bibitem{liu2024improved}
H.~Liu, C.~Li, Y.~Li \emph{et~al.}, ``Improved baselines with visual instruction tuning,'' in \emph{CVPR}, 2024, pp. 26\,296--26\,306.

\bibitem{dai2023instructblip}
W.~Dai, J.~Li, D.~Li \emph{et~al.}, ``Instructblip: towards general-purpose vision-language models with instruction tuning,'' in \emph{NeurIPS}, 2023, pp. 49\,250--49\,267.

\bibitem{bai2023qwen}
J.~Bai, S.~Bai, S.~Yang \emph{et~al.}, ``Qwen-vl: a versatile vision-language model for understanding, localization, text reading, and beyond,'' \emph{arXiv:2308.12966}, 2023.

\bibitem{liu2023visual}
H.~Liu, C.~Li, Q.~Wu \emph{et~al.}, ``Visual instruction tuning,'' in \emph{NeurIPS}, 2023, pp. 34\,892--34\,916.

\bibitem{chen2024expanding}
Z.~Chen, W.~Wang, Y.~Cao \emph{et~al.}, ``Expanding performance boundaries of open-source multimodal models with model, data, and test-time scaling,'' \emph{arXiv:2412.05271}, 2024.

\bibitem{li2024monkey}
Z.~Li, B.~Yang, Q.~Liu \emph{et~al.}, ``Monkey: image resolution and text label are important things for large multi-modal models,'' in \emph{CVPR}, 2024, pp. 26\,763--26\,773.

\bibitem{kim2024openvla}
M.~J. Kim, K.~Pertsch, S.~Karamcheti \emph{et~al.}, ``Openvla: an open-source vision-language-action model,'' in \emph{CoRL}, 2024, pp. 2679--2713.

\bibitem{brohan2023rt2visionlanguageactionmodelstransfer}
B.~Zitkovich, T.~Yu, S.~Xu \emph{et~al.}, ``Rt-2: vision-language-action models transfer web knowledge to robotic control,'' in \emph{CoRL}, 2023, pp. 2165--2183.

\bibitem{belkhale2024rt}
S.~Belkhale, T.~Ding, T.~Xiao \emph{et~al.}, ``Rt-h: action hierarchies using language,'' in \emph{RSS}, 2024.

\bibitem{black2024pi0}
K.~Black, N.~Brown, D.~Driess \emph{et~al.}, ``$\pi_0$: A vision-language-action flow model for general robot control,'' in \emph{RSS}, 2025.

\bibitem{pi052025phy}
P.~Intelligence, N.~B. Kevin~Black, K.~D. James~Darpinian \emph{et~al.}, ``$\pi_{0.5}$: a vision language-action model with open-world generalization,'' \emph{arXiv:2504.16054}, 2025.

\bibitem{smolvla}
M.~Shukor, D.~Aubakirova, F.~Capuano \emph{et~al.}, ``Smolvla: a vision-language-action model for affordable and efficient robotics,'' \emph{arXiv:2506.01844}, 2025.

\bibitem{bjorck2025gr00t}
J.~Bjorck, F.~Casta{\~n}eda, N.~Cherniadev \emph{et~al.}, ``Gr00t n1: an open foundation model for generalist humanoid robots,'' \emph{arXiv:2503.14734}, 2025.

\bibitem{cotVLA}
Q.~Zhao, Y.~Lu, M.~J. Kim \emph{et~al.}, ``Cot-vla: visual chain-of-thought reasoning for vision-language-action models,'' in \emph{CVPR}, 2025, pp. 1702--1713.

\bibitem{hybridvla}
J.~Liu, H.~Chen, P.~An \emph{et~al.}, ``Hybridvla: collaborative diffusion and autoregression in a unified vision-language-action model,'' \emph{arXiv:2503.10631}, 2025.

\bibitem{li2025bridgevla}
P.~Li, Y.~Chen, H.~Wu \emph{et~al.}, ``Bridgevla: input-output alignment for efficient 3d manipulation learning with vision-language models,'' \emph{arXiv:2506.07961}, 2025.

\bibitem{deervla}
Y.~Yue, Y.~Wang, B.~Kang \emph{et~al.}, ``Deer-vla: dynamic inference of multimodal large language models for efficient robot execution,'' in \emph{NeurIPS}, 2024, pp. 56\,619--56\,643.

\bibitem{dan2025pi05ki}
J.~T.~S. Danny~Driess, L.~Y. Brian~Ichter, K.~P. Adrian Li-Bell \emph{et~al.}, ``Knowledge insulating vision-language-action models: train fast, run fast, generalize better,'' \emph{arXiv:2505.23705}, 2025.

\bibitem{cen2025worldvla}
J.~Cen, C.~Yu, H.~Yuan \emph{et~al.}, ``Worldvla: towards autoregressive action world model,'' \emph{arXiv:2506.21539}, 2025.

\bibitem{zhang2025rewind}
J.~Zhang, Y.~Luo, A.~Anwar \emph{et~al.}, ``Rewind: language-guided rewards teach robot policies without new demonstrations,'' \emph{arXiv:2505.10911}, 2025.

\bibitem{chen2025fast}
H.~Chen, J.~Liu, C.~Gu \emph{et~al.}, ``Fast-in-slow: a dual-system foundation model unifying fast manipulation within slow reasoning,'' \emph{arXiv:2506.01953}, 2025.

\bibitem{black2025real}
K.~Black, M.~Y. Galliker, and S.~Levine, ``Real-time execution of action chunking flow policies,'' \emph{arXiv:2506.07339}, 2025.

\bibitem{zhou2025mitigatinghumanrobotdomaindiscrepancy}
J.~Zhou, T.~Ma, K.~Y. Lin \emph{et~al.}, ``Mitigating the human-robot domain discrepancy in visual pre-training for robotic manipulation,'' in \emph{CVPR}, 2025, pp. 22\,551--22\,561.

\bibitem{chen2025world4omnizeroshotframeworkimage}
H.~Chen, B.~Wang, J.~Guo \emph{et~al.}, ``World4omni: a zero-shot framework from image generation world model to robotic manipulation,'' \emph{arXiv:2506.23919}, 2025.

\bibitem{kim2025oft}
M.~J. Kim, C.~Finn, and P.~Liang, ``Fine-tuning vision-language-action models: optimizing speed and success,'' in \emph{RSS}, 2025.

\bibitem{hung2025nora}
C.-Y. Hung, Q.~Sun, P.~Hong \emph{et~al.}, ``Nora: a small open-sourced generalist vision language action model for embodied tasks,'' \emph{arXiv:2504.19854}, 2025.

\bibitem{song2025maniplvm}
Z.~Song, G.~Ouyang, M.~Li \emph{et~al.}, ``Maniplvm-r1: reinforcement learning for reasoning in embodied manipulation with large vision-language models,'' \emph{arXiv:2505.16517}, 2025.

\bibitem{zhang2025embodied}
W.~Zhang, M.~Wang, G.~Liu \emph{et~al.}, ``Embodied-reasoner: synergizing visual search, reasoning, and action for embodied interactive tasks,'' \emph{arXiv:2503.21696}, 2025.

\bibitem{li2025hamster}
Y.~Li, Y.~Deng, J.~Zhang \emph{et~al.}, ``Hamster: hierarchical action models for open-world robot manipulation,'' in \emph{ICLR}, 2025.

\bibitem{xu2025a0}
R.~Xu, J.~Zhang, M.~Guo \emph{et~al.}, ``A0: an affordance-aware hierarchical model for general robotic manipulation,'' \emph{arXiv:2504.12636}, 2025.

\bibitem{huang2024rekep}
W.~Huang, C.~Wang, Y.~Li \emph{et~al.}, ``Rekep: spatio-temporal reasoning of relational keypoint constraints for robotic manipulation,'' in \emph{CoRL}, 2024.

\bibitem{intro22}
Y.~Du, Z.~Liu, J.~Li \emph{et~al.}, ``A survey of vision-language pre-trained models,'' in \emph{IJCAI}, 2022.

\bibitem{intro15}
J.~Zhang, J.~Huang, S.~Jin \emph{et~al.}, ``Vision-language models for vision tasks: A survey,'' \emph{TPAMI}, vol.~46, pp. 5625--5644, 2024.

\bibitem{intro19}
J.~Wu, W.~Gan, Z.~Chen \emph{et~al.}, ``Multimodal large language models: A survey,'' in \emph{BigData}, 2023, pp. 2247--2256.

\bibitem{intro20}
Z.~Li, X.~Wu, H.~Du \emph{et~al.}, ``A survey of state of the art large vision language models: Alignment, benchmark, evaluations and challenges,'' \emph{arXiv:2501.02189}, 2025.

\bibitem{intro21}
D.~Shu, H.~Zhao, J.~Hu \emph{et~al.}, ``Large vision-language model alignment and misalignment: A survey through the lens of explainability,'' \emph{arXiv:2501.01346}, 2025.

\bibitem{intro16}
M.~Song, X.~Deng, Z.~Zhou \emph{et~al.}, ``A survey on diffusion policy for robotic manipulation: Taxonomy, analysis, and future directions,'' \emph{Authorea Preprints}, 2025.

\bibitem{intro17}
R.~Wolf, Y.~Shi, S.~Liu \emph{et~al.}, ``Diffusion models for robotic manipulation: A survey,'' \emph{arXiv:2504.08438}, 2025.

\bibitem{intro18}
C.~Cui, P.~Ding, W.~Song \emph{et~al.}, ``Openhelix: A short survey, empirical analysis, and open-source dual-system vla model for robotic manipulation,'' \emph{arXiv:2505.03912}, 2025.

\bibitem{intro23}
Y.~Zheng, L.~Yao, Y.~Su \emph{et~al.}, ``A survey of embodied learning for object-centric robotic manipulation,'' \emph{MIR}, vol.~22, pp. 588--626, 2025.

\bibitem{chen2024lion}
G.~Chen, L.~Shen, R.~Shao \emph{et~al.}, ``Lion: empowering multimodal large language model with dual-level visual knowledge,'' in \emph{CVPR}, 2024, pp. 26\,540--26\,550.

\bibitem{zhang2025falcon}
R.~Zhang, R.~Shao, G.~Chen \emph{et~al.}, ``Falcon: Resolving visual redundancy and fragmentation in high-resolution multimodal large language models via visual registers,'' in \emph{ICCV}, 2025.

\bibitem{shen2024mome}
L.~Shen, G.~Chen, R.~Shao \emph{et~al.}, ``Mome: Mixture of multimodal experts for generalist multimodal large language models,'' in \emph{NeurIPS}, 2024.

\bibitem{li2025lion}
W.~Li, B.~Hu, R.~Shao \emph{et~al.}, ``Lion-fs: Fast \& slow video-language thinker as online video assistant,'' in \emph{CVPR}, 2025.

\bibitem{li2025optimus}
Z.~Li, Y.~Xie, R.~Shao \emph{et~al.}, ``Optimus-1: Hybrid multimodal memory empowered agents excel in long-horizon tasks,'' in \emph{NeurIPS}, 2024.

\bibitem{li2025optimus2}
------, ``Optimus-2: Multimodal minecraft agent with goal-observation-action conditioned policy,'' in \emph{CVPR}, 2025.

\bibitem{ye2024cat}
Q.~Ye, Z.~Yu, R.~Shao \emph{et~al.}, ``Cat: Enhancing multimodal large language model to answer questions in dynamic audio-visual scenarios,'' in \emph{ECCV}, 2024.

\bibitem{ye2025cat+}
------, ``Cat+: investigating and enhancing audio-visual understanding in large language models,'' \emph{IEEE Transactions on Pattern Analysis and Machine Intelligence}, 2025.

\bibitem{alayrac2022flamingo}
J.-B. Alayrac, J.~Donahue, P.~Luc \emph{et~al.}, ``Flamingo: a visual language model for few-shot learning,'' \emph{NeurIPS}, vol.~35, pp. 23\,716--23\,736, 2022.

\bibitem{peng2023kosmos}
Z.~Peng, W.~Wang, L.~Dong \emph{et~al.}, ``Kosmos-2: Grounding multimodal large language models to the world,'' \emph{arXiv:2306.14824}, 2023.

\bibitem{ma2023crepe}
Z.~Ma, J.~Hong, M.~O. Gul \emph{et~al.}, ``Crepe: Can vision-language foundation models reason compositionally?'' in \emph{CVPR}, 2023, pp. 10\,910--10\,921.

\bibitem{hsieh2023sugarcrepe}
C.-Y. Hsieh, J.~Zhang, Z.~Ma \emph{et~al.}, ``Sugarcrepe: Fixing hackable benchmarks for vision-language compositionality,'' \emph{NeurIPS}, vol.~36, pp. 31\,096--31\,116, 2023.

\bibitem{shah2023lm}
D.~Shah, B.~Osi{\'n}ski, S.~Levine \emph{et~al.}, ``Lm-nav: Robotic navigation with large pre-trained models of language, vision, and action,'' in \emph{CoRL}.\hskip 1em plus 0.5em minus 0.4em\relax pmlr, 2023, pp. 492--504.

\bibitem{liu2025msnav}
C.~Liu, Z.~Zhou, J.~Zhang \emph{et~al.}, ``Msnav: Zero-shot vision-and-language navigation with dynamic memory and llm spatial reasoning,'' \emph{arXiv:2508.16654}, 2025.

\bibitem{tiandrivevlm}
X.~Tian, J.~Gu, B.~Li \emph{et~al.}, ``Drivevlm: the convergence of autonomous driving and large vision-language models,'' in \emph{CoRL}, 2025, pp. 4698--4726.

\bibitem{hong2024cogagent}
W.~Hong, W.~Wang, Q.~Lv \emph{et~al.}, ``Cogagent: a visual language model for gui agents,'' in \emph{CVPR}, 2024, pp. 14\,281--14\,290.

\bibitem{chen2025less}
G.~Chen, X.~Zhou, R.~Shao \emph{et~al.}, ``Less is more: Empowering gui agent with context-aware simplification,'' in \emph{ICCV}, 2025.

\bibitem{lyu2025puma}
Y.~Lyu, R.~Shao, G.~Chen \emph{et~al.}, ``Puma: Layer-pruned language model for efficient unified multimodal retrieval with modality-adaptive learning,'' in \emph{ACM MM}, 2025.

\bibitem{zhu2025emosym}
Y.~Zhu, Y.~Lyu, Z.~Yu \emph{et~al.}, ``Emosym: A symbiotic framework for unified emotional understanding and generation via latent reasoning,'' in \emph{ACM MM}, 2025.

\bibitem{xie2025gui}
B.~Xie, R.~Shao, G.~Chen \emph{et~al.}, ``Gui-explorer: Autonomous exploration and mining of transition-aware knowledge for gui agent,'' in \emph{ACL}, 2025.

\bibitem{shao2025robust}
R.~Shao, T.~Wu, and Z.~Liu, ``Robust sequential deepfake detection,'' \emph{International Journal of Computer Vision}, vol. 133, pp. 3278--3295, 2025.

\bibitem{shao2024detecting}
R.~Shao, T.~Wu, J.~Wu \emph{et~al.}, ``Detecting and grounding multi-modal media manipulation and beyond,'' \emph{IEEE Transactions on Pattern Analysis and Machine Intelligence}, vol.~46, pp. 5556--5574, 2024.

\bibitem{shao2023detecting}
R.~Shao, T.~Wu, and Z.~Liu, ``Detecting and grounding multi-modal media manipulation,'' in \emph{CVPR}, 2023.

\bibitem{shao2019multi}
R.~Shao, X.~Lan, J.~Li \emph{et~al.}, ``Multi-adversarial discriminative deep domain generalization for face presentation attack detection,'' in \emph{CVPR}, 2019.

\bibitem{shao2025deepfake}
R.~Shao, T.~Wu, L.~Nie \emph{et~al.}, ``Deepfake-adapter: Dual-level adapter for deepfake detection,'' \emph{International Journal of Computer Vision}, vol. 133, pp. 3613--3628, 2025.

\bibitem{chen2025spabench}
J.~Chen, D.~Yuen, B.~Xie \emph{et~al.}, ``Spa-bench: A comprehensive benchmark for smartphone agent evaluation,'' in \emph{ICLR}, 2025.

\bibitem{li2025llavaonevision}
B.~Li, Y.~Zhang, D.~Guo \emph{et~al.}, ``{Ll}a{va}-onevision: easy visual task transfer,'' \emph{TMLR}, 2025.

\bibitem{bai2025qwen2}
S.~Bai, K.~Chen, X.~Liu \emph{et~al.}, ``Qwen2.5-vl technical report,'' \emph{arXiv:2502.13923}, 2025.

\bibitem{huang2025visionr1}
W.~Huang, B.~Jia, Z.~Zhai \emph{et~al.}, ``Vision-r1: incentivizing reasoning capability in multimodal large language models,'' \emph{arXiv:2503.06749}, 2025.

\bibitem{radford2021learning}
A.~Radford, J.~W. Kim, C.~Hallacy \emph{et~al.}, ``Learning transferable visual models from natural language supervision,'' in \emph{ICML}, 2021, pp. 8748--8763.

\bibitem{tan2019efficientnet}
M.~Tan and Q.~Le, ``Efficientnet: rethinking model scaling for convolutional neural networks,'' in \emph{ICML}, 2019, pp. 6105--6114.

\bibitem{shridhar2022peract}
M.~Shridhar, L.~Manuelli, and D.~Fox, ``Perceiver-actor: A multi-task transformer for robotic manipulation,'' in \emph{CoRL}, 2022.

\bibitem{jiang2022vima}
Y.~Jiang, A.~Gupta, Z.~Zhang \emph{et~al.}, ``Vima: general robot manipulation with multimodal prompts,'' in \emph{NeurIPS}, 2022.

\bibitem{zheng2022vlmbench}
K.~Zheng, X.~Chen, O.~Jenkins \emph{et~al.}, ``{VLM}bench: A compositional benchmark for vision-and-language manipulation,'' in \emph{NeurIPS}, 2022.

\bibitem{li2026semanticvla}
W.~Li, R.~Zhang, R.~Shao \emph{et~al.}, ``Semanticvla: Semantic-aligned sparsification and enhancement for efficient robotic manipulation,'' in \emph{Proceedings of the AAAI Conference on Artificial Intelligence}, vol.~40, no.~22, 2026, pp. 18\,397--18\,405.

\bibitem{li2026consisvla}
W.~Li, J.~Liu, L.~Yixing \emph{et~al.}, ``Consisvla-4d: Advancing spatiotemporal consistency in efficient 3d-perception and 4d-reasoning for robotic manipulation,'' \emph{arXiv preprint arXiv:2605.05126}, 2026.

\bibitem{li2026global}
Z.~Li, B.~Hu, R.~Shao \emph{et~al.}, ``Global prior meets local consistency: Dual-memory augmented vision-language-action model for efficient robotic manipulation,'' \emph{arXiv preprint arXiv:2602.20200}, 2026.

\bibitem{hu2026abstraction}
B.~Hu, Z.~Li, R.~Shao \emph{et~al.}, ``From abstraction to instantiation: Learning behavioral representation for vision-language-action model,'' \emph{arXiv preprint arXiv:2605.22671}, 2026.

\bibitem{zhu2026h}
Y.~Zhu, R.~Shao, Z.~Liu \emph{et~al.}, ``H-gar: A hierarchical interaction framework via goal-driven observation-action refinement for robotic manipulation,'' in \emph{Proceedings of the AAAI Conference on Artificial Intelligence}, vol.~40, no.~22, 2026, pp. 18\,882--18\,890.

\bibitem{lv2025spatial}
Q.~Lv, H.~Li, X.~Deng \emph{et~al.}, ``Spatial-temporal graph diffusion policy with kinematic modeling for bimanual robotic manipulation,'' in \emph{2025 IEEE/CVF Conference on Computer Vision and Pattern Recognition (CVPR)}.\hskip 1em plus 0.5em minus 0.4em\relax IEEE, 2025, pp. 17\,394--17\,404.

\bibitem{lv2025f1}
Q.~Lv, W.~Kong, H.~Li \emph{et~al.}, ``F1: A vision-language-action model bridging understanding and generation to actions,'' \emph{arXiv preprint arXiv:2509.06951}, 2025.

\bibitem{chen2023pali}
X.~Chen, J.~Djolonga, P.~Padlewski \emph{et~al.}, ``Pali-x: on scaling up a multilingual vision and language model,'' in \emph{CVPR}, 2024, pp. 14\,432--14\,444.

\bibitem{hu2022lora}
E.~J. Hu, Y.~Shen, P.~Wallis \emph{et~al.}, ``Lo{ra}: low-rank adaptation of large language models,'' in \emph{ICLR}, 2022.

\bibitem{wang2024large}
J.~Wang, E.~Shi, H.~Hu \emph{et~al.}, ``Large language models for robotics: opportunities, challenges, and perspectives,'' \emph{JAI}, vol.~4, pp. 52--64, 2025.

\bibitem{sun2025review}
J.~Sun, P.~Mao, L.~Kong \emph{et~al.}, ``A review of embodied grasping,'' \emph{Sensors}, vol.~25, p. 852, 2025.

\bibitem{yao2023bridging}
X.~Yao, H.~Zhou, O.~Mees \emph{et~al.}, ``Bridging language and action: A survey of language-conditioned robot manipulation,'' \emph{arXiv:2312.10807}, 2023.

\bibitem{li2024foundation}
D.~Li, Y.~Jin, Y.~Sun \emph{et~al.}, ``What foundation models can bring for robot learning in manipulation: A survey,'' \emph{The International Journal of Robotics Research}, p. 02783649251390579, 2024.

\bibitem{o2024open}
A.~O’Neill, A.~Rehman, A.~Maddukuri \emph{et~al.}, ``Open x-embodiment: robotic learning datasets and rt-x models: open x-embodiment collaboration 0,'' in \emph{ICRA}, 2024, pp. 6892--6903.

\bibitem{huang2024embodied}
J.~Huang, S.~Yong, X.~Ma \emph{et~al.}, ``An embodied generalist agent in 3d world,'' in \emph{ICML}, 2024, pp. 20\,413--20\,451.

\bibitem{chen2025trainingstrategiesefficientembodied}
W.~Chen, S.~Belkhale, S.~Mirchandani \emph{et~al.}, ``Training strategies for efficient embodied reasoning,'' \emph{arXiv:2505.08243}, 2025.

\bibitem{dey2024revla}
S.~Dey, J.-N. Zaech, N.~Nikolov \emph{et~al.}, ``Revla: reverting visual domain limitation of robotic foundation models,'' in \emph{ICRA}, 2025.

\bibitem{zheng2024tracevla}
R.~Zheng, Y.~Liang, S.~Huang \emph{et~al.}, ``Tracevla: visual trace prompting enhances spatial-temporal awareness for generalist robotic policies,'' in \emph{ICLR}, 2025.

\bibitem{jones2025beyond}
J.~Jones, O.~Mees, C.~Sferrazza \emph{et~al.}, ``Beyond sight: finetuning generalist robot policies with heterogeneous sensors via language grounding,'' in \emph{ICRA}, 2025.

\bibitem{zheng2025universal}
J.~Zheng, J.~Li, D.~Liu \emph{et~al.}, ``Universal actions for enhanced embodied foundation models,'' in \emph{CVPR}, 2025, pp. 22\,508--22\,519.

\bibitem{qu2025spatialvlaexploringspatialrepresentations}
D.~Qu, H.~Song, Q.~Chen \emph{et~al.}, ``Spatialvla: exploring spatial representations for visual-language-action model,'' in \emph{RSS}, 2025.

\bibitem{zhang2025up}
J.~Zhang, Y.~Guo, Y.~Hu \emph{et~al.}, ``Up-vla: a unified understanding and prediction model for embodied agent,'' in \emph{ICML}, 2025.

\bibitem{zhao2025vlas}
W.~Zhao, P.~Ding, Z.~Min \emph{et~al.}, ``Vlas: vision-language-action model with speech instructions for customized robot manipulation,'' in \emph{ICLR}, 2025.

\bibitem{zhang2025vtla}
C.~Zhang, P.~Hao, X.~Cao \emph{et~al.}, ``Vtla: vision-tactile-language-action model with preference learning for insertion manipulation,'' \emph{arXiv:2505.09577}, 2025.

\bibitem{zhao2025unveilingpotentialvisionlanguageactionmodels}
W.~Zhao, G.~Li, Z.~Gong \emph{et~al.}, ``Unveiling the potential of vision-language-action models with open-ended multimodal instructions,'' \emph{arXiv:2505.11214}, 2025.

\bibitem{vanvo2025refinevlareasoningawareteacherguidedtransfer}
T.~Van~Vo, T.~Q. Nguyen, K.~M. Nguyen \emph{et~al.}, ``Refinevla: reasoning-aware teacher-guided transfer fine-tuning,'' \emph{arXiv:2505.19080}, 2025.

\bibitem{yang2025lohovlaunifiedvisionlanguageactionmodel}
Y.~Yang, J.~Sun, S.~Kou \emph{et~al.}, ``Lohovla: a unified vision-language-action model for long-horizon embodied tasks,'' \emph{arXiv:2506.00411}, 2025.

\bibitem{wang2025unified}
Y.~Wang, X.~Li, W.~Wang \emph{et~al.}, ``Unified vision-language-action model,'' \emph{arXiv:2506.19850}, 2025.

\bibitem{zhang20254dvla}
J.~Zhang, Y.~Chen, Y.~Xu \emph{et~al.}, ``4d-vla: spatiotemporal vision-language-action pretraining with cross-scene calibration,'' \emph{arXiv:2506.22242}, 2025.

\bibitem{lin2025vote}
J.~Lin, A.~Taherin, A.~Akbari \emph{et~al.}, ``Vote: vision-language-action optimization with trajectory ensemble voting,'' \emph{arXiv:2507.05116}, 2025.

\bibitem{patratskiy2025stvla}
M.~A. Patratskiy, A.~K. Kovalev, and A.~I. Panov, ``Spatial traces: Enhancing vla models with spatial-temporal understanding,'' \emph{arXiv:2508.09032}, 2025.

\bibitem{li2023vision}
X.~Li, M.~Liu, H.~Zhang \emph{et~al.}, ``Vision-language foundation models as effective robot imitators,'' in \emph{ICLR}, 2024, pp. 1--19.

\bibitem{liu2024robomamba}
J.~Liu, M.~Liu, Z.~Wang \emph{et~al.}, ``Robomamba: efficient vision-language-action model for robotic reasoning and manipulation,'' in \emph{NeurIPS}, vol.~37, 2024, pp. 40\,085--40\,110.

\bibitem{pdvla}
W.~Song, J.~Chen, P.~Ding \emph{et~al.}, ``Accelerating vision-language-action model integrated with action chunking via parallel decoding,'' \emph{arXiv:2503.02310}, 2025.

\bibitem{zhang2025mole}
R.~Zhang, M.~Dong, Y.~Zhang \emph{et~al.}, ``Mole-vla: dynamic layer-skipping vision language action model via mixture-of-layers for efficient robot manipulation,'' \emph{arXiv:2503.20384}, 2025.

\bibitem{tan2025thinktwiceactonce}
X.~Tan, Y.~Yang, P.~Ye \emph{et~al.}, ``Think twice, act once: token-aware compression and action reuse for efficient inference in vision-language-action models,'' \emph{arXiv:2505.21200}, 2025.

\bibitem{wang2025bitvla}
H.~Wang, C.~Xiong, R.~Wang \emph{et~al.}, ``Bitvla: 1-bit vision-language-action models for robotics manipulation,'' \emph{arXiv:2506.07530}, 2025.

\bibitem{wang2025spec}
S.~Wang, R.~Yu, Z.~Yuan \emph{et~al.}, ``Spec-vla: speculative decoding for vision-language-action models with relaxed acceptance,'' \emph{arXiv:2507.22424}, 2025.

\bibitem{li2025cogvla}
W.~Li, R.~Zhang, R.~Shao \emph{et~al.}, ``Cogvla: Cognition-aligned vision-language-action model via instruction-driven routing \& sparsification,'' \emph{arXiv:2508.21046}, 2025.

\bibitem{vaswani2017attention}
A.~Vaswani, N.~Shazeer, N.~Parmar \emph{et~al.}, ``Attention is all you need,'' in \emph{NeurIPS}, 2017.

\bibitem{zhai2023sigmoid}
X.~Zhai, B.~Mustafa, A.~Kolesnikov \emph{et~al.}, ``Sigmoid loss for language image pre-training,'' in \emph{ICCV}, 2023, pp. 11\,975--11\,986.

\bibitem{oquab2023dinov2}
M.~Oquab, T.~Darcet, T.~Moutakanni \emph{et~al.}, ``{Dino}v2: learning robust visual features without supervision,'' \emph{TMLR}, 2024.

\bibitem{qi2017pointnet++}
C.~R. Qi, L.~Yi, H.~Su \emph{et~al.}, ``Pointnet++: deep hierarchical feature learning on point sets in a metric space,'' in \emph{NeurIPS}, vol.~30, 2017.

\bibitem{chen2022language}
S.~Chen, P.-L. Guhur, M.~Tapaswi \emph{et~al.}, ``Language conditioned spatial relation reasoning for 3d object grounding,'' \emph{NeurIPS}, vol.~35, pp. 20\,522--20\,535, 2022.

\bibitem{zawalski2025robotic}
M.~Zawalski, W.~Chen, K.~Pertsch \emph{et~al.}, ``Robotic control via embodied chain-of-thought reasoning,'' in \emph{CoRL}, 2025, pp. 3157--3181.

\bibitem{gu2023mamba}
A.~Gu and T.~Dao, ``Mamba: linear-time sequence modeling with selective state spaces,'' \emph{arXiv:2312.00752}, 2023.

\bibitem{nielsen2024bitnet}
J.~Nielsen and P.~Schneider-Kamp, ``Bitnet b1. 58 reloaded: state-of-the-art performance also on smaller networks,'' in \emph{DeLTA}, 2024, pp. 301--315.

\bibitem{pertsch2025fast}
K.~Pertsch, K.~Stachowicz, B.~Ichter \emph{et~al.}, ``Fast: efficient action tokenization for vision-language-action models,'' in \emph{RSS}, 2025.

\bibitem{han2024dual}
B.~Han, J.~Kim, and J.~Jang, ``A dual process vla: efficient robotic manipulation leveraging vlm,'' \emph{arXiv:2410.15549}, 2024.

\bibitem{bu2024towards}
Q.~Bu, H.~Li, L.~Chen \emph{et~al.}, ``Towards synergistic, generalized, and efficient dual-system for robotic manipulation,'' \emph{arXiv:2410.08001}, 2024.

\bibitem{shentu2024llms}
Y.~Shentu, P.~Wu, A.~Rajeswaran \emph{et~al.}, ``From llms to actions: latent codes as bridges in hierarchical robot control,'' in \emph{IROS}, 2024, pp. 8539--8546.

\bibitem{li2024cogact}
Q.~Li, Y.~Liang, Z.~Wang \emph{et~al.}, ``Cogact: a foundational vision-language-action model for synergizing cognition and action in robotic manipulation,'' \emph{arXiv:2411.19650}, 2024.

\bibitem{zhang2024hirt}
J.~Zhang, Y.~Guo, X.~Chen \emph{et~al.}, ``Hirt: enhancing robotic control with hierarchical robot transformers,'' in \emph{CoRL}, 2025, pp. 933--946.

\bibitem{chatvla}
Z.~Zhou, Y.~Zhu, M.~Zhu \emph{et~al.}, ``Chatvla: unified multimodal understanding and robot control with vision-language-action model,'' \emph{arXiv:2502.14420}, 2025.

\bibitem{chatvla2}
Z.~Zhou, Y.~Zhu, J.~Wen \emph{et~al.}, ``Chatvla-2: vision-language-action model with open-world embodied reasoning from pretrained knowledge,'' \emph{arXiv:2505.21906}, 2025.

\bibitem{diffusionvla}
J.~Wen, Y.~Zhu, M.~Zhu \emph{et~al.}, ``Diffusionvla: scaling robot foundation models via unified diffusion and autoregression,'' in \emph{ICML}, 2025.

\bibitem{liu2025trivla}
Z.~Liu, Y.~Gu, S.~Zheng \emph{et~al.}, ``Trivla: a unified triple-system-based unified vision-language-action model for general robot control,'' \emph{arXiv:2507.01424}, 2025.

\bibitem{li2025information}
S.~Li, L.~Gao, J.~Wang \emph{et~al.}, ``Information-theoretic graph fusion with vision-language-action model for policy reasoning and dual robotic control,'' \emph{arXiv:2508.05342}, 2025.

\bibitem{song2025rationalvla}
W.~Song, J.~Chen, W.~Li \emph{et~al.}, ``Rationalvla: a rational vision-language-action model with dual system,'' \emph{arXiv:2506.10826}, 2025.

\bibitem{wang2025vq}
J.~Jones, O.~Mees, C.~Sferrazza \emph{et~al.}, ``Vq-vla: improving vision-language-action models via scaling vector-quantized action tokenizers,'' in \emph{ICCV}, 2025.

\bibitem{wen2025tinyvla}
J.~Wen, Y.~Zhu, J.~Li \emph{et~al.}, ``Tinyvla: towards fast, data-efficient vision-language-action models for robotic manipulation,'' \emph{RA-L}, 2025.

\bibitem{yu2025forcevla}
J.~Yu, H.~Liu, Q.~Yu \emph{et~al.}, ``Forcevla: enhancing vla models with a force-aware moe for contact-rich manipulation,'' \emph{arXiv:2505.22159}, 2025.

\bibitem{lin2025onetwovla}
F.~Lin, R.~Nai, Y.~Hu \emph{et~al.}, ``Onetwovla: a unified vision-language-action model with adaptive reasoning,'' \emph{arXiv:2505.11917}, 2025.

\bibitem{huang2025tactile}
J.~Huang, S.~Wang, F.~Lin \emph{et~al.}, ``Tactile-vla: unlocking vision-language-action model's physical knowledge for tactile generalization,'' \emph{arXiv:2507.09160}, 2025.

\bibitem{cheang2025gr3}
C.~Cheang, S.~Chen, Z.~Cui \emph{et~al.}, ``Gr-3 technical report,'' \emph{arXiv:2507.15493}, 2025.

\bibitem{chen2025villa}
X.~Chen, H.~Wei, P.~Zhang \emph{et~al.}, ``Villa-x: enhancing latent action modeling in vision-language-action models,'' \emph{arXiv:2507.23682}, 2025.

\bibitem{deng25gra}
S.~Deng, M.~Yan, S.~Wei \emph{et~al.}, ``Graspvla: a grasping foundation model pre-trained on billion-scale synthetic action data,'' \emph{arXiv:2505.03233}, 2025.

\bibitem{peebles2023dit}
W.~Peebles and S.~Xie, ``Scalable diffusion models with transformers,'' in \emph{ICCV}, 2023, pp. 4195--4205.

\bibitem{mandlekar2021matters}
A.~Mandlekar, D.~Xu, J.~Wong \emph{et~al.}, ``What matters in learning from offline human demonstrations for robot manipulation,'' in \emph{CoRL}, 2022, pp. 1678--1690.

\bibitem{van2017VQVAE}
A.~van~den Oord, O.~Vinyals, and k.~kavukcuoglu, ``Neural discrete representation learning,'' in \emph{NeurIPS}, 2017, pp. 1--10.

\bibitem{ronneberger2015u}
O.~Ronneberger, P.~Fischer, and T.~Brox, ``U-net: Convolutional networks for biomedical image segmentation,'' in \emph{MICCAI}, 2015.

\bibitem{noam2017outr}
A.~M. Noam~Shazeer, A.~D. Krzysztof~Maziarz, G.~H. Quoc~Le \emph{et~al.}, ``Outrageously large neural networks: the sparsely-gated mixture-of-experts layer,'' in \emph{ICLR}, 2017.

\bibitem{momanipvla}
Z.~Wu, Y.~Zhou, X.~Xu \emph{et~al.}, ``Momanipvla: transferring vision-language-action models for general mobile manipulation,'' in \emph{CVPR}, 2025, pp. 1714--1723.

\bibitem{yuan2024robopoint}
W.~Yuan, J.~Duan, V.~Blukis \emph{et~al.}, ``Robopoint: a vision-language model for spatial affordance prediction in robotics,'' in \emph{CoRL}, 2024.

\bibitem{wu2025reinforced}
D.~Wu, J.~Fan, J.~Zang \emph{et~al.}, ``Reinforced reasoning for embodied planning,'' \emph{arXiv:2505.22050}, 2025.

\bibitem{wang2025chain}
C.~Wang, F.~Xia, W.~Yu \emph{et~al.}, ``Chain-of-modality: learning manipulation programs from multimodal human videos with vision-language-models,'' in \emph{ICRA}, 2025.

\bibitem{li2025robotic}
Y.~Li, Z.~Gong, H.~Li \emph{et~al.}, ``Robotic visual instruction,'' in \emph{CVPR}, 2025, pp. 12\,155--12\,165.

\bibitem{liu2025longhorizonembodiedplanningimplicit}
S.~Liu, J.~Du, S.~Xiang \emph{et~al.}, ``Long-horizon embodied planning with implicit logical inference and hallucination mitigation,'' \emph{arXiv:2409.15658}, 2025.

\bibitem{hu2023lookleapunveilingpower}
Y.~Hu, F.~Lin, T.~Zhang \emph{et~al.}, ``Look before you leap: unveiling the power of gpt-4v in robotic vision-language planning,'' \emph{arXiv:2311.17842}, 2023.

\bibitem{ji2025robobrain}
Y.~Ji, H.~Tan, J.~Shi \emph{et~al.}, ``Robobrain: a unified brain model for robotic manipulation from abstract to concrete,'' in \emph{CVPR}, 2025, pp. 1724--1734.

\bibitem{shi2025hi}
L.~X. Shi, B.~Ichter, M.~Equi \emph{et~al.}, ``Hi robot: open-ended instruction following with hierarchical vision-language-action models,'' in \emph{ICML}, 2025.

\bibitem{yang2025agentic}
Z.~Yang, Y.~Chen, X.~Zhou \emph{et~al.}, ``Agentic robot: a brain-inspired framework for vision-language-action models in embodied agents,'' \emph{arXiv:2505.23450}, 2025.

\bibitem{wen2025dexvla}
J.~Wen, Y.~Zhu, J.~Li \emph{et~al.}, ``Dexvla: vision-language model with plug-in diffusion expert for general robot control,'' in \emph{CoRL}, 2025.

\bibitem{huang2023instruct2act}
S.~Huang, Z.~Jiang, H.~Dong \emph{et~al.}, ``Instruct2act: mapping multi-modality instructions to robotic actions with large language model,'' \emph{arXiv:2305.11176}, 2023.

\bibitem{mao2024robomatrix}
W.~Mao, W.~Zhong, Z.~Jiang \emph{et~al.}, ``Robomatrix: a skill-centric hierarchical framework for scalable robot task planning and execution in open-world,'' \emph{arXiv:2412.00171}, 2024.

\bibitem{li2025pointvla}
C.~Li, J.~Wen, Y.~Peng \emph{et~al.}, ``Pointvla: injecting the 3d world into vision-language-action models,'' \emph{arXiv:2503.07511}, 2025.

\bibitem{yuan2025seeing}
Y.~Yuan, H.~Cui, Y.~Chen \emph{et~al.}, ``From seeing to doing: bridging reasoning and decision for robotic manipulation,'' \emph{arXiv:2505.08548}, 2025.

\bibitem{zhang2025robridge}
K.~Zhang, R.~Xu, P.~Ren \emph{et~al.}, ``Robridge: a hierarchical architecture bridging cognition and execution for general robotic manipulation,'' in \emph{ICCV}, 2025.

\bibitem{han2025robocerebralargescalebenchmarklonghorizon}
S.~Han, B.~Qiu, Y.~Liao \emph{et~al.}, ``Robocerebra: a large-scale benchmark for long-horizon robotic manipulation evaluation,'' \emph{arXiv:2506.06677}, 2025.

\bibitem{zhong2025dexgraspvla}
Y.~Zhong, X.~Huang, R.~Li \emph{et~al.}, ``Dexgraspvla: a vision-language-action framework towards general dexterous grasping,'' \emph{arXiv:2502.20900}, 2025.

\bibitem{huang2023voxposer}
W.~Huang, C.~Wang, R.~Zhang \emph{et~al.}, ``Voxposer: composable 3d value maps for robotic manipulation with language models,'' in \emph{CoRL}, 2023.

\bibitem{liang2024skilldiffuser}
Z.~Liang, Y.~Mu, H.~Ma \emph{et~al.}, ``Skilldiffuser: interpretable hierarchical planning via skill abstractions in diffusion-based task execution,'' in \emph{CVPR}, 2024, pp. 16\,467--16\,476.

\bibitem{nasiriany2024rtaffordanceaffordancesversatileintermediate}
S.~Nasiriany, S.~Kirmani, T.~Ding \emph{et~al.}, ``{RT}-affordance: reasoning about robotic manipulation with affordances,'' in \emph{CoRL}, 2024.

\bibitem{wu2025hibernac}
H.~Wu, H.~Zhang, P.~Zhang \emph{et~al.}, ``Hibernac: hierarchical brain-emulated robotic neural agent collective for disentangling complex manipulation,'' \emph{arXiv:2506.08296}, 2025.

\bibitem{niu2024llarva}
D.~Niu, Y.~Sharma, G.~Biamby \emph{et~al.}, ``{Llarva}: vision-action instruction tuning enhances robot learning,'' in \emph{CoRL}, 2024.

\bibitem{singh2024malmm}
H.~Singh, R.~J. Das, M.~Han \emph{et~al.}, ``Malmm: multi-agent large language models for zero-shot robotics manipulation,'' in \emph{CVPR}, 2024.

\bibitem{bi2025vlatouch}
J.~Bi, K.~Y. Ma, C.~Hao \emph{et~al.}, ``Vla-touch: Enhancing vision-language-action models with dual-level tactile feedback,'' \emph{arXiv:2507.17294}, 2025.

\bibitem{guo2025deepseek}
D.~Guo, D.~Yang, H.~Zhang \emph{et~al.}, ``Deepseek-r1: incentivizing reasoning capability in llms via reinforcement learning,'' \emph{arXiv:2501.12948}, 2025.

\bibitem{zhao2025embodied}
B.~Zhao, Z.~Wang, J.~Fang \emph{et~al.}, ``Embodied-r: collaborative framework for activating embodied spatial reasoning in foundation models via reinforcement learning,'' \emph{arXiv:2504.12680}, 2025.

\bibitem{kirillov2023segment}
A.~Kirillov, E.~Mintun, N.~Ravi \emph{et~al.}, ``Segment anything,'' in \emph{ICCV}, 2023, pp. 4015--4026.

\bibitem{hurst2024gpt}
A.~Hurst, A.~Lerer, A.~P. Goucher \emph{et~al.}, ``Gpt-4o system card,'' \emph{arXiv:2410.21276}, 2024.

\bibitem{gu2023rttrajectory}
J.~Gu, S.~Kirmani, P.~Wohlhart \emph{et~al.}, ``Rt-trajectory: Robotic task generalization via hindsight trajectory sketches,'' \emph{arXiv:2311.01977}, 2023.

\bibitem{pmlr-v270-sundaresan25a}
P.~Sundaresan, Q.~Vuong, J.~Gu \emph{et~al.}, ``Rt-sketch: Goal-conditioned imitation learning from hand-drawn sketches,'' in \emph{CoRL}, vol. 270.\hskip 1em plus 0.5em minus 0.4em\relax pmlr, 2025, pp. 70--96.

\bibitem{zhang2024grape}
Z.~Zhang, K.~Zheng, Z.~Chen \emph{et~al.}, ``Grape: generalizing robot policy via preference alignment,'' \emph{arXiv:2411.19309}, 2024.

\bibitem{liu2025can}
J.~Liu, F.~Gao, B.~Wei \emph{et~al.}, ``What can rl bring to vla generalization? an empirical study,'' \emph{arXiv:2505.19789}, 2025.

\bibitem{ouyang2022training}
L.~Ouyang, J.~Wu, X.~Jiang \emph{et~al.}, ``Training language models to follow instructions with human feedback,'' in \emph{NeurIPS}, 2022, pp. 27\,730--27\,744.

\bibitem{jaech2024openai}
A.~Jaech, A.~Kalai, A.~Lerer \emph{et~al.}, ``Openai o1 system card,'' \emph{arXiv:2412.16720}, 2024.

\bibitem{lu2025vlarl}
G.~Lu, W.~Guo, C.~Zhang \emph{et~al.}, ``Vla-rl: towards masterful and general robotic manipulation with scalable reinforcement learning,'' \emph{arXiv:2505.18719}, 2025.

\bibitem{chen2025tgrpo}
Z.~Chen, R.~Niu, H.~Kong \emph{et~al.}, ``Tgrpo: fine-tuning vision-language-action model via trajectory-wise group relative policy optimization,'' \emph{arXiv:2506.08440}, 2025.

\bibitem{achiam2023gpt}
J.~Achiam, S.~Adler, S.~Agarwal \emph{et~al.}, ``Gpt-4 technical report,'' \emph{arXiv:2303.08774}, 2023.

\bibitem{luo2024precise}
J.~Luo, C.~Xu, J.~Wu \emph{et~al.}, ``Precise and dexterous robotic manipulation via human-in-the-loop reinforcement learning,'' \emph{arXiv:2410.21845}, 2024.

\bibitem{chen2025conrft}
Y.~Chen, S.~Tian, S.~Liu \emph{et~al.}, ``Conrft: a reinforced fine-tuning method for vla models via consistency policy,'' \emph{arXiv:2502.05450}, 2025.

\bibitem{xu2024rldg}
C.~Xu, Q.~Li, J.~Luo \emph{et~al.}, ``Rldg: robotic generalist policy distillation via reinforcement learning,'' in \emph{RSS}, 2025.

\bibitem{guo2025improving}
Y.~Guo, J.~Zhang, X.~Chen \emph{et~al.}, ``Improving vision-language-action model with online reinforcement learning,'' in \emph{ICRA}, 2025.

\bibitem{yang2025efficientvlatrainingfreeaccelerationcompression}
Y.~Yang, Y.~Wang, Z.~Wen \emph{et~al.}, ``Efficientvla: training-free acceleration and compression for vision-language-action models,'' \emph{arXiv:2506.10100}, 2025.

\bibitem{xu2025vlacache}
S.~Xu, Y.~Wang, C.~Xia \emph{et~al.}, ``Vla-cache: towards efficient vision-language-action model via adaptive token caching in robotic manipulation,'' \emph{arXiv:2502.02175}, 2025.

\bibitem{li2025sp}
Y.~Li, Y.~Meng, Z.~Sun \emph{et~al.}, ``Sp-vla: a joint model scheduling and token pruning approach for vla model acceleration,'' \emph{arXiv:2506.12723}, 2025.

\bibitem{ji2025bac}
K.~Ji, Y.~Meng, H.~Cui \emph{et~al.}, ``Block-wise adaptive caching for accelerating diffusion policy,'' \emph{arXiv:2506.13456}, 2025.

\bibitem{bu2025learning}
Q.~Bu, Y.~Yang, J.~Cai \emph{et~al.}, ``Univla: learning to act anywhere with task-centric latent actions,'' \emph{arXiv:2505.06111}, 2025.

\bibitem{ye2024latent}
S.~Ye, J.~Jang, B.~Jeon \emph{et~al.}, ``Latent action pretraining from videos,'' in \emph{ICLR}, 2025.

\bibitem{he2024learningactionablediscretediffusion}
H.~He, C.~Bai, L.~Pan \emph{et~al.}, ``Learning an actionable discrete diffusion policy via large-scale actionless video pre-training,'' in \emph{NeurIPS}, 2024, pp. 31\,124--31\,153.

\bibitem{zhen20243dvla3dvisionlanguageactiongenerative}
H.~Zhen, X.~Qiu, P.~Chen \emph{et~al.}, ``3d-vla: a 3d vision-language-action generative world model,'' in \emph{ICML}, 2024, pp. 61\,229--61\,245.

\bibitem{ding2025humanoid}
P.~Ding, J.~Ma, X.~Tong \emph{et~al.}, ``Humanoid-vla: towards universal humanoid control with visual integration,'' \emph{arXiv:2502.14795}, 2025.

\bibitem{patel2025robotic}
S.~Patel, S.~Mohan, H.~Mai \emph{et~al.}, ``Robotic manipulation by imitating generated videos without physical demonstrations,'' \emph{arXiv:2507.00990}, 2025.

\bibitem{wen2024foundationpose}
B.~Wen, W.~Yang, J.~Kautz \emph{et~al.}, ``Foundationpose: unified 6d pose estimation and tracking of novel objects,'' in \emph{CVPR}, 2024, pp. 17\,868--17\,879.

\bibitem{assran2025v}
M.~Assran, A.~Bardes, D.~Fan \emph{et~al.}, ``V-jepa 2: self-supervised video models enable understanding, prediction and planning,'' \emph{arXiv:2506.09985}, 2025.

\bibitem{kostrikov2021offline}
I.~Kostrikov, A.~Nair, and S.~Levine, ``Offline reinforcement learning with implicit q-learning,'' in \emph{ICLR}, 2022.

\bibitem{haarnoja2018soft}
T.~Haarnoja, A.~Zhou, P.~Abbeel \emph{et~al.}, ``Soft actor-critic: off-policy maximum entropy deep reinforcement learning with a stochastic actor,'' in \emph{ICML}, 2018, pp. 1861--1870.

\bibitem{tan2025interactive}
S.~Tan, K.~Dou, Y.~Zhao \emph{et~al.}, ``Interactive post-training for vision-language-action models,'' \emph{arXiv:2505.17016}, 2025.

\bibitem{wang2024omnijarvisunifiedvisionlanguageactiontokenization}
Z.~Wang, S.~Cai, Z.~Mu \emph{et~al.}, ``Omni{jarvis}: unified vision-language-action tokenization enables open-world instruction following agents,'' in \emph{NeurIPS}, 2024.

\bibitem{liu2023libero}
B.~Liu, Y.~Zhu, C.~Gao \emph{et~al.}, ``Libero: benchmarking knowledge transfer for lifelong robot learning,'' in \emph{NeurIPS}, vol.~36, 2023, pp. 44\,776--44\,791.

\bibitem{fei25libero-plus}
S.~Fei, S.~Wang, J.~Shi \emph{et~al.}, ``Libero-plus: In-depth robustness analysis of vision-language-action models,'' \emph{arXiv:2510.13626}, 2025.

\bibitem{mees2022calvin}
O.~Mees, L.~Hermann, E.~Rosete-Beas \emph{et~al.}, ``Calvin: a benchmark for language-conditioned policy learning for long-horizon robot manipulation tasks,'' \emph{RA-L}, vol.~7, pp. 7327--7334, 2022.

\bibitem{ha2023scaling}
H.~Ha, H.~Ha, P.~Florence \emph{et~al.}, ``Scaling up and distilling down: Language-guided robot skill acquisition,'' in \emph{Conference on Robot Learning}.\hskip 1em plus 0.5em minus 0.4em\relax PMLR, 2023, pp. 3766--3777.

\bibitem{wen2023atm}
C.~Wen, X.~Lin, J.~So \emph{et~al.}, ``Any-point trajectory modeling for policy learning,'' \emph{arXiv:2401.00025}, 2023.

\bibitem{zhao2023act}
T.~Z. Zhao, V.~Kumar, S.~Levine \emph{et~al.}, ``Learning fine-grained bimanual manipulation with low-cost hardware,'' \emph{arXiv:2304.13705}, 2023.

\bibitem{chi2023diffusion}
C.~Chi, Z.~Xu, S.~Feng \emph{et~al.}, ``Diffusion policy: visuomotor policy learning via action diffusion,'' \emph{IJRR}, 2023.

\bibitem{lee2024behavior}
S.~Lee, Y.~Wang, H.~Etukuru \emph{et~al.}, ``Behavior generation with latent actions,'' in \emph{ICML}, 2024, pp. 26\,991--27\,008.

\bibitem{team2024octo}
O.~M. Team, D.~Ghosh, H.~Walke \emph{et~al.}, ``Octo: an open-source generalist robot policy,'' \emph{arXiv:2405.12213}, 2024.

\bibitem{black2025pi0}
K.~Black, N.~Brown, J.~Darpinian \emph{et~al.}, ``$\pi$0. 5: a vision-language-action model with open-world generalization,'' \emph{arXiv:2504.16054}, 2025.

\bibitem{Bc-z}
E.~Jang, A.~Irpan, M.~Khansari \emph{et~al.}, ``Bc-z: zero-shot task generalization with robotic imitation learning,'' in \emph{CoRL}, 2022, pp. 991--1002.

\bibitem{ebert2021bridge}
F.~Ebert, Y.~Yang, K.~Schmeckpeper \emph{et~al.}, ``Bridge data: boosting generalization of robotic skills with cross-domain datasets,'' in \emph{RSS}, 2022, pp. 1--7.

\bibitem{fang2023rh20t}
H.-S. Fang, H.~Fang, Z.~Tang \emph{et~al.}, ``Rh20t: a comprehensive robotic dataset for learning diverse skills in one-shot,'' in \emph{ICRA}, 2024, pp. 653--660.

\bibitem{khazatsky2024droid}
A.~Khazatsky, K.~Pertsch, S.~Nair \emph{et~al.}, ``Droid: a large-scale in-the-wild robot manipulation dataset,'' in \emph{RSS}, 2024, pp. 1--26.

\bibitem{li2024behavior}
C.~Li, R.~Zhang, J.~Wong \emph{et~al.}, ``Behavior-1k: a human-centered, embodied ai benchmark with 1,000 everyday activities and realistic simulation,'' in \emph{CoRL}, 2022.

\bibitem{shridhar2020alfred}
M.~Shridhar, J.~Thomason, D.~Gordon \emph{et~al.}, ``Alfred: a benchmark for interpreting grounded instructions for everyday tasks,'' in \emph{CVPR}, 2020, pp. 10\,740--10\,749.

\bibitem{james2020rlbench}
S.~James, Z.~Ma, D.~R. Arrojo \emph{et~al.}, ``Rlbench: the robot learning benchmark \& learning environment,'' \emph{RA-L}, vol.~5, pp. 3019--3026, 2020.

\bibitem{grotz2024peract2}
M.~Grotz, M.~Shridhar, T.~Asfour \emph{et~al.}, ``Peract2: benchmarking and learning for robotic bimanual manipulation tasks,'' \emph{arXiv:2407.00278}, 2024.

\bibitem{yu2020meta}
T.~Yu, D.~Quillen, Z.~He \emph{et~al.}, ``Meta-world: a benchmark and evaluation for multi-task and meta reinforcement learning,'' in \emph{CoRL}, 2020, pp. 1094--1100.

\bibitem{FrankaKitchen}
A.~Gupta, V.~Kumar, C.~Lynch \emph{et~al.}, ``Relay policy learning: solving long-horizon tasks via imitation and reinforcement learning,'' in \emph{CoRL}, 2019.

\bibitem{cherepanov2025mikasa}
E.~Cherepanov, N.~Kachaev, A.~K. Kovalev \emph{et~al.}, ``Memory, benchmark \& robots: a benchmark for solving complex tasks with reinforcement learning,'' \emph{arXiv:2502.10550}, 2025.

\bibitem{li24simpler}
X.~Li, K.~Hsu, J.~Gu \emph{et~al.}, ``Evaluating real-world robot manipulation policies in simulation,'' in \emph{CoRL}, 2024.

\bibitem{savva2019habitat}
M.~Savva, A.~Kadian, O.~Maksymets \emph{et~al.}, ``Habitat: a platform for embodied ai research,'' in \emph{ICCV}, 2019, pp. 9339--9347.

\bibitem{szot2021habitat}
A.~Szot, A.~Clegg, E.~Undersander \emph{et~al.}, ``Habitat 2.0: training home assistants to rearrange their habitat,'' in \emph{NeurIPS}, vol.~34, 2021, pp. 251--266.

\bibitem{puig2023habitat}
X.~Puig, E.~Undersander, A.~Szot \emph{et~al.}, ``Habitat 3.0: a co-habitat for humans, avatars and robots,'' \emph{ICLR}, 2024.

\bibitem{xiang2020sapien}
F.~Xiang, Y.~Qin, K.~Mo \emph{et~al.}, ``Sapien: a simulated part-based interactive environment,'' in \emph{CVPR}, 2020, pp. 11\,097--11\,107.

\bibitem{pumacay2024colosseum}
W.~Pumacay, I.~Singh, J.~Duan \emph{et~al.}, ``The colosseum: a benchmark for evaluating generalization for robotic manipulation,'' in \emph{RSS}, 2024.

\bibitem{zhang2024vlabench}
S.~Zhang, Z.~Xu, P.~Liu \emph{et~al.}, ``Vlabench: a large-scale benchmark for language-conditioned robotics manipulation with long-horizon reasoning tasks,'' in \emph{ICCV}, 2025.

\bibitem{kris2021ego}
K.~Grauman, A.~Westbury, E.~Byrne \emph{et~al.}, ``Ego4d: around the world in 3,000 hours of egocentric video,'' in \emph{CVPR}, 2022.

\bibitem{grauman2024ego}
K.~Grauman, A.~Westbury, L.~Torresani \emph{et~al.}, ``Ego-exo4d: understanding skilled human activity from first-and third-person perspectives,'' in \emph{CVPR}, 2024, pp. 19\,383--19\,400.

\bibitem{chen2023egoplan}
Y.~Chen, Y.~Ge, Y.~Ge \emph{et~al.}, ``Egoplan-bench: benchmarking egocentric embodied planning with multimodal large language models,'' \emph{arxiv:2312.06722}, 2024.

\bibitem{wang2024egovid}
X.~Wang, K.~Zhao, F.~Liu \emph{et~al.}, ``Egovid-5m: a large-scale video-action dataset for egocentric video generation,'' \emph{arXiv:2411.08380}, 2024.

\bibitem{dima2018epic}
D.~Damen, H.~Doughty, G.~M. Farinella \emph{et~al.}, ``Scaling egocentric vision: the epic-kitchens dataset,'' in \emph{ECCV}, 2018.

\bibitem{maeda2024kitchens}
K.~Maeda, T.~Hirasawa, A.~Hashimoto \emph{et~al.}, ``Com kitchens: an unedited overhead-view video dataset as a vision-language benchmark,'' in \emph{ECCV}, 2024, pp. 123--140.

\bibitem{fan2019egovqa}
C.~Fan, ``Egovqa-an egocentric video question answering benchmark dataset,'' in \emph{ICCV}, 2019.

\bibitem{jia2022egotaskqa}
B.~Jia, T.~Lei, S.-C. Zhu \emph{et~al.}, ``Egotaskqa: understanding human tasks in egocentric videos,'' in \emph{NeurIPS}, vol.~35, 2022, pp. 3343--3360.

\bibitem{hoque2025egodex}
R.~Hoque, P.~Huang, D.~J. Yoon \emph{et~al.}, ``Egodex: learning dexterous manipulation from large-scale egocentric video,'' \emph{arXiv:2505.11709}, 2025.

\bibitem{wang2024dexcap}
C.~Wang, H.~Shi, W.~Wang \emph{et~al.}, ``Dexcap: scalable and portable mocap data collection system for dexterous manipulation,'' in \emph{RSS}, 2024.

\bibitem{kareer2024egomimic}
S.~Kareer, D.~Patel, R.~Punamiya \emph{et~al.}, ``Egomimic: scaling imitation learning via egocentric video,'' \emph{arXiv:2410.24221}, 2024.

\bibitem{PH2D}
R.-Z. Qiu, S.~Yang, X.~Cheng \emph{et~al.}, ``Humanoid policy\~{} human policy,'' \emph{arXiv:2503.13441}, 2025.

\bibitem{yu2025real2render2real}
J.~Yu, L.~Fu, H.~Huang \emph{et~al.}, ``Real2render2real: scaling robot data without dynamics simulation or robot hardware,'' \emph{arXiv:2505.09601}, 2025.

\bibitem{das2018embodied}
A.~Das, S.~Datta, G.~Gkioxari \emph{et~al.}, ``Embodied question answering,'' in \emph{CVPR}, 2018, pp. 1--10.

\bibitem{gordon2018iqa}
D.~Gordon, A.~Kembhavi, M.~Rastegari \emph{et~al.}, ``Iqa: visual question answering in interactive environments,'' in \emph{CVPR}, 2018, pp. 4089--4098.

\bibitem{MT-EQA}
L.~Yu, X.~Chen, G.~Gkioxari \emph{et~al.}, ``Multi-target embodied question answering,'' in \emph{CVPR}, 2019, pp. 6309--6318.

\bibitem{MP3D-EQA}
E.~Wijmans, S.~Datta, O.~Maksymets \emph{et~al.}, ``Embodied question answering in photorealistic environments with point cloud perception,'' in \emph{CVPR}, 2019, pp. 6659--6668.

\bibitem{islam2023eqa}
M.~M. Islam, A.~Gladstone, R.~Islam \emph{et~al.}, ``Eqa-mx: embodied question answering using multimodal expression,'' in \emph{ICLR}, 2023.

\bibitem{majumdar2024openeqa}
A.~Majumdar, A.~Ajay, X.~Zhang \emph{et~al.}, ``Openeqa: embodied question answering in the era of foundation models,'' in \emph{CVPR}, 2024, pp. 16\,488--16\,498.

\bibitem{choi2024lota}
J.-W. Choi, Y.~Yoon, H.~Ong \emph{et~al.}, ``Lota-bench: benchmarking language-oriented task planners for embodied agents,'' in \emph{ICLR}, 2024.

\end{thebibliography}
\endgroup

\IEEEdisplaynontitleabstractindextext

\IEEEpeerreviewmaketitle

\appendices

\section{{ Experimental Evaluation}}
\label{Experimental}
{In this section, we present a rigorous comparative analysis of state of the art large VLM based vision language action (VLA) models. To comprehensively assess these architectures, we synthesize quantitative results from three primary simulation benchmarks alongside diverse real world deployments. Specifically, we utilize LIBERO~\cite{liu2023libero} to evaluate lifelong learning and long horizon capabilities. We then employ LIBERO-Plus~\cite{fei25libero-plus} and CALVIN~\cite{mees2022calvin} to deeply explore model generalization under severe visual and kinematic distribution shifts. Beyond simulations, we systematically investigate model performance on real world custom tasks to examine practical capabilities across varied robotic hardware. Finally, drawing on these empirical findings, we discuss the inherent tradeoffs between monolithic and hierarchical paradigms in physical applications.}

\subsection{{Benchmark Performance Analysis}}
{To ensure a standardized comparison, we employ Success Rate (SR) as our primary Evaluation Metric. SR is defined as the percentage of episodes in which the robot successfully completes the target task within a predefined maximum step limit following a natural language instruction. For multi-task evaluations, we calculate the average SR across all tasks within a given suite to quantify the model's overall robustness and generalization proficiency.}

\subsubsection{{Evaluation on Diverse Tasks}}
\begin{table}[htbp]
\caption{{Comprehensive LIBERO Benchmark Results.}}
\label{tab:libero}
\centering
\footnotesize
\setlength{\tabcolsep}{4pt}
\begin{tabular}{l|cccc|c}
\toprule
\textbf{Method} & \textbf{Spatial} & \textbf{Object} & \textbf{Goal} & \textbf{Long} & \textbf{Overall} \\ 
    \midrule
    \rowcolor{lightgray}
    \multicolumn{6}{c}{\textbf{Early Language-Conditioned Robotic System}} \\ 
    \midrule
    Distill-D~\cite{ha2023scaling} & 46.8 & 72.0 & 63.8 & 47.3 & 57.5 \\
    Transformer-BC~\cite{liu2023libero} & 71.8 & 71.0 & 76.3 & 24.2 & 60.8 \\
    ResNet-T~\cite{liu2023libero} & 75.7 & 78.9 & 52.7 & 45.0 & 63.1 \\
    ATM~\cite{wen2023atm} & 68.5 & 68.0 & 77.8 & 39.3 & 63.4 \\
    ACT~\cite{zhao2023act} & 82.0 & 78.8 & 66.1 & 44.0 & 67.7 \\
    Diffusion Policy~\cite{chi2023diffusion} & 78.3 & 92.5 & 68.3 & 50.5 & 72.4 \\
    VQ-BeT~\cite{lee2024behavior} & 88.7 & 90.3 & 61.3 & 59.7 & 75.0 \\
    \midrule
    \rowcolor{lightgray}
    \multicolumn{6}{c}{\textbf{Monolithic VLA models: Single-system Model}} \\ 
    \midrule
    FlashVLA~\cite{tan2025thinktwiceactonce} & 84.2 & 86.4 & 75.4 & 51.4 & 74.4 \\
    TraceVLA~\cite{zheng2024tracevla} & 84.6 & 85.2 & 75.1 & 54.1 & 74.8 \\
    Octo~\cite{team2024octo} & 78.9 & 85.7 & 84.6 & 51.1 & 75.1 \\
    Spec-VLA~\cite{wang2025spec} & 85.8 & 85.0 & 74.7 & 55.0 & 75.1 \\
    OpenVLA~\cite{kim2024openvla} & 84.7 & 88.4 & 79.2 & 53.7 & 76.5 \\
    UniAct~\cite{zheng2025universal} & 77.0 & 87.0 & 77.0 & 70.0 & 77.8 \\
    SpatialVLA~\cite{qu2025spatialvlaexploringspatialrepresentations} & 88.2 & 89.9 & 78.6 & 55.5 & 78.1 \\
    WorldVLA~\cite{cen2025worldvla} & 87.6 & 96.2 & 83.4 & 60.0 & 81.8 \\
    CoT-VLA~\cite{cotVLA} & 87.5 & 91.6 & 87.6 & 69.0 & 83.9 \\
    NORA~\cite{hung2025nora} & 92.2 & 95.4 & 89.4 & 74.6 & 87.9 \\
    4D-VLA~\cite{zhang20254dvla} & 88.9 & 95.2 & 90.9 & 79.1 & 88.6 \\
    PD-VLA~\cite{pdvla} & 95.5 & 96.7 & 94.9 & 91.7 & 94.7 \\
    VOTE~\cite{lin2025vote} & 96.0 & 98.5 & 94.0 & 91.0 & 94.9 \\
    UnifiedVLA~\cite{wang2025unified} & 95.4 & 98.8 & 93.6 & 94.0 & 95.5 \\
    BitVLA~\cite{wang2025bitvla} & 96.6 & 99.0 & 95.4 & 92.8 & 96.0 \\ 
    OpenVLA-OFT~\cite{kim2025oft} & 97.6 & 98.4 & 97.9 & 94.5 & 97.1 \\

    \midrule
    \rowcolor{lightgray}
    \multicolumn{6}{c}{\textbf{Monolithic VLA models: Dual-system Model}} \\ 
    \midrule
    GraspVLA~\cite{deng25gra} & - & 94.1 & 91.2 & 82.0 & - \\
    $\pi_0$ + FAST~\cite{pertsch2025fast} & 96.4 & 96.8 & 88.6 & 60.2 & 85.5 \\
    TriVLA~\cite{liu2025trivla} & 91.2 & 93.8 & 89.8 & 73.2 & 87.0 \\
    SmolVLA~\cite{smolvla} & 93.0 & 94.0 & 91.0 & 77.0 & 88.8 \\
    $\pi_0$~\cite{black2024pi0} & 96.8 & 98.8 & 95.8 & 85.2 & 94.2 \\
    villa-X~\cite{chen2025villa} & 97.5 & 97.0 & 91.5 & 74.5 & 90.1 \\
    \midrule
    \rowcolor{lightgray}
    \multicolumn{6}{c}{\textbf{Hierarchical VLA models}} \\ 
    \midrule
    DexVLA~\cite{wen2025dexvla} & 97.2 & 99.1 & 95.6 & - & - \\
    Agentic Robot~\cite{yang2025agentic} & 85.8 & 89.0 & 81.8 & 61.6 & 79.6 \\
    \midrule
    \rowcolor{lightgray}
    \multicolumn{6}{c}{\textbf{Other Advanced Field}} \\ 
    \midrule
    LAPA~\cite{ye2024latent} & 73.8 & 74.6 & 58.8 & 55.4 & 65.7 \\
    VLA-Cache~\cite{xu2025vlacache} & 83.8 & 85.8 & 76.4 & 52.8 & 74.7 \\
    VLA-RL~\cite{lu2025vlarl} & 90.2 & 91.8 & 82.2 & 59.8 & 81.0 \\
    WorldVLA~\cite{cen2025worldvla} & 87.6 & 96.2 & 83.4 & 60.0 & 81.8 \\
    UniVLA~\cite{bu2025learning} & 96.5 & 96.8 & 95.6 & 92.0 & 95.2 \\
\bottomrule
\end{tabular}
\end{table}

{\textbf{The Capability Leap of Monolithic Scaling.} Table \ref{tab:libero} presents the evaluation of various paradigms on the LIBERO benchmark~\cite{liu2023libero}, spanning Spatial, Object, Goal, and Long-horizon tasks. A prominent phenomenon is the massive capability leap from Early Language-Conditioned Systems to Monolithic VLA models. Early models, relying on smaller backbones or decoupled feature extractors (e.g., ACT~\cite{zhao2023act}, Diffusion Policy~\cite{chi2023diffusion}), plateau around a 65\% to 75\% overall success rate. In stark contrast, Monolithic Single-system models powered by large VLMs demonstrate exceptional scalability, with state-of-the-art models like OpenVLA-OFT~\cite{kim2025oft}, UniVLA~\cite{bu2025learning}, and UnifiedVLA~\cite{wang2025unified} achieving overall success rates exceeding 95\%. This clearly indicates that leveraging massive pretrained VLMs as unified backbones fundamentally shatters the performance ceiling of early decoupled systems, establishing monolithic scaling as the primary driver for broad tasks.}

{\textbf{Overcoming Long Horizon Bottlenecks.} Crucially, the ``Long'' task category highlights a significant bottleneck for earlier architectures. While models like Transformer-BC~\cite{liu2023libero} and ACT~\cite{zhao2023act} achieve high performance on spatial and object tasks, their performance drops to roughly 20\% to 35\% on long-horizon reasoning. This exposes a fundamental limitation in naive autoregressive decoding over long temporal windows, where context saturation and error accumulation rapidly degrade performance. Conversely, recent single-system advancements (e.g., PD-VLA~\cite{pdvla}, VOTE~\cite{lin2025vote}) and dual-system models (e.g., $\pi_0$~\cite{black2024pi0}) explicitly address this through parallel decoding mechanisms, flow-matching, or continuous latent feature propagation, effectively pushing long-horizon success rates above 90\%. This provides a deep insight: while simply scaling VLM parameters effectively solves basic spatial semantic grounding, sustained long horizon execution strictly requires architectural innovations that capture temporal dynamics and tightly integrate high level cognitive reasoning with robust continuous control.}

\subsubsection{{Evaluation on Generalization Performance}}

\begin{table*}[htbp]
\caption{{Comprehensive LIBERO-Plus Benchmark Results under Various Perturbations. $\downarrow$ denotes the relative decrease from the original success rate.}}
\label{tab:libero-plus}
\centering
\footnotesize
\setlength{\tabcolsep}{3.5pt}
\begin{tabular}{l|c|ccccccc|c}
\toprule
\textbf{Method} & \textbf{Original} & \textbf{Camera} & \textbf{Robot} & \textbf{Language} & \textbf{Light} & \textbf{ Background} & \textbf{Noise} & \textbf{Layout} & \textbf{Overall} \\ 
    \midrule
    OpenVLA~\cite{kim2024openvla} & 76.5 & 1.1 (75.4 $\downarrow$) & 4.1 (72.4 $\downarrow$) & 26.8 (49.7 $\downarrow$) & 4.4 (72.1 $\downarrow$) & 25.3 (51.2 $\downarrow$) & 19.3 (57.2 $\downarrow$) & 31.6 (44.9 $\downarrow$) & 16.1 (60.4 $\downarrow$) \\
    OpenVLA-OFT~\cite{kim2025oft} & 97.1 & 59.7 (37.4 $\downarrow$) & 37.2 (59.9 $\downarrow$) & 81.5 (15.6 $\downarrow$) & 85.8 (11.3 $\downarrow$) & 92.4 (4.7 $\downarrow$) & 76.7 (20.4 $\downarrow$) & 77.1 (20.0 $\downarrow$) & 72.9 (24.2 $\downarrow$) \\
    $\pi_0$~\cite{black2024pi0} & 94.2 & 15.8 (78.4 $\downarrow$) & 6.6 (87.6 $\downarrow$) & 61.0 (33.2 $\downarrow$) & 79.6 (14.6 $\downarrow$) & 78.5 (15.7 $\downarrow$) & 79.4 (14.8 $\downarrow$) & 70.4 (23.8 $\downarrow$) & 55.9 (38.3 $\downarrow$) \\
    $\pi_0$ + FAST~\cite{pertsch2025fast} & 85.5 & 66.4 (19.1 $\downarrow$) & 24.8 (60.7 $\downarrow$) & 63.3 (22.2 $\downarrow$) & 73.0 (12.5 $\downarrow$) & 67.7 (17.8 $\downarrow$) & 75.8 (9.7 $\downarrow$) & 70.3 (15.2 $\downarrow$) & 63.0 (22.5 $\downarrow$) \\
    NORA~\cite{hung2025nora} & 87.9 & 4.0 (83.9 $\downarrow$) & 41.1 (46.8 $\downarrow$) & 67.0 (20.9 $\downarrow$) & 31.0 (56.9 $\downarrow$) & 50.5 (37.4 $\downarrow$) & 17.6 (70.3 $\downarrow$) & 63.9 (24.0 $\downarrow$) & 39.3 (48.6 $\downarrow$) \\
    WorldVLA~\cite{cen2025worldvla} & 79.1 & 0.3 (78.8 $\downarrow$) & 30.2 (48.9 $\downarrow$) & 44.2 (34.9 $\downarrow$) & 29.4 (49.7 $\downarrow$) & 14.5 (64.6 $\downarrow$) & 12.2 (66.9 $\downarrow$) & 39.4 (39.7 $\downarrow$) & 24.3 (54.8 $\downarrow$) \\
    UniVLA~\cite{bu2025learning} & 95.2 & 4.3 (90.9 $\downarrow$) & 50.3 (44.9 $\downarrow$) & 71.8 (23.4 $\downarrow$) & 59.1 (36.1 $\downarrow$) & 80.0 (15.2 $\downarrow$) & 25.3 (69.9 $\downarrow$) & 34.3 (60.9 $\downarrow$) & 46.4 (48.8 $\downarrow$) \\
\bottomrule
\end{tabular}
\end{table*}

{\textbf{Visual Kinematic Overfitting and Perceptual Brittleness.} The LIBERO-Plus~\cite{fei25libero-plus} results in Table \ref{tab:libero-plus} detail performance under a comprehensive suite of perturbations. A detailed analysis of these results exposes a multifaceted overfitting phenomenon. For instance, OpenVLA~\cite{kim2024openvla} suffers a catastrophic overall performance drop to 16.1\%. The near total collapse under camera viewpoint shifts (1.1\%) and robot initial state perturbations (4.1\%) confirms a severe visual kinematic entanglement. These models heavily memorize specific 2D pixel to joint mappings rather than acquiring an embodiment agnostic understanding of physical mechanics or invariant spatial geometry. Even the state of the art OpenVLA-OFT~\cite{kim2025oft} experiences a massive performance drop when facing unseen camera angles. Furthermore, substantial degradation under varying light conditions, background textures, and sensor noise indicates a critical overreliance on pristine high frequency visual features. Instead of learning robust geometric object affordances, current architectures overfit to the specific rendering pipelines and photometric properties of their training environments.}

{\textbf{Cognitive Limitations and Semantic Fragility.} Beyond perceptual brittleness, the performance drops associated with language instruction rewriting and object layout alterations reveal profound cognitive limitations. When task instructions are infused with conversational distractors, commonsense synonyms, or complex reasoning chains, models exhibit significant capability regression. This demonstrates that the integrated VLMs often function as shallow linguistic pattern matchers rather than exhibiting true deep semantic reasoning or robust multimodal information filtering. Additionally, the introduction of confounding distractor objects or shifted target poses severely disrupts visual attention, proving that current architectures struggle to maintain stable object permanence and relational focus in cluttered scenes. Dual system models like $\pi_0$~\cite{black2024pi0} and $\pi_0$ + FAST~\cite{chen2025fast} exhibit similar vulnerabilities across these dimensions, proving that merely decoupling the action expert does not inherently resolve these foundational geometric and semantic flaws.}

{\textbf{Cross Domain Resilience and Robust Action Generation.} As shown in Table~\ref{tab:calvin}, the ABC$\rightarrow$D split represents the most difficult evaluation setting within CALVIN benchmark~\cite{mees2022calvin}, forcing methods to generalize to unseen scene (Environment D) after training exclusively on three distinct environments (A, B, and C). In this rigorous cross domain and long horizon setting, the sequential nature of the 1/5 to 5/5 tasks severely tests a model's resistance to compounding covariate shift. Early architectures like RT-1~\cite{brohan2022rt} suffer immediate collapse, dropping from a 53.3\% success rate at the first step to a mere 1.3\% at the fifth step. However, advanced monolithic and dual system architectures, such as TriVLA~\cite{liu2025trivla}, UnifiedVLA~\cite{wang2025unified}, and $\pi_0$~\cite{black2024pi0}, demonstrate remarkable resilience by maintaining high success rates even at the final 5/5 stage. We attribute this success to the integration of robust action generation paradigms, specifically diffusion and flow matching mechanisms, which provide dynamic action correction capabilities. This strongly implies that tightly coupling the deep semantic planning of a VLM with an expressive and noise resistant low level action expert is currently the most viable path toward achieving generalized long horizon robotic autonomy in unseen environments.}

\begin{table}[htbp]
\caption{{Comprehensive CALVIN Benchmark Results under ABC$\rightarrow$D Setting.}}
\label{tab:calvin}
\centering
\footnotesize
\setlength{\tabcolsep}{5pt} 
\begin{tabular}{l|ccccc|c} 
\toprule
\textbf{Method} & \textbf{1/5} & \textbf{2/5} & \textbf{3/5} & \textbf{4/5} & \textbf{5/5} & \textbf{Avg. Len.} \\
    \midrule
    RT-1~\cite{brohan2022rt} & 53.3 & 22.2 & 9.4 & 3.8 & 1.3 & 0.90 \\
    LCB~\cite{shentu2024llms} & 73.6 & 50.2 & 28.5 & 16.0 & 9.9 & 1.78 \\
    RationalVLA~\cite{song2025rationalvla} & 74.3 & 58.3 & 42.3 & 30.0 & 20.7 & 2.26 \\
    VLAS~\cite{zhao2025vlas} & 87.2 & 64.2 & 40.9 & 28.1 & 19.6 & 2.40 \\
    RoboFlamingo~\cite{li2023vision} & 82.4 & 61.9 & 46.6 & 33.1 & 23.5 & 2.47 \\
    DeeR-VLA~\cite{deervla} & 84.8 & 72.3 & 54.9 & 44.6 & 33.5 & 2.90 \\
    OE-VLA~\cite{zhao2025unveilingpotentialvisionlanguageactionmodels} & 91.8 & 73.8 & 56.2 & 44.2 & 33.4 & 2.99 \\
    OpenVLA~\cite{kim2024openvla} & 91.3 & 77.8 & 62.0 & 52.1 & 43.5 & 3.27 \\
    RoboDual~\cite{bu2024towards} & 94.4 & 82.7 & 72.1 & 62.4 & 54.4 & 3.66 \\
    UniVLA~\cite{bu2025learning} & 95.5 & 85.8 & 75.4 & 66.9 & 56.5 & 3.80 \\
    $\pi_0$~\cite{black2024pi0} & 93.8 & 85.0 & 76.7 & 68.1 & 59.9 & 3.92 \\
    UP-VLA~\cite{zhang2025up} & 92.8 & 86.5 & 81.5 & 76.9 & 69.9 & 4.08 \\
    TriVLA~\cite{liu2025trivla} & 96.8 & 92.4 & 86.8 & 83.2 & 81.8 & 4.37 \\
    UnifiedVLA~\cite{wang2025unified} & 98.9 & 94.8 & 89.0 & 82.8 & 75.1 & 4.41 \\
\bottomrule
\end{tabular}
\end{table}

\subsubsection{{Evaluation on Real-World Custom Tasks}}
\begin{table*}[htbp]
\caption{{Analysis of Real-world Experiments. $\uparrow$ denotes the improvement in success rate compared to the baseline method under the same task settings.}}
\label{tab:real_exp}
\centering
\scriptsize
\renewcommand{\arraystretch}{1.2}
\begin{tabularx}{\textwidth}{C{2.1cm} C{0.7cm} C{1.4cm} C{1.6cm} X}
    \toprule
    \makecell[c]{\textbf{Model} \\[0.4ex]}
    & \makecell[c]{\textbf{Robot} \\ [0.4ex]} 
    & \makecell[c]{\textbf{Baseline} \\ [0.4ex]} 
    & \makecell[c]{\textbf{SR$(\%)$} \\[0.4ex]}  
    & \makecell[c]{\textbf{Real-World Task Characteristics} \\[0.4ex]} \\ \midrule
    \rowcolor{lightgray}
    \multicolumn{5}{c}{\textbf{MONOLITHIC}} \\ \midrule

    RT-2 \cite{brohan2023rt2visionlanguageactionmodelstransfer}
    & -
    & RT-1
    & $71.3(24.3\uparrow)$
    & Custom zero-shot tasks assessing generalization and emergent capabilities. \\

    RT-2-X \cite{o2024open}
    & WidowX
    & RT-1
    & $40(5\uparrow)$
    & Using diverse test environments and robot arms to assess cross-entity transfer. \\

    OpenVLA 
    \cite{kim2024openvla}
    & Panda
    & DP
    & $61.1(16.8\uparrow)$
    & Custom tasks with out-of-distribution target objects and distractors. \\

    ECoT \cite{chen2025trainingstrategiesefficientembodied}
    & WidowX
    & OpenVLA
    & $65(28\uparrow)$
    & Custom spatial-relation and out-of-distribution tasks assessing generalization. \\

    TraceVLA
    \cite{zheng2024tracevla}
    & WidowX
    & OpenVLA
    & $75(57.5\uparrow)$
    & Custom tasks with randomized initial positions, unseen tasks, and distractor objects. \\

    FuSe \cite{jones2025beyond}
    & WidowX
    & Octo
    & $68.7(22.7\uparrow)$
    & Custom tasks assessing joint visual, auditory, and tactile reasoning. \\

    UniAct
    \cite{zheng2025universal}
    & UnitreeG1
    & LangWBC
    & $83.1(24.9\uparrow)$
    & Custom tasks assessing heterogeneous instruction understanding. \\

    SpatialVLA \cite{qu2025spatialvlaexploringspatialrepresentations}
    & WidowX
    & OpenVLA
    & $74.0(9\uparrow)$
    & Custom unseen-background and pose tasks assessing semantic and motor robustness. \\

    UP-VLA
    \cite{zhang2025up}
    & Panda
    & DP
    & $65(51.7\uparrow)$
    & Custom unseen and precision tasks assessing generalization and perception. \\

    HybridVLA \cite{hybridvla}
    & FR3
    & CogACT
    & $83(22\uparrow)$
    & Custom tasks assessing spatial understanding and long-horizon capabilities. \\

    CoT-VLA \cite{cotVLA}
    & Panda
    & OpenVLA
    & $78.8(11.5\uparrow)$
    & Custom long-horizon tasks assessing out-of-distribution generalization. \\

    VTLA \cite{zhang2025vtla}
    & UR3 
    & Qwen2-VL
    & $97.5(2.5\uparrow)$
    & Custom peg-in-hole assembly tasks assessing contact-rich manipulation. \\

    BridgeVLA
    \cite{li2025bridgevla}
    & FR3
    & RVT-2
    & $96.9(6.9\uparrow)$
    & Custom basic and generalization evaluation settings. \\

    4D-VLA \cite{zhang20254dvla}
    & Franka
    & OpenVLA
    & $85.6(57.9\uparrow)$
    & Custom out-of-distribution and novel-view generalization tasks. \\

    ST-VLA \cite{patratskiy2025stvla}
    & Panda
    & OpenVLA
    & $72.5(32.5\uparrow)$
    & Custom tasks assessing generalization, robustness, and long-horizon capabilities. \\

    OFT \cite{kim2025oft}
    & ViperX
    & RDT-1B
    & $87.8(9.4\uparrow)$
    & Custom unseen dexterous bimanual manipulation tasks. \\

    PD-VLA \cite{pdvla}
    & UnitreeZ1
    & LLaVA-VLA
    & $70(33.3\uparrow)$
    & Custom distractor-augmented tasks assessing action coherence. \\

    MoLe-VLA
    \cite{zhang2025mole}
    & FR3
    & CogAct
    & $70(3.3\uparrow)$
    & Custom tasks assessing precise 3D spatial understanding and continuous control. \\

    NORA \cite{hung2025nora}
    & WidowX
    & OpenVLA
    & $56.7(16.7\uparrow)$
    & Custom tasks assessing out-of-distribution generalization and multi-target grasping. \\

    BitVLA \cite{wang2025bitvla}
    & Franka
    & $\pi_0$
    & $65(12.5\uparrow)$
    & Custom basic and OOD tasks with unseen objects and background substitutions. \\
    
    Tactile-VLA \cite{huang2025tactile}
    & -
    & $\pi_0$
    & $90(50\uparrow)$
    & Custom contact-rich manipulation tasks assessing language-driven force reasoning. \\
    
    GR-3 \cite{cheang2025gr3}
    & ByteMini
    & $\pi_0$
    & $57.8(17.8\uparrow)$
    & Custom tabletop tasks assessing generalization across unseen scenarios. \\
    
    TinyVLA \cite{wen2025tinyvla}
    & Franka
    & OpenVLA 
    & $94.0(25.7\uparrow)$
    & Custom single-arm and bimanual tasks with diverse action spaces. \\
    
    RoboDual \cite{bu2024towards}
    & ALOHA
    & OpenVLA 
    & $82.7(26.7\uparrow)$
    & Custom tasks comparing specialist, generalist and hybrid VLA approaches. \\
    
    GR00T N1 \cite{bjorck2025gr00t}
    & GR-1
    & DP
    & $76.8(30.4\uparrow)$
    & Custom tasks assessing data efficiency across varying scales of human teleoperation data. \\
    
    CogACT \cite{li2024cogact}
    & Realman
    & OpenVLA 
    & $71.2(51.9\uparrow)$
    & Custom tasks with object-color compositional reasoning. \\
    
    HiRT \cite{zhang2024hirt}
    & Franka
    & Vanilla-VLA 
    & $75.0(27.0\uparrow)$
    & Custom tasks assessing semantic grounding with unseen objects, distractors and target motion. \\
    
    Fast-in-Slow \cite{chen2025fast}
    & Agilex
    & $\pi_0$ 
    & $74(13\uparrow)$
    & Custom multi-condition tasks assessing generalization, robustness and distraction resilience. \\
    
    ChatVLA \cite{chatvla}
    & Franka
    & OpenVLA
    & $51.4(32.7\uparrow)$
    & Custom long-horizon and multi-task tasks assessing compositional skills. \\
    
    ChatVLA-2 \cite{chatvla2}
    & ARX
    & $\pi_0$ 
    & $81.4(65.4\uparrow)$
    & Custom mathematical and spatial reasoning tasks. \\
    
    DiVLA \cite{diffusionvla}
    & Agilex
    & OpenVLA
    & $71.0(44.3\uparrow)$
    & Custom multi-task and sorting tasks assessing generalization and robustness to visual changes. \\
    
    RationalVLA \cite{song2025rationalvla}
    & AgileX
    & 3DDA
    & $85.0(68.3\uparrow)$
    & Custom tasks with unseen instructions assessing language generalization. \\
    
    ForceVLA\cite{yu2025forcevla}
    & Franka
    & $\pi_0$ 
    & $60.5(20.3\uparrow)$
    & Custom contact-rich tasks assessing force control and multimodal manipulation. \\
    
    SmolVLA  \cite{smolvla}
    & Franka
    & $\pi_0$ 
    & $78.3(16.6\uparrow)$
    & Custom tasks assessing generalization and OOD robustness. \\
    
    OneTwoVLA \cite{lin2025onetwovla}
    & Franka
    & $\pi_0$ 
    & $87(30\uparrow)$
    & Custom long-horizon tasks assessing planning and generalization. \\
    
    \midrule \rowcolor{lightgray} \multicolumn{5}{c}{\textbf{HIERARCHICAL}} \\ \midrule

    Hamster \cite{li2025hamster}
    & Panda
    & OpenVLA
    & $69(32\uparrow)$ 
    & Custom tabletop tasks assessing cross-domain visual and semantic generalization. \\
    
    DexVLA \cite{wen2025dexvla}
    & DJI'RM
    & $\pi_0$
    & $55(13.5\uparrow)$ 
    & Custom long-horizon tasks assessing implicit reasoning and cross-embodiment dexterity. \\

    RoboMatrix \cite{mao2024robomatrix}
    & DJI'RM
    & OpenVLA
    & $100(100\uparrow)$ 
    & Custom long-horizon tasks assessing zero-shot skill recombination and generalization. \\

    PointVLA \cite{li2025pointvla}
    & AgileX
    & DexVLA
    & $57(53\uparrow)$ 
    & Custom 3D-aware tasks assessing spatial reasoning, height adaptability, and anti-hallucination. \\

    $A_0$ \cite{xu2025a0}
    & Gen3
    & ReKep
    & $53.75(20\uparrow)$ 
    & Custom zero-shot tasks assessing affordance points and continuous visual traces. \\

    FSD \cite{yuan2025seeing}
    & xArm6
    & RoboPoint
    & $72(30\uparrow)$ 
    & Custom zero-shot tasks evaluating affordance- and visual trace-guided manipulation. \\

    Robridge \cite{zhang2025robridge}
    & Gen3
    & ReKep
    & $83.3(34.1\uparrow)$ 
    & Custom multi-stage tasks assessing zero-shot generalization and cognition-execution bridging. \\

    DexGrasp \cite{zhong2025dexgraspvla}
    & RM+GO
    & $\pi_0$
    & $91.7(61.4\uparrow)$ 
    & Custom zero-shot tasks evaluating dexterous generalization across extreme domain shifts. \\

    RT-H \cite{belkhale2024rt}
    & EDR
    & RT-2
    & $40(15\uparrow)$ 
    & High-Precision tasks assessing contextual sequential execution and language interventions. \\

    ReKep \cite{huang2024rekep}
    & Franka
    & VoxPoser
    & $61(51\uparrow)$ 
    & Custom tasks assessing zero-shot spatio-temporal reasoning and reactive coordination. \\

    VoxPoser \cite{huang2023voxposer}
    & Panda
    & LLM+Prim
    & $88(64\uparrow)$ 
    & Custom tabletop tasks assessing zero-shot spatial composition and dynamic robustness. \\

    RT-Affo \cite{nasiriany2024rtaffordanceaffordancesversatileintermediate}
    & EDR
    & RT-2
    & $68(40\uparrow)$ 
    & Custom manipulation tasks testing fine-grained part-level and precise spatial reasoning. \\

    HiBerNAC \cite{wu2025hibernac}
    & Franka
    & Octo/RDT
    & $60(39/5.6\uparrow)$ 
    & Custom dynamic tasks for neuro-inspired long-horizon adaptation and error-correction. \\
    
    LLARVA \cite{niu2024llarva}
    & Panda
    & RPT/Octo
    & $83(13/37\uparrow)$ 
    & Randomized cube tasks assessing fine-grained 2D-to-3D visual-action alignment. \\
    
    MALMM \cite{singh2024malmm}
    & Panda
    & SA
    & $56(30\uparrow)$ 
    & Custom zero-shot tabletop tasks assessing semantic reasoning and visual grounding. \\
    
    VLA-Touch \cite{bi2025vlatouch}
    & Panda
    & RDT-1B
    & $53.3(23.3\uparrow)$ 
    & Custom contact-rich tasks assessing tactile grasping, pressure, and sustained contact. \\

    \bottomrule
\end{tabularx}
\end{table*}



{
Evaluating VLA models in the real world presents a unique set of challenges. As summarized in Table \ref{tab:real_exp}, real-world experimental setups are highly customized, varying significantly in robotic hardware (e.g., Franka, WidowX, Unitree), baseline comparisons, and task complexity. Consequently, directly comparing absolute Success Rates (SR) across different methods is often infeasible. Instead, a more rigorous analytical approach involves examining the relative performance gains over established baselines (such as OpenVLA \cite{kim2024openvla}, RT-1 \cite{brohan2022rt}, or Diffusion Policy \cite{chi2023diffusion}) and identifying the specific task characteristics that highlight the structural advantages of either monolithic or hierarchical paradigms.

\textbf{Capabilities and Strengths of Monolithic Architectures.} Monolithic models, encompassing both single-system and tightly coupled dual-system designs, consistently demonstrate strong performance in dynamic, contact-rich, and multi-modal environments. By mapping observations directly to actions within a unified latent space, these models excel at tasks requiring high-frequency reactive control and seamless sensory integration. For instance, as shown in Table \ref{tab:real_exp}, VILA \cite{zhang2025vtla} achieves a remarkable 97.5\% SR in contact-rich peg-in-hole assemblies, highlighting the efficacy of an end-to-end architecture in preserving subtle physical and tactile nuances that might be lost during intermediate feature extraction. Furthermore, monolithic models exhibit robust local generalization against visual distractors and out-of-distribution objects. Models like OpenVLA \cite{kim2024openvla} and TraceVLA \cite{zheng2024tracevla} show significant SR improvements (16.8\% and 57.5\%, respectively) over standard baselines when confronted with unseen backgrounds and randomized initial positions. Moreover, unified architectures naturally facilitate multi-modal fusion; methods like FuSe \cite{jones2025beyond} successfully integrate visual, auditory, and tactile reasoning, yielding a 22.7\% performance boost in custom multimodal tasks.

\textbf{Capabilities and Strengths of Hierarchical Architectures.} Conversely, hierarchical architectures demonstrate profound advantages in extreme zero-shot generalization, long-horizon multi-stage planning, and cross-embodiment transfer. By explicitly decoupling high-level semantic reasoning from low-level continuous control, these models can leverage the immense cognitive capabilities of foundational VLMs without being bottlenecked by the strict real-time constraints of actuator control. Table \ref{tab:real_exp} reveals that hierarchical frameworks often achieve massive relative improvements in structurally complex scenarios. For example, RoboMatrix \cite{mao2024robomatrix} demonstrates up to a 100\% relative improvement in zero-shot skill recombination, while DexGrasp \cite{zhong2025dexgraspvla} yields a 61.4\% gain in dexterous generalization across extreme domain shifts. Because the intermediate representations (such as 3D keypoints, affordance maps, or visual traces) are highly structured, hierarchical models like PointVLA \cite{li2025pointvla} and $A_0$ \cite{xu2025a0} excel in 3D-aware spatial reasoning and affordance-guided manipulation. This modularity allows the high-level planner to adapt to novel tasks and instructions intuitively, bridging the cognition-execution gap with highly interpretable sub-goals.

\textbf{Analytical Trade-offs: Monolithic vs. Hierarchical Designs.} A direct comparison between the two paradigms emerges when analyzing hierarchical models that explicitly benchmark against established monolithic systems (e.g., OpenVLA \cite{kim2024openvla}, $\pi_0$ \cite{black2025pi0}, and RT-2 \cite{brohan2023rt2visionlanguageactionmodelstransfer}). The empirical data indicates that while monolithic models provide a robust foundation for generalist control, hierarchical models significantly outperform them in long-horizon planning and structurally complex reasoning. For example, in zero-shot skill recombination and long-horizon tasks, RoboMatrix \cite{mao2024robomatrix} achieves a 100\% relative SR improvement over OpenVLA, while DexVLA \cite{wen2025dexvla} outperforms $\pi_0$ by 13.5\% in tasks evaluating implicit reasoning and cross-embodiment dexterity. Furthermore, when tasks require contextual sequential execution or spatial composition, hierarchical models like RT-H \cite{belkhale2024rt} and VoxPoser \cite{huang2023voxposer} surpass RT-2 by 15\% and 64\%, respectively, highlighting the benefits of decoupling dynamic robustness from language interventions. Ultimately, this comparative data suggests a clear trade-off: monolithic models are highly capable standalone generalists optimized for reactive, visually diverse, and contact-rich environments, whereas hierarchical models are structurally necessary for overcoming the performance ceilings of end-to-end systems in long-horizon, multi-stage, and zero-shot skill transfer scenarios.
}

\section{Datasets and Benchmarks}
\label{db}

\subsection{Real-world Robot Datasets}

Real-world data, capturing environmental complexity such as dynamic lighting, background motion, and varied object properties, is vital for training and evaluating VLA models. Effective language grounding and semantic generalization rely on aligning natural language instructions, perceptual inputs, and precise action responses.
To support this, a series of large-scale real-world datasets have emerged. BC-Z~\cite{Bc-z} provides expert demonstrations with language commands across 100 tasks; RT-1~\cite{brohan2022rt} and its kitchen variant extend this to over 700 daily activities, while RT-2~\cite{brohan2023rt2visionlanguageactionmodelstransfer} incorporates web-scale vision-language data for broader vocabulary and instruction generalization. RoboFlamingo~\cite{li2023vision} exemplifies the transfer of pretrained multimodal priors into real-world control. BridgeData V2~\cite{ebert2021bridge} and RH20T~\cite{fang2023rh20t} further enhance cross-domain generalization through extensive, language-annotated demonstrations, with RH20T enabling one-shot learning across 147 tasks. DROID~\cite{khazatsky2024droid} expands this landscape with teleoperated demonstrations ``in the wild,'' offering diverse visual contexts over 564 tasks. Most notably, the OXE dataset~\cite{o2024open} unifies over 1 million multimodal demonstrations across 22 robot platforms and 500+ skills, marking a significant step toward web-scale, cross-embodiment generalization. Although these efforts improve real-world applicability, the long-tail distribution of open-world objects, scenes, and skills remains underrepresented, underscoring the need for broader and more diverse real-world data to scale large VLM-based VLA models further.

\begin{figure}[t]
    \centering
    \includegraphics[width=1.0\linewidth]{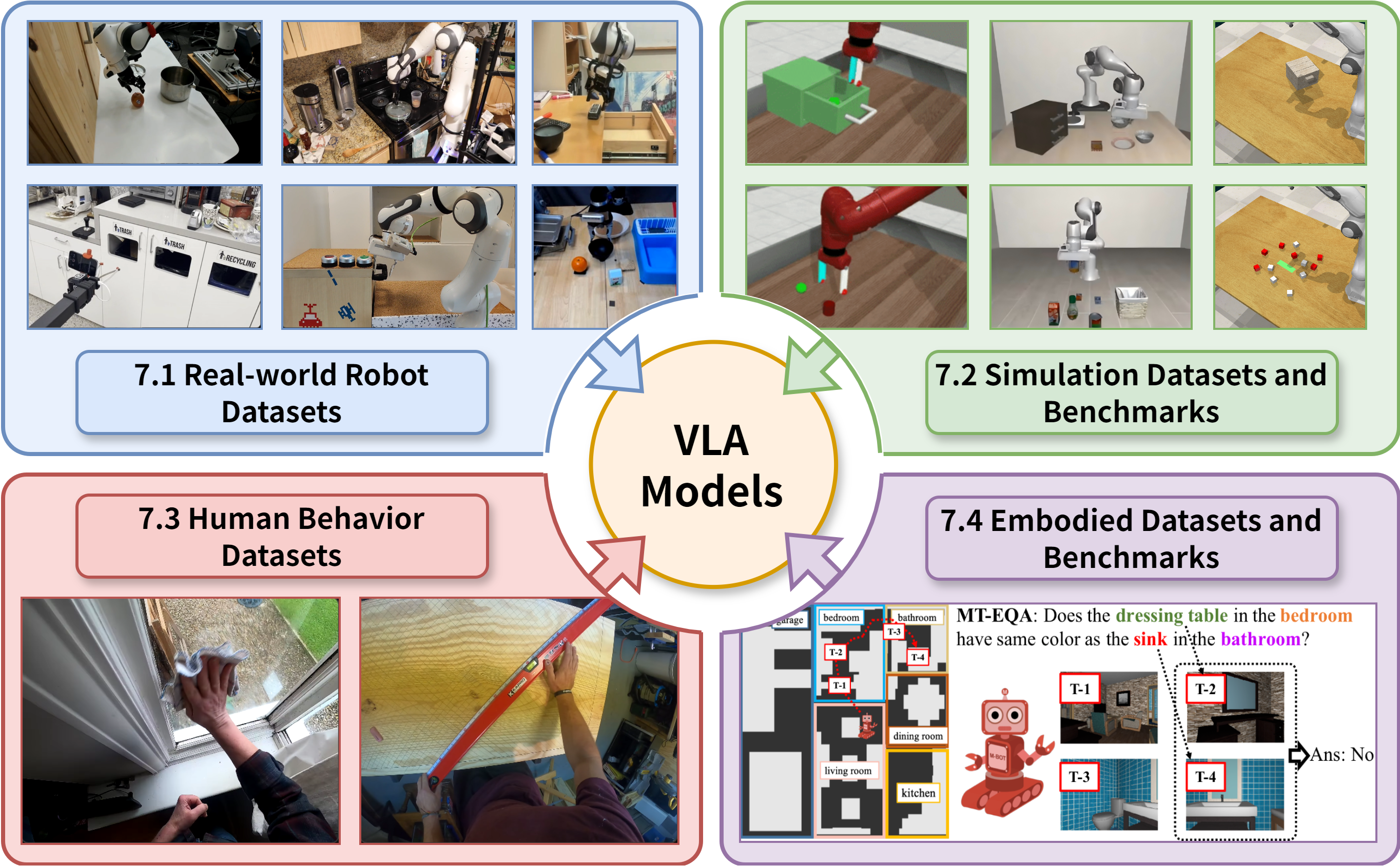}
    \caption{Illustration of four dataset types that underpin large VLM-based VLA models for robotic manipulation.}
    \label{fig:data}
\end{figure}

\subsection{Simulation Datasets and Benchmarks}
To overcome real-world data limitations, physics-based simulators and virtual environments offer scalable, safe, and reproducible interaction data, aligning with the multimodal, large-scale training needs of large VLM-based VLA models. They enable complex instruction following, multi-stage planning, and consistent language grounding with automatic evaluation.
BEHAVIOR~\cite{li2024behavior} supports multi-step semantic control in cluttered household settings; ALFRED~\cite{shridhar2020alfred} focuses on long-horizon tasks specified by egocentric language instructions; RLBench~\cite{james2020rlbench} allows fine-grained policy learning through RGB-D tabletop manipulation, and RLBench2~\cite{grotz2024peract2} extends it to bimanual manipulation with language-conditioned action prediction. Benchmarks such as Meta-World~\cite{yu2020meta}, Franka Kitchen~\cite{FrankaKitchen}, and LIBERO~\cite{liu2023libero} further advance this direction by incorporating multi-skill scenarios, often enriched with language annotations and vision-language rewards to encourage generalization. CALVIN~\cite{mees2022calvin} enables multi-stage manipulation under unconstrained language instructions to support long-horizon behaviors. MIKASA-Robo~\cite{cherepanov2025mikasa} addresses memory-centric challenges by evaluating agents under partial observability in tabletop tasks. SIMPLER~\cite{li24simpler} bridges the sim-to-real gap through calibrated simulated environments aligned with common real-world robot configurations, reducing discrepancies in perception and control. High-fidelity platforms such as Habitat~\cite{savva2019habitat, szot2021habitat, puig2023habitat} and SAPIEN~\cite{xiang2020sapien} support spatial reasoning and manipulation learning in realistic 3D environments, enabling scalable embodied VLA research. 
Beyond traditional simulation benchmarks, THE COLOSSEUM~\cite{pumacay2024colosseum} and VLABench~\cite{zhang2024vlabench} extend simulation benchmarks by assessing robustness under distribution shifts and supporting universal language-conditioned manipulation, respectively.
Despite limitations such as imperfect physics and visual artifacts, simulation remains crucial for efficient training, systematic evaluation, and robust real-world generalization.

\subsection{Human Behavior Datasets}

Large-scale human behavior datasets offer semantically rich and contextually diverse demonstrations that are well-suited for pretraining VLA models. 
Egocentric video corpora such as Ego4D~\cite{kris2021ego}, Ego-Exo4D~\cite{grauman2024ego}, EgoPlan-Bench~\cite{chen2023egoplan}, and EgoVid-5~\cite{wang2024egovid} capture diverse daily activities from a first-person perspective. In contrast, domain-specific datasets like EPIC-Kitchens~\cite{dima2018epic} and COM-Kitchens~\cite{maeda2024kitchens} focus on fine-grained, temporally structured cooking tasks. These resources support learning of object recognition, action sequencing, and task decomposition. Reasoning-oriented datasets such as EgoVQA~\cite{fan2019egovqa} and EgoTaskQA~\cite{jia2022egotaskqa} further enhance cognitive grounding by introducing spatial, temporal, and causal reasoning over egocentric video. 
To address the lack of precise manipulation signals in general-purpose corpora like Ego4D, EgoDex~\cite{hoque2025egodex} extends this line of work with 829 hours of egocentric video paired with dense 3D hand and finger tracking. Similarly, smaller manipulation-focused datasets such as DexCap~\cite{wang2024dexcap}, EgoMimic~\cite{kareer2024egomimic}, and PH$^2$D~\cite{PH2D} offer fine-grained supervision for hand-object coordination and embodiment-aware policy learning.
Complementing these human-centric datasets, R2R2R~\cite{yu2025real2render2real} offers a scalable pipeline that replaces teleoperation with smartphone scans and human videos to generate high-fidelity demonstrations. Together, these resources provide a rich spectrum of semantic grounding, embodiment alignment, and data scalability essential for training generalist VLA models.

\subsection{Embodied Datasets and Benchmarks}

As robots advance toward greater autonomy, understanding and planning tasks become essential. Leveraging strong language and world knowledge, VLMs empower VLA systems to plan over long horizons, parse complex instructions, and build cross-modal reasoning for semantic decision-making. In response, a growing set of benchmarks has emerged to evaluate embodied agents' planning and reasoning abilities. Early efforts such as EmbodiedQA~\cite{das2018embodied} and IQUAD~\cite{gordon2018iqa} established the integration of vision, language, and spatial navigation through question-answering in 3D environments. These have since evolved into more cognitively demanding tasks. MT-EQA~\cite{MT-EQA} incorporates multi-target reasoning, MP3D-EQA~\cite{MP3D-EQA} introduces 3D point cloud inputs for enhanced spatial grounding, EQA-MX~\cite{islam2023eqa} adds non-verbal cues for naturalistic multimodal interaction, and OpenEQA~\cite{majumdar2024openeqa} expands the scope to open-ended questions involving functional and commonsense reasoning. Complementarily, LoTa-Bench~\cite{choi2024lota} directly evaluates plan executability by deploying LLM-generated plans in simulation.
Collectively, these benchmarks emphasize active perception, reasoning, and execution, offering rigorous protocols to assess high-level semantic planning in general-purpose VLA.

\section{Outlook}
\subsection{Future Directions}
\label{future}

\textbf{Datasets and Benchmarking.} As VLA models advance, the reliance on simulated datasets reveals a significant reality gap. Synthetic scenes lack the visual complexity of real environments. By contrast, collecting real-world data is costly, limiting both diversity and scale. Meanwhile, existing benchmarks typically focus on short-horizon pick-and-place tasks and report simple success rates. These settings offer little insight into practical challenges such as long-term planning, mobile manipulation and multi-agent collaboration. Future datasets may combine large-scale real-world data acquisition with task suites that capture practical challenges. Evaluations should include richer metrics, such as subtask success, time efficiency, and robustness to disturbances. Such datasets and benchmarks would enable systematic progress on the broader capabilities of VLA systems.

\textbf{Memory Mechanisms and Long-Term Planning.} Effective real-world manipulation often requires planning over extended horizons and recalling past observations. However, most current VLAs rely on frame-by-frame reasoning, resulting in short-sighted behavior and a limited ability to leverage historical context. 
To address this, a promising direction is to design architectures that incorporate forward-looking planning and memory mechanisms with episodic awareness.
By grounding high-level decisions in contextual memory, agents may move beyond reactive responses toward coherent, goal-driven action sequences, enabling more consistent and effective manipulation in complex scenarios.

\textbf{3D and 4D Perception.}
Robotic manipulation inherently involves interacting with objects in a three-dimensional and temporally dynamic environment. However, existing VLA models operate primarily on static 2D visual inputs. 
This limits the agent's ability to reason about complex spatial and temporal information, including depth, affordances, object movement, and human actions.
Moving beyond static 2D snapshots, 4D perception demands a model to understand how 3D scenes evolve over time. The progression from 2D to 4D perception requires integrating depth or point cloud observations, as well as fusing multi-modal inputs into unified representations. It also involves embedding perception that is grounded in temporal context and replanning actions on the fly. Together, these 4D-aware abilities would enable more robust and adaptive manipulation in the real world.

\textbf{Mobile Manipulation.}
Real-world tasks often demand the simultaneous execution of locomotion and manipulation, giving rise to the specialized domain of mobile manipulation.
This task requires a synergistic integration of navigation capabilities and interactive skills with the environment, introducing tightly coupled demands on perception and control. 
Rather than treating locomotion and grasping as separate stages, future VLA models may benefit from learning integrated policies that adaptively prioritize locomotion and arm coordination, ultimately leading to more robust and flexible mobile manipulation.

\textbf{Multi-Agent Cooperation.} Many real-world collaborative tasks, such as bulky objects moving or joint tool use, require agents to negotiate intentions, adapt to teammates’ actions, and jointly reason over multi-step goals. However, single-agent VLAs often falter when communication and role assignment become critical, highlighting the need for interaction-aware representations. Integrating emergent dialogue protocols with shared world models may enable agents to synchronize plans and flexibly allocate subtasks. Such cooperative capabilities promise to elevate robotic teams from loosely coordinated individuals to cohesive collaborators capable of addressing complex group objectives.

\textbf{Lifelong Learning in Open-World.} Robotic autonomy demands continual skill acquisition without catastrophic forgetting.
However, most VLA models trained on static datasets struggle to handle unfamiliar objects and novel interaction modes. They also have difficulty incorporating new experiences. This reveals the limits of static training paradigms.
Future VLA systems may benefit from mechanisms that incrementally accumulate knowledge through exploration and feedback in open-world. Incorporating memory structures that grow over time and enabling agents to self-organize experiences into reusable abstractions may also provide a foundation for long-term competence.

\textbf{Model Efficiency.} Though VLA models excel at understanding vision-language instructions and planning actions, they often suffer from prohibitive computational and memory costs. Deploying these models on resource-constrained robotic platforms raises concerns about latency, memory usage, and sustained operation. The key challenges lie in balancing model capacity with real-time inference requirements and preserving multimodal alignment to maintain accuracy during compression. Promising directions include task-aware dynamic token pruning, asynchronous inference for seamless transitions between action chunks, and hardware-friendly quantization schemes.

\subsection{{Discussion}}
{As the field of vision language action modeling rapidly matures, several core agreements, emerging paradigms, and theoretical disputes have surfaced. By synthesizing the recent advancements reviewed in this survey, we provide a critical subjective analysis of the current research trajectory.}

{\textbf{Established and Emerging Consensus.} A firm consensus has emerged within the community regarding the efficacy of transferring internet-scale visual-semantic priors to physical embodiments. Researchers uniformly agree that initializing robotic policies with massive pre-trained large vision-language models fundamentally addresses the semantic grounding bottleneck, enabling models to demonstrate zero-shot generalization potential when encountering novel objects and unconstrained linguistic instructions. Furthermore, an emerging consensus is shifting the perspective on low-level action generation. While early models relied heavily on pure discrete autoregressive token prediction for control, the community increasingly recognizes that representing continuous, high-precision physical movements via discrete text tokens suffers from compounding spatial errors and temporal instability in long-horizon tasks. Consequently, integrating continuous action generation paradigms (such as diffusion policies or flow-matching networks) as the execution engine is rapidly becoming the accepted standard for achieving smooth and reliable control.}

{\textbf{Key Controversies and Opposing Views.} Despite these agreements, significant theoretical divides remain. The most prominent controversy centers on the integration of explicit world models versus purely implicit end to end architectures. One camp argues that large monolithic VLA models can implicitly internalize the laws of physics and environmental dynamics simply by scaling up model parameters and robot interaction data, rendering separate simulation modules redundant. Conversely, opposing researchers argue that without explicit predictive world models to simulate future states and enforce geometric constraints, VLAs will always remain vulnerable to severe hallucinations and visual kinematic overfitting in out of distribution scenarios. Critics of the explicit world model approach counter this by pointing out that simulating high fidelity video or 3D environments adds prohibitive computational overhead, creating unacceptable latency for real time reactive control.}

{\textbf{Critical Insights: Successes and Persistent Bottlenecks.} Although current VLA architectures excel at interpreting complex human intents and achieving broad semantic generalization, a critical examination reveals profound structural limitations in physical grounding, data scaling, temporal control, and explicit environmental modeling that severely bottleneck true real world autonomy.}
\begin{itemize}
    \item {\textbf{Illusory Physical Reasoning and Statistical Shortcuts.} Large vision-language models fundamentally operate as statistical pattern learners over 2D data, lacking true causal grounding in 3D physical dynamics. Apparent task success (e.g., object placement) largely stems from exploiting language priors and visual co-occurrence, rather than understanding underlying physics, revealing a reliance on trajectory imitation instead of genuine embodied cognition.}
    \item {\textbf{Embodiment Heterogeneity and the Breakdown of Scaling.} Robot data is intrinsically heterogeneous, entangled with kinematics, sensing configurations, and embodiment-specific constraints. Naïve data aggregation induces distributional conflicts and performance collapse, in contrast to the scaling benefits observed in language models. Coupled with the high cost of teleoperation data, this limits scalability unless new paradigms beyond imitation learning (such as large-scale autonomous reinforcement learning) are realized.}
    \item {\textbf{Pseudo Closed-Loop Control and Missing Temporal Dynamics.} Despite end-to-end claims, most VLAs generate open-loop action sequences delegated to low-level controllers. This induces temporal misalignment and weak responsiveness in dynamic, contact-rich settings, leading to error accumulation over long horizons due to the absence of continuous-time dynamics modeling.}
    \item {\textbf{Implicit Memorization vs. Explicit World Modeling.} Monolithic VLAs rely on implicit parameterized memorization of physical regularities, lacking explicit geometric and causal constraints, and thus are prone to hallucination under distribution shift. Explicit world models provide principled physical simulation but incur prohibitive computational costs for real-time control. This trade-off between scalability and physical fidelity remains a core barrier to safe deployment.}
\end{itemize}
{Ultimately, overcoming these foundational barriers requires the robotics community to move beyond simply scaling static two dimensional pattern matchers. Future breakthroughs will depend on developing native multimodal architectures that seamlessly integrate explicit causal physical reasoning, scalable autonomous exploration, and high frequency continuous control.}

\newcolumntype{L}[1]{>{\raggedright\arraybackslash}p{#1}}

\section{Practical Model-Selection Guidelines}
\label{app:model_selection}
This appendix complements the architecture-oriented taxonomy in the main
manuscript with a model-selection perspective. For each representative
method, we identify: (1) the principal baseline or architectural predecessor;
(2) the limitation addressed by the proposed innovation; (3) the corresponding trade-off; and (4) the task or deployment scenario for which the method is most
appropriate. Because the reported results are obtained using different
platforms, datasets, and evaluation protocols, numerical values
should not be compared directly across rows and should not be interpreted as a
global ranking.

\textbf{Single-System Models.}
As summarized in Table~\ref{tab:selection-single}, OpenVLA~\cite{kim2024openvla} is a strong default when openness, reproducibility, and a mature fine-tuning ecosystem are priorities. When the main bottleneck is semantic or visual generalization, ECoT~\cite{chen2025trainingstrategiesefficientembodied}, ReVLA~\cite{dey2024revla}, TraceVLA~\cite{zheng2024tracevla}, and SpatialVLA~\cite{qu2025spatialvlaexploringspatialrepresentations} target reasoning, visual-domain shift, temporal motion cues, and 3D spatial structure,
respectively, but introduce additional annotations, sensing assumptions, or
computation. For deployment speed, RoboMamba~\cite{liu2024robomamba} reduces backbone and head cost, whereas OpenVLA-OFT~\cite{kim2025oft} preserves the OpenVLA starting point and obtains high-rate continuous action chunks through task-specific adaptation. Therefore, single-system efficiency claims should always be read together with the action interface and evaluation scope.

\begin{table*}[t]
\centering
\caption{Practical model-selection guide for representative single-system VLA
methods. A baseline denotes the architectural predecessor or principal
comparator used by the cited paper. Numeric gains are paper-specific and are
not comparable across rows because datasets, robots, metrics, and protocols
differ.}
\label{tab:selection-single}
\scriptsize
\setlength{\tabcolsep}{2.4pt}
\renewcommand{\arraystretch}{1.14}
\begin{tabularx}{\textwidth}{L{0.095\textwidth}L{0.13\textwidth}L{0.245\textwidth}L{0.245\textwidth}X}
\toprule
Method & Primary baseline(s) & Limitation addressed and evidence & Trade-off & Recommended scenario \\
\midrule
RT-2~\cite{brohan2023rt2visionlanguageactionmodelstransfer} & RT-1; VC-1 & Co-fines-tunes web vision-language and robot data; PaLI-X-55B reports about 59\% average on emergent-skill tests versus about 17\% for RT-1. & The strongest version is very large and closed; autoregressive action tokens raise compute, latency, and reproducibility barriers. & Semantic OOD commands and web-knowledge transfer when datacenter-scale inference is acceptable. \\

\rowcolor{lightgray}
OpenVLA~\cite{kim2024openvla} & RT-2-X; Diffusion Policy for adaptation & Open 7B generalist trained on 970k demonstrations; reports $+16.5$ points over RT-2-X across 29 tasks and $+20.4$ points over Diffusion Policy. & Still a 7B autoregressive policy; new embodiments or high-frequency tasks generally require task-specific fine-tuning and serving optimization. & Default open, reproducible baseline for generalist manipulation, ablations, and downstream fine-tuning. \\

ECoT~\cite{chen2025trainingstrategiesefficientembodied} & OpenVLA & Adds grounded reasoning over plans, subtasks, motions, boxes, and end-effector states; reports a 28-point gain on generalization tasks. & Requires synthetic reasoning annotations and extra autoregressive reasoning tokens, increasing pre-action latency. & Semantic ambiguity, OOD generalization, failure diagnosis, and human correction rather than tight control loops. \\

\rowcolor{lightgray}
ReVLA~\cite{dey2024revla} & OpenVLA & Merges the vision backbone toward its pretrained state to reduce forgetting; reports 77\% and 66\% relative gains for visual-OOD grasping and lifting. & Targets visual domain shift, not control frequency or dynamics; performance depends on the merging schedule. & Stable action spaces with large shifts in lighting, textures, backgrounds, or object appearance. \\

TraceVLA~\cite{zheng2024tracevla} & OpenVLA & Encodes state-action history as traces; reports $+10$\% on SimplerEnv and $3.5\times$ success on its physical-robot tasks. & Needs trace construction and 150k additional trajectories; benefits depend on tracking quality. & Motion sensitive or progress dependent tasks where a single frame is insufficient. \\

\rowcolor{lightgray}
SpatialVLA~\cite{qu2025spatialvlaexploringspatialrepresentations} & OpenVLA; RT-1-X/RT-2-X; Octo & Adds depth-derived egocentric 3D encoding and action grids; reports 20.1~Hz versus 5.0~Hz for OpenVLA, with stronger spatial transfer. & Depends on depth estimation and camera intrinsics, uses 1.1M episodes, and re-discretizes action grids for a new robot. & 3D layout changes, cross-robot spatial transfer, and geometry-dominated tasks. \\

RoboMamba \cite{liu2024robomamba} & OpenVLA-7B; ManipLLM-7B & A 2.8B Mamba plus 3.7M pose head runs at 9.0~Hz on A100 versus 3.4/1.1~Hz; the OpenVLA comparison is about $2.6\times$. & Speed is reported for a narrower SE(3) pose interface, not a matched cross-embodiment generalist-policy comparison. & Efficient pose-level reasoning/control within the supported setup. \\

\rowcolor{lightgray}
OpenVLA-OFT~\cite{kim2025oft} & OpenVLA; adapted $\pi_0$/RDT-1B & Parallel decoding, chunks, continuous actions, and L1 training raise LIBERO average success from 76.5\% to 97.1\% and throughput by $26\times$. & An adaptation recipe, not a zero-shot replacement; needs robot/task demonstrations and changes the base action interface. & High-frequency, dexterous, or bimanual deployment starting from OpenVLA. \\
\bottomrule
\end{tabularx}
\end{table*}

\textbf{Dual-System Models.}
As summarized in Table~\ref{tab:selection-dual}, dual-system designs are preferable when continuous or high-frequency action generation is the primary requirement. $\pi_0$ \cite{black2024pi0} provides a strong generalist flow-matching foundation, while $\pi_{0.5}$ \cite{pi052025phy} adds semantic co-training for open-world long-horizon tasks. CogACT \cite{li2024cogact} and RoboDual \cite{bu2024towards} are suitable when a specialized action expert can be trained for higher precision or control
rate. SmolVLA \cite{smolvla} targets resource-constrained deployment, GR00T~N1 \cite{bjorck2025gr00t} targets humanoid and bimanual systems, and ForceVLA \cite{yu2025forcevla} targets contact-rich tasks with force-torque sensing. These benefits come at the cost of an additional action module, or a more complex training/data recipe.

\begin{table*}[t]
\centering
\caption{Practical model-selection guide for representative dual-system VLA methods.}
\label{tab:selection-dual}
\scriptsize
\setlength{\tabcolsep}{2.4pt}
\renewcommand{\arraystretch}{1.14}
\begin{tabularx}{\textwidth}{L{0.095\textwidth}L{0.13\textwidth}L{0.235\textwidth}L{0.245\textwidth}X}
\toprule
Method & Primary baseline(s) & Limitation addressed and evidence & Trade-off & Recommended scenario \\
\midrule
$\pi_0$~\cite{black2024pi0} & Autoregressive VLAs such as RT-2/OpenVLA & Adds a flow-matching action expert for continuous chunks and reports control up to 50~Hz across dexterous, cross-embodiment tasks. & Relies on more than 10,000 hours of robot data, substantial compute, and iterative flow generation; the latent interface is not human-interpretable. & High-frequency continuous control and dexterity when large-scale data and compute are available. \\

\rowcolor{lightgray}
$\pi_{0.5}$~\cite{pi052025phy} & $\pi_0$ & Adds heterogeneous co-training and semantic/subtask prediction, enabling long-horizon manipulation in entirely unseen homes. & Broader capability comes from a more complex data mixture and annotation/training recipe; the system remains resource intensive. & Open-world household tasks with long horizons, semantic decisions, and novel environments. \\

CogACT~\cite{li2024cogact} & OpenVLA; RT-2-X & Replaces action quantization with a diffusion action transformer; reports over 35\% gains over OpenVLA in simulation and 55\% in real-world. & Adds a separately parameterized denoising module and tighter VLM-policy coupling, increasing training and serving complexity. & Precise, multimodal action distributions when minimal architecture and explicit plans are secondary. \\

\rowcolor{lightgray}
RoboDual~\cite{bu2024towards} & OpenVLA & Adds a 20M specialist; reports $+26.7$ points in real-world tests, $+12$ points on CALVIN, and $3.8\times$ control frequency. & The specialist needs in-domain demonstrations and another maintained module; gains may not transfer unchanged across embodiments. & Known deployment domains with limited demonstrations and strict control-rate requirements. \\
SmolVLA~\cite{smolvla} & $\pi_0$; ACT; OpenVLA/Octo comparisons & A 0.45B model reaches 87.3\% LIBERO average versus 86.0\% for robotics pretrained $\pi_0$, while training about 40\% faster with $6\times$ less memory. & Pretraining uses about 23k trajectories and one robot family; longer action execution improves speed but reduces responsiveness. & Consumer GPUs/CPUs, low-cost arms, teaching labs, and rapid iteration. \\

\rowcolor{lightgray}
GR00T~N1~\cite{bjorck2025gr00t} & BC-Transformer; Diffusion Policy & Combines an Eagle VLM, diffusion transformer, and real/human-video/synthetic data; reports strong multi-embodiment humanoid results. & Humanoid-oriented post-training, simulation assets, and embodiment-specific data remain substantial integration requirements. & Humanoid or bimanual industrial manipulation with synthetic-data and post-training infrastructure. \\

TinyVLA~\cite{wen2025tinyvla} & OpenVLA & Uses a compact backbone and diffusion decoder without robot pretraining; reports faster, more data-efficient learning with comparable evaluated success. & Savings come from target-domain fine-tuning, so broad zero-shot and cross-embodiment capability is less established. & Fast prototyping or new tasks with few demonstrations and no large pretraining stage. \\

\rowcolor{lightgray}
ForceVLA~\cite{yu2025forcevla} & $\pi_0$-based baselines & Fuses 6-axis force through a force-aware MoE; reports $+23.2$\% average success and up to 80\% on plug insertion. & Requires synchronized force-torque sensing, calibration, and specialized data; the modality does not help ordinary free-space tasks by default. & Insertion, assembly, wiping, and other contact-rich tasks with occlusion or uncertain contact. \\
\bottomrule
\end{tabularx}
\end{table*}

\textbf{Hierarchical Models.}
As summarized in Table~\ref{tab:selection-hierarchical}, hierarchical models are preferable when a task requires long-horizon decomposition, explicit intermediate representations, human intervention, or verifiable geometric constraints. PaLM-E \cite{driess2023palme} and HiRobot \cite{shi2025hi} are suitable for
high-level semantic planning when a reliable library of low-level skills is
already available. RoboPoint \cite{yuan2024robopoint} and HAMSTER \cite{li2025hamster} use keypoints or paths to bridge
high-level visual reasoning and low-level control, making them attractive for
geometry-intensive manipulation and cross-domain transfer. VoxPoser \cite{huang2023voxposer} and ReKep \cite{huang2024rekep}
are preferable when training-free deployment, explicit 3D constraints, and
inspectable planning are priorities, whereas RT-H \cite{belkhale2024rt} supports interpretable language-level intervention during execution. These advantages come at the cost of additional planner--policy interfaces, possible error propagation between levels, increased inference latency, or dependence on low-level controllers.
The choice of a hierarchical method should be based on the desired
intermediate representation and the available downstream control components.

\begin{table*}[t]
\centering
\caption{Practical model-selection guide for representative hierarchical VLA methods.}
\label{tab:selection-hierarchical}
\scriptsize
\setlength{\tabcolsep}{2.4pt}
\renewcommand{\arraystretch}{1.14}
\begin{tabularx}{\textwidth}{L{0.095\textwidth}L{0.13\textwidth}L{0.235\textwidth}L{0.245\textwidth}X}
\toprule
Method & Primary baseline(s) & Limitation addressed and evidence & Trade-off & Recommended scenario \\
\midrule
PaLM-E~\cite{driess2023palme} & PaLM plus task-specific multimodal/robot planners & Injects continuous sensor observations into an LLM and jointly trains robotics, VQA, and captioning, demonstrating positive transfer across embodiments. & Planner outputs need a downstream controller; the largest 562B model is costly and does not directly provide high-frequency actions. & High-level multimodal planning when broad reasoning is primary and reliable skills/controllers already exist. \\

\rowcolor{lightgray}
RoboPoint~\cite{yuan2024robopoint} & GPT-4o; PIVOT & Trains a VLM to emit 2D affordance keypoints from synthetic data; reports $+21.8$\% affordance accuracy and $+30.5$\% downstream success. & Produces points rather than control; accuracy can inherit synthetic-data and projection biases and needs a motion planner/policy. & Scalable spatial grounding, pick/place targets, navigation points, and AR assistance. \\

VoxPoser~\cite{huang2023voxposer} & Code as Policies; language-conditioned manipulation baselines & Composes 3D value maps and model-based planning for zero-shot closed-loop trajectories over open-set instructions and objects. & Depends on RGB-D, open-vocabulary models, code generation, and optimization; component errors compound and contact-rich dynamics need extra modeling. & Training-free open-vocabulary tabletop tasks with calibrated 3D sensing and accessible model APIs. \\

\rowcolor{lightgray}
ReKep~\cite{huang2024rekep} & VoxPoser-style value-map planning & Uses executable relational keypoint-cost functions and hierarchical optimization for bimanual and reactive tasks without task data. & Sensitive to segmentation/tracking and constraint-generation errors; nonlinear optimization can be slow or locally infeasible. & Relational geometry, reactivity, and inspectable constraints over end-to-end simplicity. \\

RT-H~\cite{belkhale2024rt} & RT-2-style flat task-to-action prediction & Inserts language motions between task and action, improving sharing across diverse tasks and enabling language intervention/correction. & Requires language-motion supervision or intervention data; language can mis-specify fine geometry and still relies on the low-level policy. & Multitask policies that must accept interpretable human-in-the-loop corrections. \\

\rowcolor{lightgray}
HAMSTER~\cite{li2025hamster} & OpenVLA & Uses VLM-generated 2D paths to condition a 3D policy; reports $+20$ points average across seven generalization axes over OpenVLA. & A 2D path discards depth/contact detail and needs path data plus a trained 3D policy; high-level errors propagate. & Cross-domain transfer from videos, sketches, or simulation to precise real-robot control. \\

HiRobot~\cite{shi2025hi} & Direct-instruction $\pi_0$-style control & Converts open-ended prompts and feedback into atomic commands for a low-level VLA, supporting complex household workflows. & Two-stage inference adds latency and error propagation; success is bounded by the low-level policy's atomic skill coverage. & Long-horizon household tasks with user feedback, semantic preferences, and reliable skills. \\

\rowcolor{lightgray}
DexVLA~\cite{wen2025dexvla} & Octo; OpenVLA; Diffusion Policy & Scales a plug-in diffusion expert to 1B parameters with an embodiment curriculum for complex long-horizon and cross-embodiment control. & The large expert and three-stage recipe raise compute and engineering cost; the latent/subtask interface is less inspectable than keypoint/program hierarchies. & Dexterous long-horizon tasks across single-arm, bimanual, or dexterous-hand platforms. \\
\bottomrule
\end{tabularx}
\end{table*}

Single-system models provide a simple and reproducible starting point, dual-system models are generally better suited to high-frequency and precise continuous control, and hierarchical models are preferable for long-horizon, interpretable, or constraint-driven tasks. Model selection should balance task requirements against control latency, data budgets, sensing assumptions, interpretability, and integration complexity. Tables~\ref{tab:selection-single}--\ref{tab:selection-hierarchical} instantiate these rules with named baselines and explicit trade-offs.

\end{document}